\documentclass{article}

\PassOptionsToPackage{numbers, compress}{natbib}

\usepackage[final]{neurips_2023}

\usepackage[utf8]{inputenc} %
\usepackage[T1]{fontenc}    %
\usepackage[hidelinks]{hyperref}       %
\usepackage{url}            %
\usepackage{booktabs}       %
\usepackage{amsfonts}       %
\usepackage{nicefrac}       %
\usepackage{microtype}      %
\usepackage{xcolor}         %

\usepackage{times}
\usepackage{soul}
\usepackage{url}
\usepackage[small]{caption}
\usepackage{graphicx}
\usepackage{amsmath}
\usepackage{amsthm}
\usepackage{booktabs}
\usepackage{algorithm}
\usepackage{algorithmic}
\usepackage{amssymb}
\usepackage{xcolor}
\usepackage{subcaption}
\usepackage{multirow}
\usepackage{multicol}
\usepackage{wrapfig}

\usepackage{cleveref} %

\newtheorem{definition}{Definition}

\def \EvalEnvShowSize {0.46}

\title{Arbitrarily Scalable Environment Generators via Neural Cellular Automata}

\author{
Yulun Zhang$^1$\And
Matthew C. Fontaine$^2$\And
Varun Bhatt$^2$\And
Stefanos Nikolaidis$^2$\and
\textbf{Jiaoyang Li$^1$} \\
$^1$Robotics Institute, Carnegie Mellon University\\
$^2$Thomas Lord Department of Computer Science, University of Southern California\\
\texttt{yulunzhang@cmu.edu,
\{mfontain,vsbhatt,nikolaid\}@usc.edu,}\\
\texttt{jiaoyangli@cmu.edu}
}

\begin{document}

\maketitle

\begin{abstract}
We study the problem of generating arbitrarily large environments to improve the throughput of multi-robot systems. Prior work proposes Quality Diversity (QD) algorithms as an effective method for optimizing the environments of automated warehouses. However, these approaches optimize only relatively small environments, falling short when it comes to replicating real-world warehouse sizes. The challenge arises from the exponential increase in the search space as the environment size increases. Additionally, the previous methods have only been tested with up to 350 robots in simulations, while practical warehouses could host thousands of robots. In this paper, instead of optimizing environments, we propose to optimize Neural Cellular Automata (NCA) environment generators via QD algorithms. We train a collection of NCA generators with QD algorithms in small environments and then generate arbitrarily large environments from the generators at test time. We show that NCA environment generators maintain consistent, regularized patterns regardless of environment size, significantly enhancing the scalability of multi-robot systems in two different domains with up to 2,350 robots. Additionally, we demonstrate that our method scales a single-agent reinforcement learning policy to arbitrarily large environments with similar patterns. We include the source code at \url{https://github.com/lunjohnzhang/warehouse_env_gen_nca_public}.

\end{abstract}

\section{Introduction}

We study the problem of generating arbitrarily large environments to improve the throughput of multi-robot systems. As a motivating example, consider a multi-robot system for automated warehousing where thousands of robots transport inventory pods in a shared warehouse environment. While numerous works have studied the underlying Multi-Agent Path Finding (MAPF) problem~\cite{SternSoCS19} to more effectively coordinate the robots to improve the throughput~\cite{ChenRAL21,ContiniRAS21,DamaniRAL21,KouAAAI20,LiAAMAS20a,LiuAAMAS19,MaAAMAS17,NguyenIJCAI17,VaramballySoCS22}, we can also optimize the throughput by designing novel warehouse environments. A well-optimized environment can alleviate traffic congestion and reduce the travel distances for the robots to fulfill their tasks in the warehouse.

\begin{figure}[!t]
    \centering
    \begin{minipage}[b]{0.215\textwidth}
        \centering
        \begin{subfigure}[t]{\linewidth}
            \includegraphics[width=\textwidth]{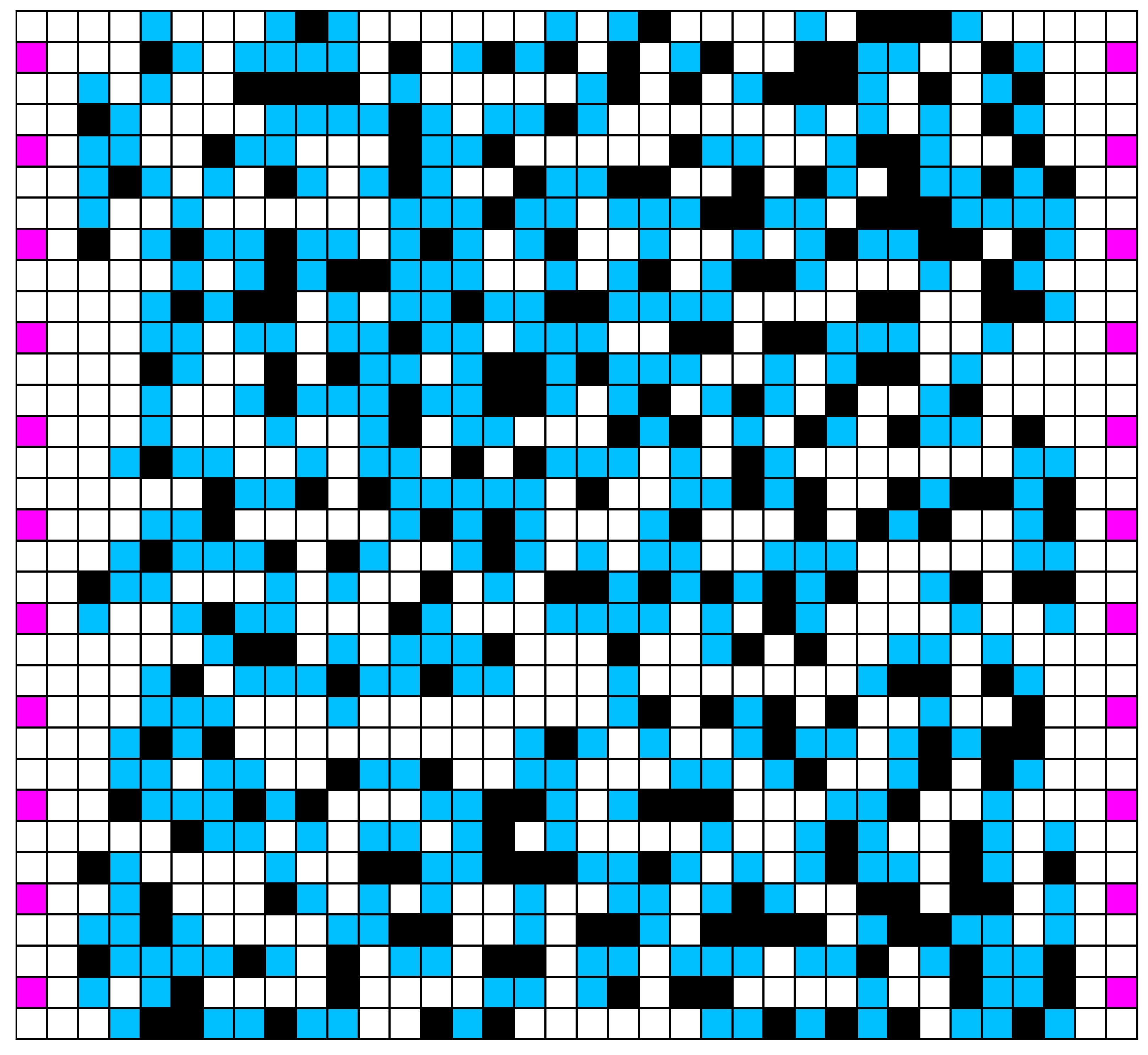}
            \caption{Optimized environment with no patterns.}
            \label{fig:front-fig:dsage-opt}
        \end{subfigure}

        \begin{subfigure}[t]{\linewidth}
            \includegraphics[width=\textwidth]{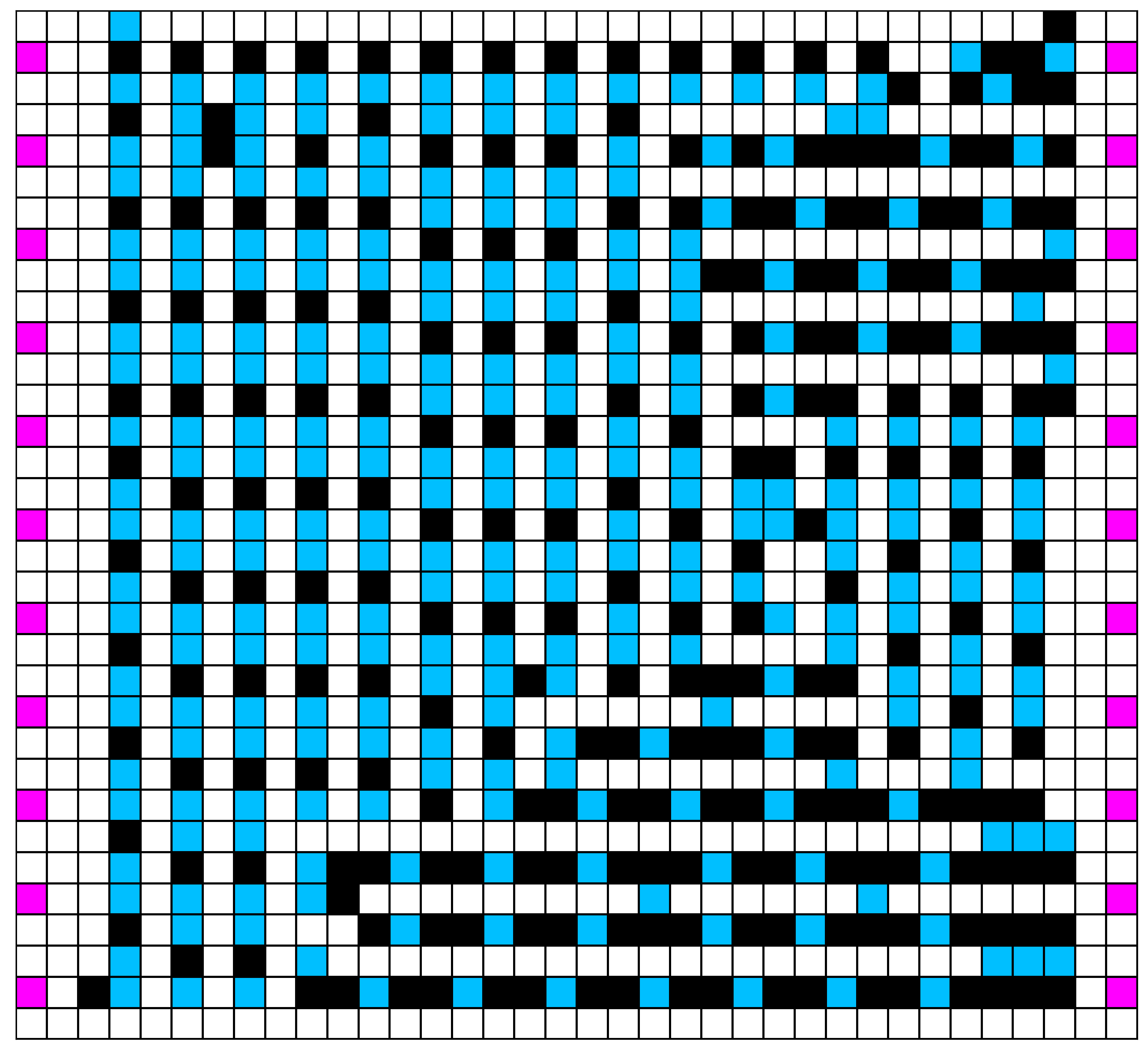}
            \caption{NCA-generated environment with patterns.}
            \label{fig:front-fig:nca-opt}
        \end{subfigure}
        
    \end{minipage}
    \hspace{0.02\textwidth} %
    \begin{minipage}[b]{0.47\textwidth}
        \centering
        \begin{subfigure}[t]{\linewidth}
            \includegraphics[width=\textwidth]{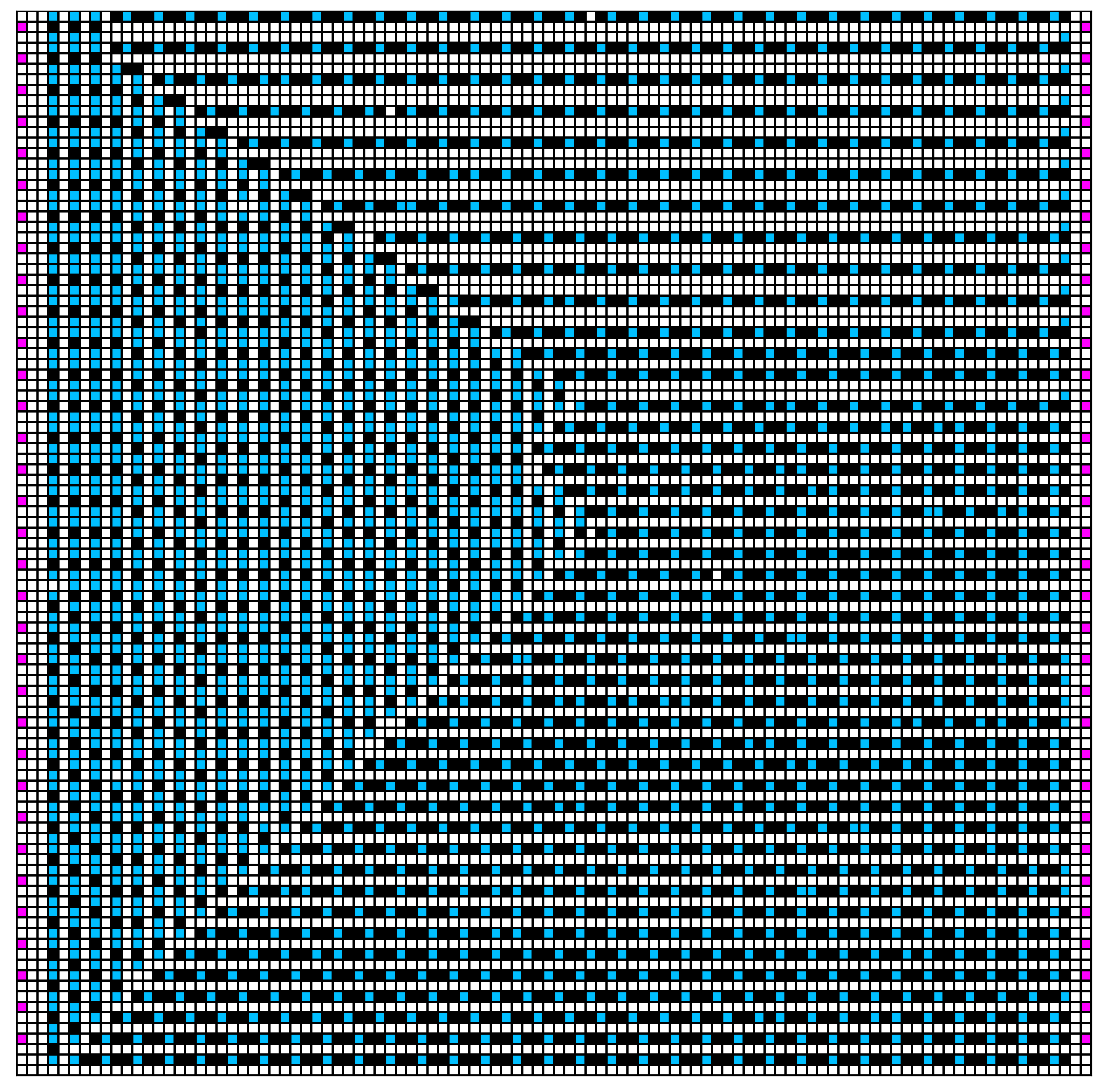}
            \caption{Scaling patterns from the NCA-generated environment to a larger environment.}
            \label{fig:front-fig:xxlarge-nca-opt}
        \end{subfigure}
    \end{minipage}
    \begin{minipage}[b]{0.15\textwidth}
        \centering
        \begin{subfigure}[b]{\linewidth}
            \includegraphics[width=1\textwidth]{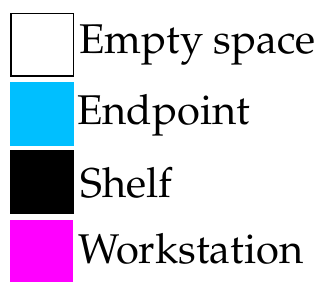}
        \end{subfigure}
        \vspace{4.1cm}
    \end{minipage}
    \caption{Examples of optimized warehouse environments. Robots move in between workstations (pink) and endpoints (blue) without traveling through shelves (black) to transport goods. %
    In this environment, the workstations on the left border are used 5 times more frequently than those on the right border. (a) shows a warehouse environment optimized directly with QD algorithms from previous work~\cite{zhangLayout23}. It has no obvious regularized patterns. (b) shows an environment generated by our NCA generator with regularized patterns. (c) shows a much larger environment generated by the same NCA generator in (b) with similar patterns.}
    \label{fig:front-fig}
\end{figure}

A recent work~\cite{zhangLayout23} formulates the environment optimization problem as a Quality Diversity (QD) optimization problem and optimizes the environments by searching for the best allocation of shelf and endpoint locations (where endpoints are locations where the robots can interact with the shelves). It uses a QD algorithm to iteratively generate new environments and then repairs them with a Mixed Integer Linear Programming (MILP) solver to enforce domain-specific constraints, such as the storage capacity and connectivity of the environment. The repaired environments are evaluated with an agent-based simulator~\cite{Li2020LifelongMP} that simulates the movements of the robots in the warehouse. \Cref{fig:front-fig:dsage-opt} shows an example optimized environment, which is shown to have much higher throughput and is more scalable than human-designed environments.

However, with the aforementioned method, the search space of the QD algorithm grows exponentially with the size of the environment. To optimize arbitrarily large environments, QD algorithms require a substantial number of runs in both the agent-based simulator and the MILP solver. Each run also takes more time to finish. For example, in the aforementioned work~\cite{zhangLayout23}, it took up to 24 hours on a 64-core local machine to optimize a warehouse environment of size only 36 $\times$ 33 with 200 robots, which is smaller than many warehouses in reality or warehouse simulations used in the literature. For example, Amazon fulfillment centers are reported to have more than 4,000 robots~\cite{Brown2023amazonrobot}. Previous research motivated by Amazon sortation centers run simulations in environments of size up to 179 $\times$ 69 with up to 1,000 robots~\cite{Yu2023,Li2020LifelongMP}.

Therefore, instead of optimizing the environments directly, we propose to train Neural Cellular Automata (NCA) environment generators capable of scaling their generated environments arbitrarily. Cellular automata (CA)~\cite{gardner1970} are well suited to arbitrary scaling as they incrementally construct environments through local interactions between cells. Each cell observes the state of its neighbors and incorporates that information into changing its own state. NCA~\cite{mordvintsev2020growing} represents the state change rules via a neural network that can be trained with any optimization method.

We follow the insight from prior work~\citep{Earle2021IlluminatingDN} and use QD algorithms to efficiently train a diverse collection of NCA generators in small environments. We then use the NCA generators to generate arbitrarily large environments with consistent and regularized patterns. In case the generated environments are invalid, we adopt the MILP solver from the previous works~\cite{zhang:aiide2020,fontaine2021importance,zhangLayout23} to repair the environments. \Cref{fig:front-fig:nca-opt} shows an example environment with regularized patterns generated by our NCA generator and then repaired by the MILP solver. \Cref{fig:front-fig:xxlarge-nca-opt} shows a much larger environment generated by the same NCA generator with similar patterns and then repaired by MILP. 
Similar to previous environment optimization methods~\cite{zhangLayout23}, we need to run the MILP solver repeatedly on small environments (\Cref{fig:front-fig:nca-opt}) to train the NCA generator. However, once the NCA generators are trained, we only run the MILP solver once after generating large environments (\Cref{fig:front-fig:xxlarge-nca-opt}).

We show that our generated environments have competitive or better throughput and scalability compared to the existing methods~\cite{zhangLayout23} in small environments. In large environments where the existing methods are not applicable due to computational limits, we show that our generated environments have significantly better throughput and scalability than human-designed environments.

We make the following contributions: (1) instead of directly searching for the best environments, we present a method to train a collection of NCA environment generators on small environments using QD algorithms, (2) we then show that the trained NCA generators can generate environments in arbitrary sizes with consistent regularized patterns, which leads to significantly higher throughput than the baseline environments in two multi-robot systems with large sizes, (3) we additionally demonstrate the general-purpose nature of our method by scaling a single-agent RL policy to arbitrarily large environments with similar patterns, maintaining high success rates.

\section{Background and Related Work}

\subsection{Neural Cellular Automata (NCA)}

Cellular automata (CA)~\citep{Chan2020LeniaAE,reinke2020intrinsically} originated in the artificial life community as a way to model incremental cell development. A CA consists of grid cells and an update rule for how to change a cell based on the state of its neighbors. Later work~\citep{mordvintsev2020growing} showed that these rules could be encoded by a convolutional neural network. By iteratively updating a grid of cells, an NCA is capable of producing complex environments. \Cref{fig:nca_process} in \Cref{appen:nca-process} shows an example NCA environment generation process of 50 iterations from an initial environment shown in \Cref{fig:nca-seed}.

Representing the rules of CAs by a neural network facilitates learning useful rules for the cellular automata. Several works have trained NCA generators to generate images~\cite{mordvintsev2020growing}, textures~\citep{niklasson2021self-organising,Mordvintsev2021NCATG}, and 3D structures~\cite{Sudhakaran2021}. Other works have replaced the convolutional network with other model architectures~\citep{Tesfaldet2022attenNCA,Gala2023EnequivariantGN}. 
In the case of environment optimization, the objective is computed from a non-differentiable agent-based simulator.
Therefore, we choose derivative-free optimization methods to train the NCA generators for environment optimization.

Prior work~\cite{Earle2021IlluminatingDN} shows that derivative-free QD algorithms can efficiently train a diverse collection of NCA video game environment generators. However, our work differs in the following ways: (1) we show how environments generated by NCA generators can scale to arbitrary sizes with similar regularized patterns, (2) in addition to encoding the constraints as part of the objective function of QD algorithms, we also use MILP to enforce the constraints on the generated environments, (3) the objective function of the previous work focus on the reliability, diversity, and validity of the generated environments, while we primarily focus on optimizing throughput, which is a simulated agent-based metric of the multi-robot systems.

\subsection{Quality Diversity (QD) Algorithms}

Although derivative-free single-objective optimization algorithms such as Covariance Matrix Adaptation Evolutionary Strategy (CMA-ES)~\cite{hansen2016cmaes} have been used to optimize derivative-free objective functions, we are interested in QD algorithms because they generate a diverse collection of solutions, providing users a wider range of options in terms of the diversity measure functions.

Inspired by evolutionary algorithms with diversity optimization~\cite{Cully_2015_robot_animal,lehman2011abandoning,lehman2011evolving}, QD algorithms simultaneously optimize an objective function and diversify a set of diversity measure functions to generate a diverse collection of high-quality solutions. QD algorithms have many different variants that incorporate different optimization methods, such as model-based approaches~\cite{Bhatt2022DeepSA,gaier2018dataefficient,Zhang2021DeepSA}, Bayesian optimization~\cite{kent2020bop}, and gradient descent~\cite{fontaine2021differentiable}. 

\noindent \textbf{MAP-Elites.} MAP-Elites~\cite{Cully_2015_robot_animal,mouret2015illuminating} constructs a discretized measure space, referred to as \textit{archive}, from the user-defined diversity measure functions. It tries to find the best solution, referred to as \textit{elite}, in each discretized cell. The QD-score measures the quality and diversity of the elites by summing the objective values of elites in the archive. MAP-Elites generates new solutions either randomly or by mutating existing elites. It then evaluates the solutions and replaces existing elites in the corresponding cells with superior ones. After a fixed number of iterations, MAP-Elites returns the archive with elites.

\noindent \textbf{CMA-MAE.}
We select Covariance Matrix Adaptation MAP-Annealing (CMA-MAE)~\cite{Fontaine2022CovarianceMA,Fontaine_CMAME_2020} as our QD method for training NCA generators because of its state-of-the-art performance in continuous domains. CMA-MAE extends MAP-Elites by incorporating the self-adaptation mechanisms of CMA-ES~\cite{hansen2016cmaes}. CMA-ES maintains a Gaussian distribution, samples from it for new solutions, evaluates them, and then updates the distribution towards the high-objective region of the search space. CMA-MAE incorporates this mechanism to optimize the QD-score. In addition, CMA-MAE implements an archive updating mechanism
to balance exploitation and exploration of the measure space. 
The mechanism introduces a threshold value to each cell in the archive, which determines whether a new solution should be added. The threshold values are iteratively updated via an archive learning rate, with lower learning rates focusing more on exploitation. 
Because of the updating mechanism, some high-quality solutions might be thrown away. Therefore, CMA-MAE maintains two archives, an \textit{optimization archive} implementing the updating mechanism, and a separate \textit{result archive} that does not use the mechanism and stores the actual elites.

\subsection{Automatic Environment Generation and Optimization}

Automatic environment generation has emerged in different research communities for various purposes such as generating diverse game content~\cite{togelius2016procedural}, training more robust RL agents~\cite{Risi2019IncreasingGI,Justesen2018IlluminatingGI}, and generating novel testing scenarios~\cite{Bhatt2022DeepSA,fontaine2021quality}.

In the MAPF community, \citet{zhangLayout23} have proposed an environment optimization method based on QD algorithms MAP-Elites~\cite{Cully_2015_robot_animal,mouret2015illuminating} and Deep Surrogate Assisted Generation of Environments (DSAGE)~\cite{Bhatt2022DeepSA} to improve the throughput of the multi-robot system in automated warehouses. In particular, they represent environments as tiled grids and use QD algorithms to repeatedly propose new warehouse environments by generating random environments or mutating existing environments by randomly replacing the tiles. Then they repair the proposed environments with a MILP solver and evaluate them in an agent-based simulator. The evaluated environments are added to an archive to create a collection of high-throughput environments.

However, since the proposed method directly optimizes warehouse environments, it is difficult to use the method to optimize arbitrarily large environments due to (1) the exponential growth of the search space and (2) the CPU runtime of the MILP solver and the agent-based simulator increase significantly. 
In comparison, we optimize environment generators instead of directly optimizing environments. In addition, we explicitly take the regularized patterns into consideration which allows us to generate arbitrarily large environments with high throughput.

\section{Problem Definition}

We define the environments and the corresponding environment optimization problem as follows.

\begin{definition}[Environment] We represent each environment as a 2D four-neighbor grid, where each tile can be one of $N_{type}$ tile types. $N_{type}$ is determined by the domains.
\end{definition}

\begin{definition}[Valid Environment] An environment is valid iff the assignment of tile types satisfies its domain-specific constraints.
\end{definition}

For example, the warehouse environments shown in \Cref{fig:front-fig} have $N_{type} = 4$ (endpoints, workstations, empty spaces, and shelves). One example domain-specific constraint for warehouse environments is that all non-shelf tiles should be connected so that the robots can reach all endpoints. 

\begin{definition}[Environment Optimization] Given an objective function $f: \mathbf{X} \rightarrow \mathbb{R}$ and a measure function $\mathbf{m}: \mathbf{X} \rightarrow \mathbb{R}^m$, where $\mathbf{X}$ is the space of all possible environments, the environment optimization problem searches for valid environments that maximize the objective function $f$ while diversifying the measure function $\mathbf{m}$.
\end{definition}

\section{Methods} \label{sec:method}

\begin{figure}[!t]
    \centering
    \includegraphics[width=1\textwidth]{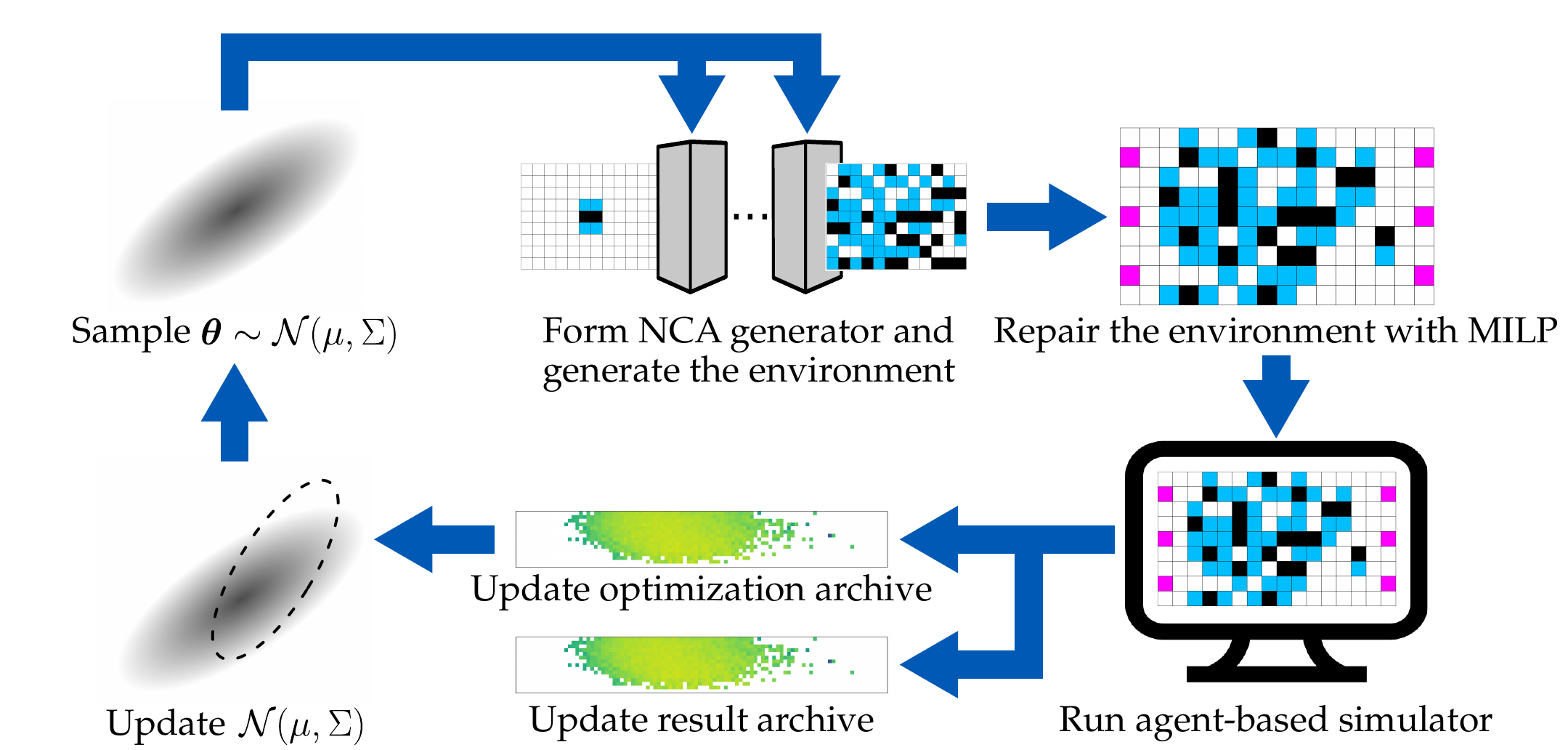}
    \caption{Overview of our method of using CMA-MAE to train diverse NCA generators.}
    \label{fig:cma_mae_nca_algo}
\end{figure}

We extend previous works~\cite{Earle2021IlluminatingDN,zhangLayout23} to use CMA-MAE to search for a diverse collection of NCA generators with the objective and diversity measures computed from an agent-based simulator that runs domain-specific simulations in generated environments. \Cref{fig:cma_mae_nca_algo} provides an overview of our method. We start by sampling a batch of $b$ parameter vectors $\boldsymbol{\theta}$ from a multivariate Gaussian distribution, which form $b$ NCA generators. Each NCA generator then generates one environment from a fixed initial environment, resulting in $b$ environments.
We then repair the environments using a MILP solver to enforce domain-specific constraints. After getting the repaired environments, we evaluate each of them by running an agent-based simulator for $N_e$ times, each with $T$ timesteps, and compute the average objective and measures. We add the evaluated environments and their corresponding NCA generators to both the optimization archive and the result archive. Finally, we update the parameters of the multivariate Gaussian distribution (i.e., $\mu$ and $\Sigma$) and sample a new batch of parameter vectors, starting a new iteration. We run CMA-MAE iteratively with batch size $b$, until the total number of evaluations reaches $N_{eval}$.

\noindent \textbf{NCA Generator.} We define an NCA generator as a function $\mathbf{g}(\mathbf{s};\boldsymbol{\theta},C): \mathbf{X} \rightarrow \mathbf{X}$, where $\mathbf{X}$ is the space of all possible environments, $\mathbf{s} \in \mathbf{X}$ is a fixed initial environment, and $C \in \mathbb{Z}^+$ is a fixed number of iterations for the NCA generator. $\mathbf{g}$ is parameterized by $\boldsymbol{\theta} \in \boldsymbol{\Theta}$.
Since both $\mathbf{s}$ and $C$ are fixed, 
each parameter vector $\boldsymbol{\theta} \in \boldsymbol{\Theta}$ corresponds to an NCA generator, which corresponds to an environment $\mathbf{x} \in \mathbf{X}$.
We discretize the measure space into $M$ cells to create an archive and attempt to find the best NCA generator in each cell to optimize the QD-score of the archive.

Our NCA generator is a convolutional neural network (CNN) with the same architecture as the previous work~\cite{Earle2021IlluminatingDN} with 3 convolutional layers of kernel size 3 $\times$ 3 followed by ReLU or sigmoid activations. \Cref{fig:nca_model_arch} in \Cref{apen:nca_gen:nca_arch} shows the architecture. 
We represent the 2D tile-based environments as 3D tensors, with each tile as a one-hot encoded vector. Depending on the tile types, our NCA model has about 1,500 $\sim$ 3,000 parameters. The input and output of the generator are one-hot environments of the same size, facilitating iterative environment generation by feeding the generator's output back into it. By using a CNN, we can use the same NCA generator to generate environments of arbitrary sizes.

\noindent \textbf{MILP Repair.} The environments generated by the NCA generators might be invalid. Therefore, we follow previous works~\cite{zhang:aiide2020,fontaine2021importance,zhangLayout23} to repair the invalid environments using a MILP solver. For each unrepaired environment $\mathbf{x}_{in} = \mathbf{g}(\mathbf{s}; \boldsymbol{\theta},C)$, we find $\mathbf{x}_{out} \in \mathbf{X}$ such that (1) the hamming distance between $\mathbf{x}_{in}$ and $\mathbf{x}_{out}$ is minimized, and (2) the domain-specific constraints are satisfied. We introduce the domain-specific constraints in detail in \Cref{sec:domain} and \Cref{appen:domain}.

\noindent \textbf{Objective Functions.} We have two objective functions $f_{opt}$ and $f_{res}$ for the optimization archive and the result archive, respectively. $f_{res}$ runs an agent-based simulator for $N_e$ times and returns the average throughput. $f_{opt}$ optimizes the weighted sum of $f_{res}$ and a similarity score $\Delta$ between the unrepaired and repaired environments $\mathbf{x}_{in}$ and $\mathbf{x}_{out}$. Specifically, $f_{opt} = f_{res} + \alpha \cdot \Delta$, where $\alpha$ is a hyperparameter that controls the weight between $f_{res}$ and $\Delta$. We incorporate the similarity score to $f_{opt}$ so that the NCA generators are more inclined to directly generate valid environments with desired regularized patterns, reducing reliance on the MILP solver for pattern generation. We show in \Cref{sec:result} that incorporating the similarity score to $f_{opt}$ significantly improves the scalability of the environments generated by the NCA generators. 

Let $n$ be the number of tiles in the environment, the similarity score $\Delta$ is computed as:

\begin{equation}
    \Delta(\mathbf{x}_{in}, \mathbf{x}_{out}) = \frac{\sum_{i=1}^{n} e_i p_i }{P} ,\text{ where } e_i = 
    \begin{cases}
        1 & \text{if } (\mathbf{x}_{in})_{i} = (\mathbf{x}_{out})_{i}\\
        0 & \text{otherwise.}
    \end{cases}
    \text{ and } P = n \cdot \max_{i} p_i
\end{equation}

$e_i$ encodes if the tile type of tile $i$ in $\mathbf{x}_{in}$ and $\mathbf{x}_{out}$ are the same. $p_i$ assigns a weight to each tile. As a result, the numerator computes the unnormalized similarity score. $P$ is the theoretical upper bound of the unnormalized similarity score of all environments of a domain. Dividing by $P$ normalizes the score to the range of 0 to 1. Intuitively, the similarity score quantifies the level of similarity between the unrepaired and repaired environments $\mathbf{x}_{in}$ and $\mathbf{x}_{out}$, with each tile weighted according to $p_i$.

\noindent \textbf{Environment Entropy Measure.} We want the NCA generators to generate environments with regularized patterns. To quantify these patterns, we introduce environment entropy as a diversity measure, inspired by previous work~\cite{simon2019tilepattern,fontaine2020illuminating}. %
We define a \emph{tile pattern} as one possible arrangement of a 2 $\times$ 2 grid in the environment. Then we count the occurrence of all possible tile patterns in the given environment, forming a tile pattern distribution. The \emph{environment entropy} is calculated as the entropy of this distribution. In essence, the environment entropy measures the degree of regularization of patterns within a given environment. A lower value of environment entropy indicates a higher degree of pattern regularization. We use environment entropy as a diversity measure to find NCA generators that can generate a broad spectrum of environments of varying patterns.

\section{Domains} \label{sec:domain}

We perform experiments in three different domains: (1) a multi-agent warehouse domain~\cite{Li2020LifelongMP,zhangLayout23}, (2) a multi-agent manufacturing domain, and (3) a single-agent maze domain~\cite{fontaine2020illuminating,gym_minigrid,Bhatt2022DeepSA}.

\noindent \textbf{Warehouse.} The warehouse domain simulates automated warehousing scenarios, where robots transport goods between shelves in the middle and human-operated workstations on the left and right borders. We use two variants of the warehouse domain, namely \textit{warehouse (even)} and \textit{warehouse (uneven)}. In warehouse (even), all workstations are visited with equal frequency, while in warehouse (uneven), the workstations on the left border are visited 5 times more frequently than those on the right border. We constrain the warehouse environments such that all non-shelf tiles are connected and the number of shelves is $N_s$ to keep consistent storage capability. We use $f_{opt}$ and $f_{res}$ introduced in \Cref{sec:method} as the objective functions, and environment entropy as one of the measures. Inspired by previous work~\cite{zhangLayout23}, we use the number of connected shelf components as the other measure. \Cref{fig:front-fig} and \Cref{fig:warehouse_exp} in \Cref{appen:domain} show example warehouse environments.

\noindent \textbf{Manufacturing.} The manufacturing domain simulates automated manufacturing scenarios, where robots go to different kinds of manufacturing workstations in a fixed order to manufacture products. We create this domain to show that our method can generate environments with more tile types for more complex robot tasks. We assume 3 kinds of workstations to which the robots must visit in a fixed order and stay for different time duration. We constrain the manufacturing environments such that all non-workstation tiles are connected and there is at least one workstation of each type. We use $f_{opt}$ and $f_{res}$ introduced in \Cref{sec:method} as the objective functions, and the environment entropy and the number of workstations as the measures. The number of workstations can approximate the cost of the manufacturing environment. By diversifying the number of workstations, our method generates a collection of environments with different price points. \Cref{fig:manufacture-exp-main} and \Cref{fig:manufacture_exp} in \Cref{appen:domain} show example manufacturing environments.

\noindent \textbf{Maze.} We conduct additional experiments in the maze domain to show that our method can scale a single-agent reinforcement learning (RL) policy trained in small environments to larger environments with similar patterns. The maze domain does not have domain-specific constraints.  We use the same objective and measure functions in the previous work~\cite{Bhatt2022DeepSA}. Specifically, $f_{opt}$ and $f_{res}$ are both binary functions that return 1 if the environment is solvable and 0 otherwise. The measures are the number of walls and the average path length of the agent over $N_e$ simulations. \Cref{fig:maze-envs-pattern} and \Cref{fig:maze_exp} in \Cref{appen:domain} show example maze environments.

We include more detailed information on the domains in \Cref{appen:domain}.

\section{Experimental Evaluation} \label{sec:result}

\begin{table}[!t]
    \centering
    \begin{center}
    \resizebox{1\linewidth}{!}{
        \begin{tabular}{c|c|c|c|c|c|c|c|c|c}
        \toprule
        Domain & $S$ & $S_{eval}$ & $N_s$ & $N_{s\_eval}$ & $N_a$ & $N_{a\_eval}$ & $N_e$ & $b$ & $N_{eval}$\\
        \hline
        \rule{0pt}{2ex}Warehouse (even) & $36 \times 33$ & $101 \times 102$ & $240$ & $2,250$ & $200$ & $1,400$ & $5$ & $50$ & $10,000$ \\
        Warehouse (uneven) &  $36 \times 33$ & $101 \times 102$ & $240$ & $2,250$ & $200$ & $1,000$ & $5$ & $50$ & $10,000$ \\
        Manufacturing & $36 \times 33$ & $101 \times 102$ & N/A & N/A & $200$ & $1,800$ & $5$ & $50$ & $10,000$ \\
        Maze & $18 \times 18$ & $66 \times 66$ & N/A & N/A & $1$ & $1$ &  $50$ & $150$ & $100,000$\\
        \bottomrule
        \end{tabular}
    }
    \end{center}
    \caption{Summary of the experiment setup. Columns 2-7 show the configurations related to the environment, and columns 8-10 show the parameters of CMA-MAE. $S$, $N_s$, and $N_a$ are the size of the environments, the number of shelves in the warehouse domain, and the number of agents used in training, respectively.  $S_{eval}$, $N_{s\_eval}$, and $N_{a\_eval}$ are their counterparts used in evaluation. %
    We choose $N_{a\_eval}$ to be large enough such that the human-designed environments of the same size are congested while our NCA-generated ones are not. Therefore, the value of $N_{a\_eval}$ depends on the tasks of the robots in each domain.
    }
    \label{tab:search-space}
\end{table}

\Cref{tab:search-space} summarizes the experiment setup. We train the NCA generators with environments of size $S$ and then evaluate them in sizes of both $S$ and $S_{eval}$. In addition, we set the number of NCA iterations $C = 50$ for environments of size $S$ and $C_{eval} = 200$ for those of size $S_{eval}$ for all domains. $C_{eval}$ is larger than $C$ because the NCA generators need more iterations to converge to a stable state while generating environments of size $S_{eval}$. For the warehouse and manufacturing domains, we use Rolling-Horizon Collision Resolution (RHCR)~\cite{Li2020LifelongMP}, a state-of-the-art centralized lifelong MAPF planner, in the simulation. We run each simulation for $T = 1,000$ timesteps during training and $T_{eval} = 5,000$ timesteps during evaluation and stop early in case of \textit{congestion}, which happens if more than half of the agents take wait actions at the same timestep. For the maze domain, we use a trained ACCEL agent~\cite{Dennis2020EmergentCA} in the simulations. We include more details of the experiment setup, compute resources, and implementation in \Cref{appen:exp_setup}.

\subsection{Multi-Agent Domains}

\noindent \textbf{Baseline Environments.} We consider two types of baseline environments for the multi-agent domains (warehouse and manufacturing), namely those designed by human and those optimized by DSAGE~\cite{zhangLayout23}, the state-of-the-art environment optimization method. In DSAGE, we use $f_{res}$ as the objective function for the result archive and $f_{opt}$ with $\alpha=5$ for the surrogate archive. With size $S_{eval}$, however, it is impractical to run DSAGE due to the computational limit. One single run of MILP repair, for example, could take 8 hours in an environment of size $S_{eval}$. We use the human-designed warehouse environments from previous work~\cite{LiuAAMAS19,Li2020LifelongMP,zhangLayout23}. The manufacturing domain is new, so we create human-designed environments for it.

\noindent \textbf{NCA-generated Environments.} We take the best NCA generator from the result archive to generate environments of size $S$ and $S_{eval}$. In the manufacturing domain, we additionally take the best NCA generator that generate an environment of size $S$ with a similar number of workstations as the DSAGE-optimized environment. This is to ensure a fair comparison, as we do not constrain the number of workstations, which is correlated with throughput, in the manufacturing domain.

We show more details of the baselines and NCA-generated environments in \Cref{appen:env}.

\subsubsection{Environments of Size \texorpdfstring{$S$}{S} } \label{subsec:large-compare}

\begin{table}[!t]
    \centering
    \begin{center}
    \resizebox{\linewidth}{!}{
        \begin{tabular}{c|c||c|c||c|c}
        \toprule
                                         &      & \multicolumn{2}{c||}{Size $S$ with $N_a$ agents} & \multicolumn{2}{c}{Size $S_{eval}$ with $N_{a\_eval}$ agents}\\
        \hline
        Domain                           & Algorithm                 & Success Rate                & Throughput                         & Success Rate                & Throughput                      \\
        \hline
        \multirow{5}{*}{warehouse (even)} & CMA-MAE + NCA ($\alpha=0$) & \textbf{100\%} & \textbf{6.79 $\pm$ 0.00} & 0\%  & N/A \\
                                         & CMA-MAE + NCA ($\alpha=1$) & \textbf{100\%} & 6.73 $\pm$ 0.00 & 0\%  & N/A \\
                                         & CMA-MAE + NCA ($\alpha=5$) & \textbf{100\%} & 6.74 $\pm$ 0.00 & \textbf{90\%} & \textbf{16.01 $\pm$ 0.00} \\
                                         & DSAGE ($\alpha=5$)   & \textbf{100\%} & 6.35 $\pm$ 0.00 & N/A  & N/A\\
                                         & Human                & 0\%            & N/A               & 0\%  & N/A\\
        \hline
        \multirow{5}{*}{warehouse (uneven)} & CMA-MAE + NCA ($\alpha=0$) & \textbf{100\%} & \textbf{6.89 $\pm$ 0.00} & 62\% & \textbf{12.32 $\pm$ 0.00} \\
                                           & CMA-MAE + NCA ($\alpha=1$) & \textbf{100\%} & 6.70 $\pm$ 0.00 & 8\%  & 11.56 $\pm$ 0.01\\
                                           & CMA-MAE + NCA ($\alpha=5$) & \textbf{100\%} & 6.82 $\pm$ 0.00 & \textbf{84\%} & 12.03 $\pm$ 0.00\\
                                           & DSAGE ($\alpha=5$)   & \textbf{100\%} & 6.40 $\pm$ 0.00 & N/A  & N/A\\
                                           & Human                & 0\%            & N/A             & 0\%  & N/A\\
        \hline
        \multirow{4}{*}{Manufacturing}       & CMA-MAE + NCA ($\alpha=5$, opt)  & 94\%  & \textbf{6.82 $\pm$ 0.00} & \textbf{100\%} & \textbf{23.11 $\pm$ 0.01}\\
                                           & CMA-MAE + NCA ($\alpha=5$, comp DSAGE) & 98\%  & 6.61 $\pm$ 0.00 & N/A  & N/A \\
                                           & DSAGE ($\alpha=5$)   & 28\%  & 5.61 $\pm$ 0.12 & N/A  & N/A\\
                                           & Human                & \textbf{100\%} & 5.92 $\pm$ 0.00 & 0\% & N/A \\
        \bottomrule
        \end{tabular}
    }
    \end{center}
    \caption{Success rates and throughput of environments of sizes $S$ and $S_{eval}$. We run 50 simulations for all environments except for the manufacturing environments of size $S_{eval}$, for which we run 20 simulations.
    The success rate is calculated as the percentage of simulations that end without congestion. We measure the throughput of only successful simulations and report both its average and standard error.
}
    \label{tab:numerical-result}
\end{table}

\begin{figure}[!t]
    \centering
    \includegraphics[width=0.9\textwidth]{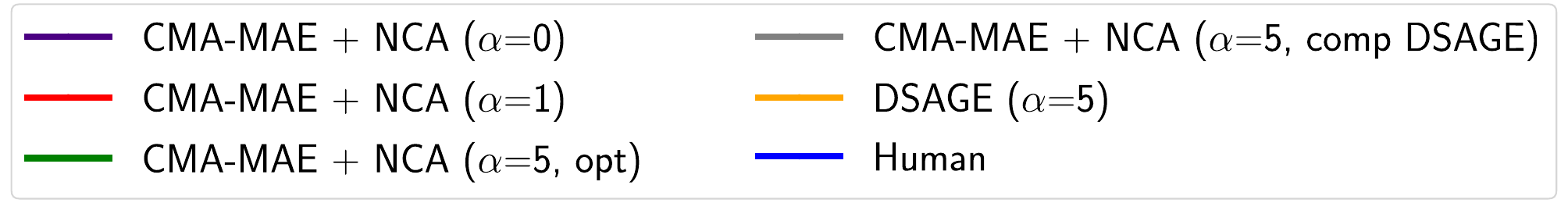}
    \begin{subfigure}[t]{0.3\textwidth}
        \includegraphics[width=1\textwidth]{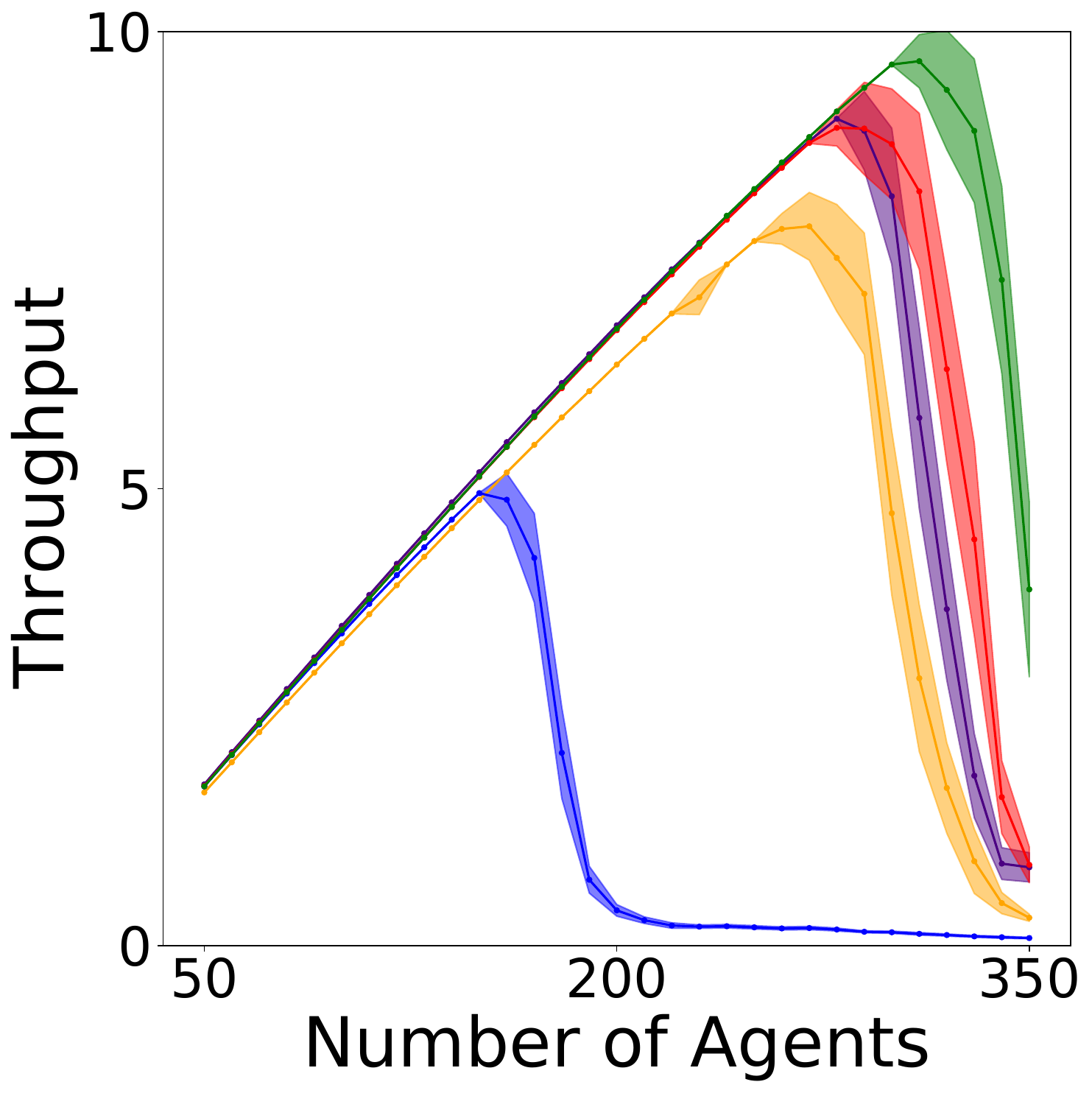}
        \caption{$S$: Warehouse (even)}
        \label{fig:warehouse-large-even-vs-n_agents}
    \end{subfigure}
    \begin{subfigure}[t]{0.3\textwidth}
        \includegraphics[width=1\textwidth]{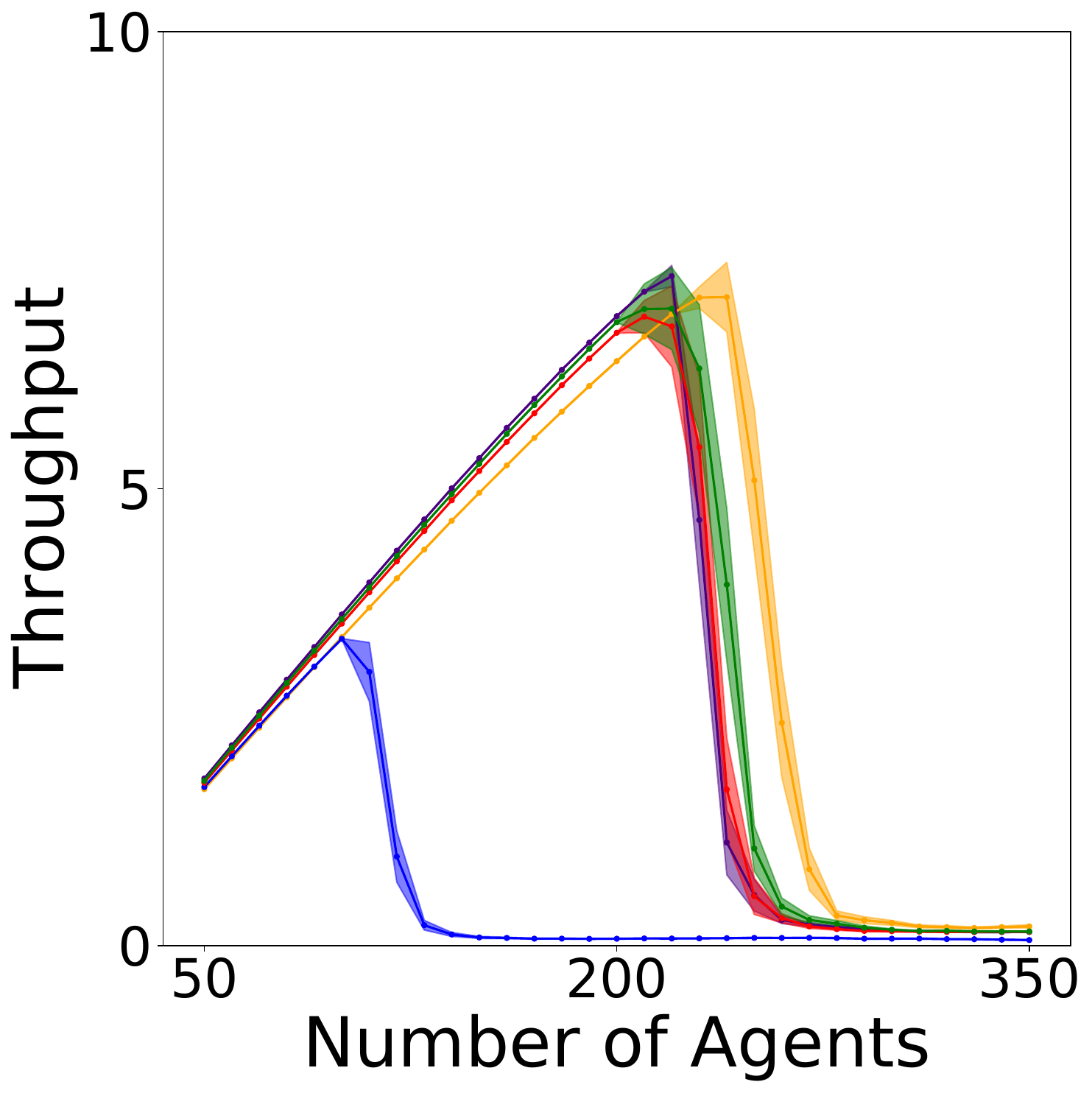}
        \caption{$S$: Warehouse (uneven)}
        \label{fig:warehouse-large-uneven-vs-n_agents}
    \end{subfigure}
    \begin{subfigure}[t]{0.3\textwidth}
        \includegraphics[width=1\textwidth]{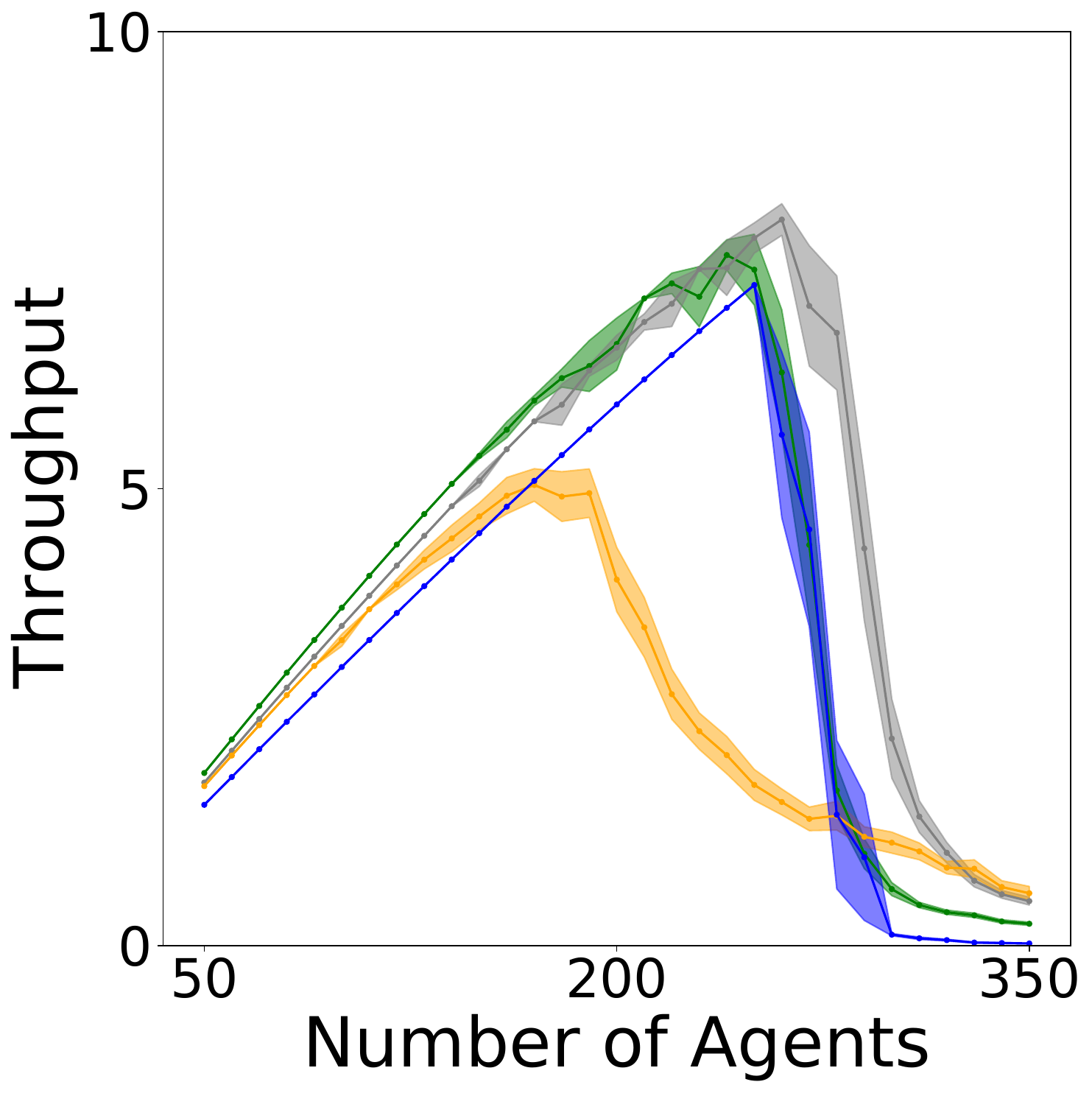}
        \caption{$S$: Manufacturing}
        \label{fig:manufacture-large-vs-n_agents}
    \end{subfigure}\\
    \begin{subfigure}[t]{0.3\textwidth}
        \includegraphics[width=1\textwidth]{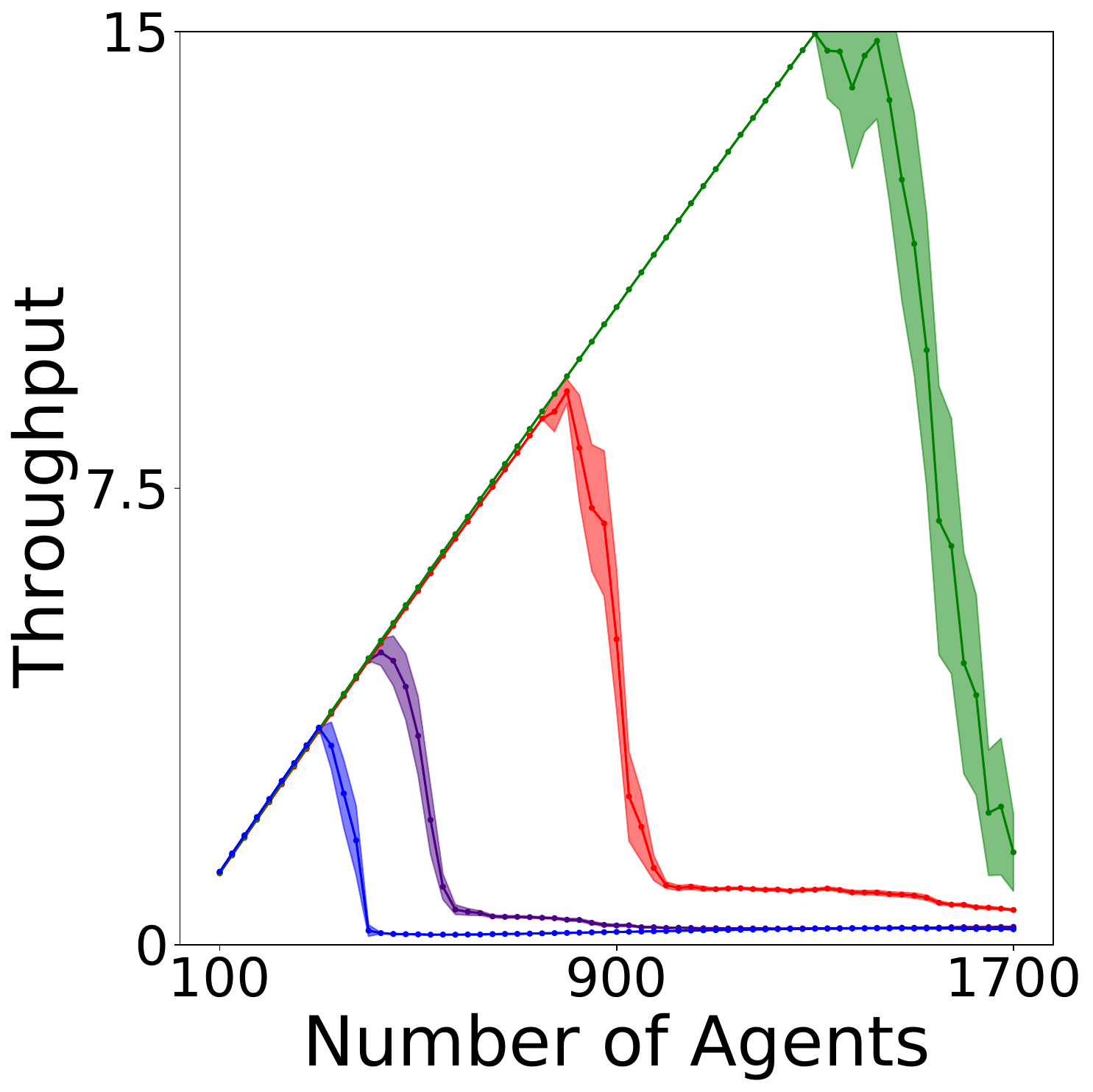}
        \caption{$S_{eval}$: Warehouse (even)}
        \label{fig:warehouse-xxlarge-compare-even}
    \end{subfigure}
    \begin{subfigure}[t]{0.3\textwidth}
        \includegraphics[width=1\textwidth]{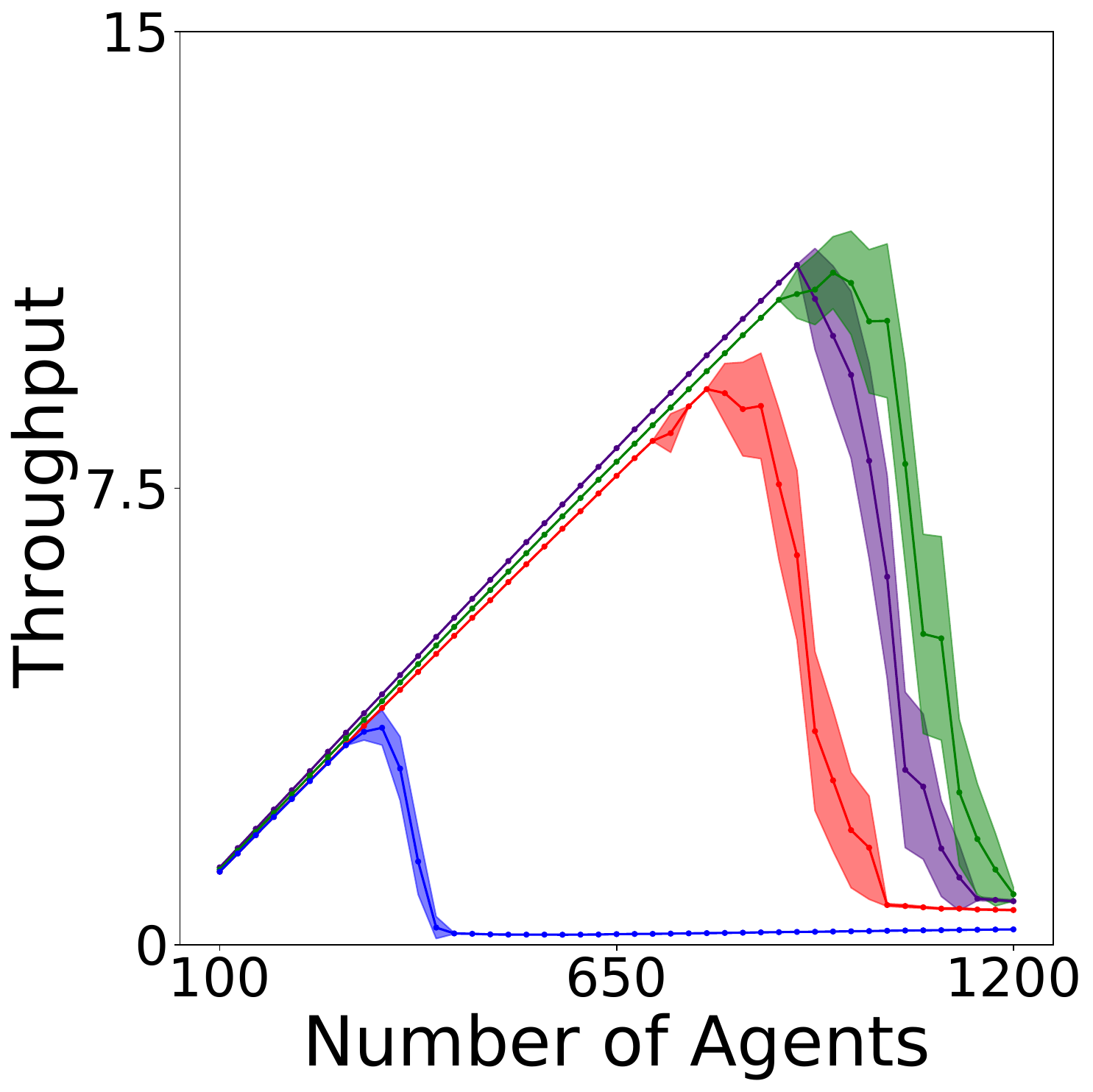}
        \caption{$S_{eval}$: Warehouse (uneven)}
        \label{fig:warehouse-xxlarge-compare-uneven}
    \end{subfigure}
    \begin{subfigure}[t]{0.3\textwidth}
        \includegraphics[width=1\textwidth]{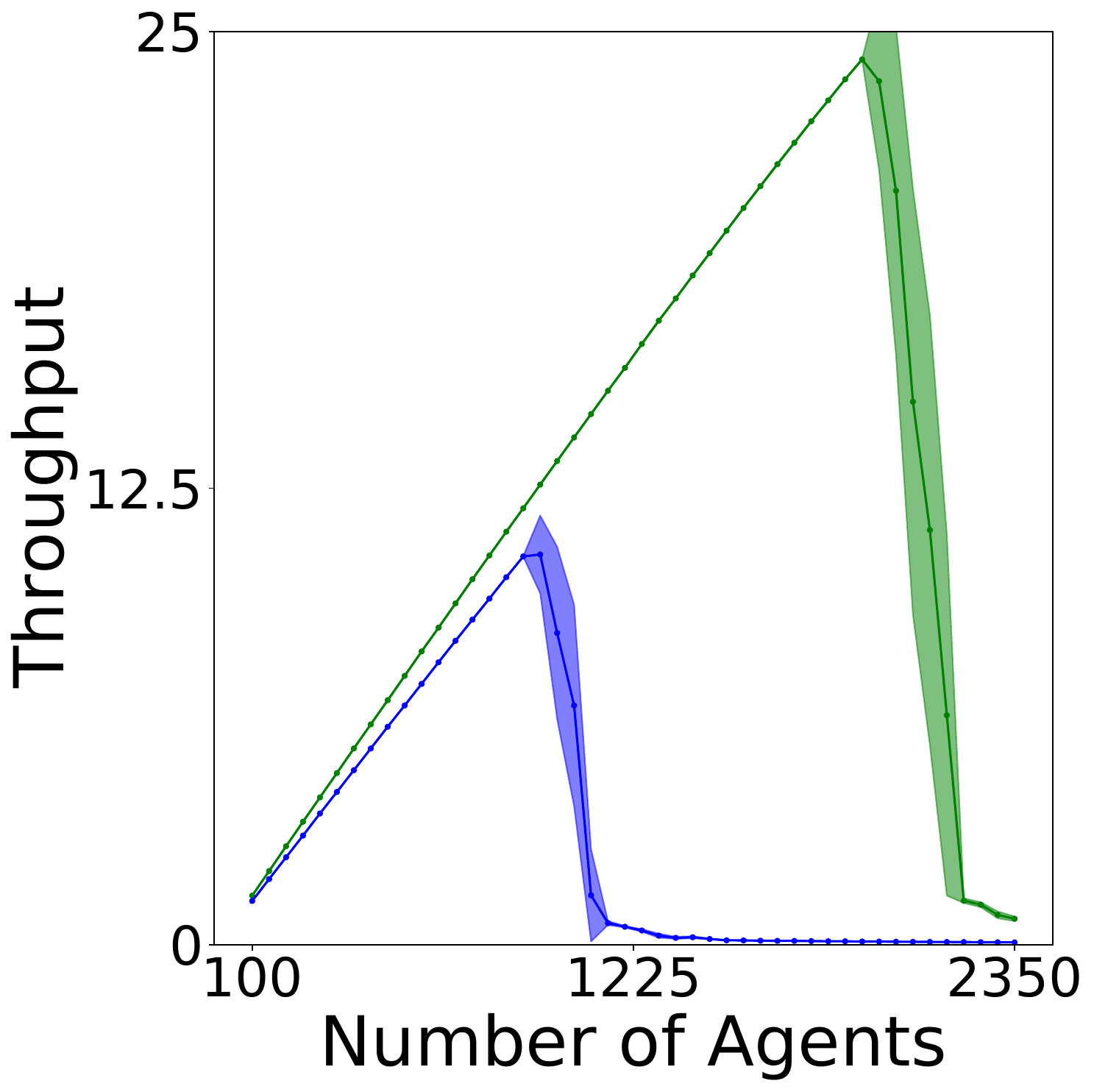}
        \caption{$S_{eval}$: Manufacturing}
        \label{fig:manufacture-xxlarge-compare}
    \end{subfigure}
    \caption{Throughput with an increasing number of agents in environments of size $S$ (a-c) and $S_{eval}$ (d-f). For size $S$, we run 50 simulations and increase the number of agents by a step size of 10. For $S_{eval}$, we run 50 and 20 simulations and increase the number of agents by step sizes of 25 and 50 in warehouse and manufacturing domains, respectively. The solid lines are the average throughput while the shaded area shows the 95\% confidence interval.}
    \label{fig:large-vs-n_agents}
\end{figure}

Columns 3 and 4 of \Cref{tab:numerical-result} show the numerical results with $N_a = 200$ agents in the warehouse and manufacturing domains with environment size $S$. All NCA-generated environments have significantly better throughput than the baseline environments. In addition, we observe that larger $\alpha$ values have no significant impact on the throughput with $N_a$ agents.

We also analyze the scalability of the environments by running simulations with varying numbers of agents and show the throughput in \Cref{fig:warehouse-large-even-vs-n_agents,fig:warehouse-large-uneven-vs-n_agents,fig:manufacture-large-vs-n_agents}. In both the warehouse (even) and manufacturing domains, our NCA-generated environments have higher throughput than the baseline environments with an increasingly larger gap with more agents. In the warehouse (uneven) domain, however, the DSAGE-optimized environment displays slightly better scalability than those generated by NCA. This could be due to a need for more traversable tiles near the popular left-side workstations to alleviate congestion, a pattern creation task that poses a significant challenge for NCA generators compared to DSAGE's direct tile search. In addition, in the manufacturing domain, the optimal NCA generator (green line) creates an environment with top throughput for 200 agents, but it is slightly less scalable than the sub-optimal one used for DSAGE comparison (grey line). We conjecture this is due to the optimizer being caught in a local optimum region due to the complexity of the tasks.
On the effect of $\alpha$ values, we observe that larger $\alpha$ values enhance the scalability of the generated environments in the warehouse (even) domain, as shown in \Cref{fig:warehouse-large-even-vs-n_agents}. We include further discussion of $\alpha$ values in \Cref{appen:alpha-val}.

To further understand the gap of throughput among different environments of size $S$ with $N_a$ agents, we present additional results on the number of finished tasks over time in \Cref{appen:add-result:task_thr_time}. All NCA-generated environments maintain a stable number of finished tasks throughout the simulation.

\subsubsection{Environments of Size \texorpdfstring{$S_{eval}$}{S\_eval}}

Columns 5 and 6 of \Cref{tab:numerical-result} compare environments of size $S_{eval}$ with $N_{a\_eval}$ agents. In the warehouse (even) domain, all environments, except for the one generated by CMA-MAE + NCA ($\alpha = 5$), run into congestion.
In the warehouse (uneven) domain, CMA-MAE + NCA ($\alpha = 5$) also achieves the best success rate, even though it does not have the highest throughput. This shows the benefit of using a larger $\alpha$ value with CMA-MAE in creating large and scalable environments. 

To further understand the scalability, \Cref{fig:warehouse-xxlarge-compare-even,fig:warehouse-xxlarge-compare-uneven,fig:manufacture-xxlarge-compare} show the throughput of the environments with an increasing number of agents. We observe that all our NCA-generated environments are much more scalable than the human-designed ones. In particular, the environments generated by CMA-MAE + NCA ($\alpha = 5$) yield the best scalability in all domains. 
This is because a higher value of $\alpha$ makes CMA-MAE focus more on minimizing the disparity between the unrepaired and repaired environments. As a result, the optimized NCA generators are more capable of generating environments with desired regularized patterns directly. On the other hand, a smaller value of $\alpha$ makes the algorithm rely more on the MILP solver to generate patterns, introducing more randomness as the MILP solver arbitrarily selects an optimal solution with minimal hamming distance when multiple solutions exist. 
We show additional results of scaling the same NCA generators to more environment sizes in \Cref{appen:add-result:scale-env-size}. We also include a more competitive baseline by tiling environments of size $S$ to create those of size $S_{eval}$ in \Cref{appen:tile-env-baseline}.

\subsection{Scaling Single-Agent RL Policy}

\begin{figure}
\centering
\begin{minipage}{.325\textwidth}
  \centering
  \includegraphics[width=1\textwidth]{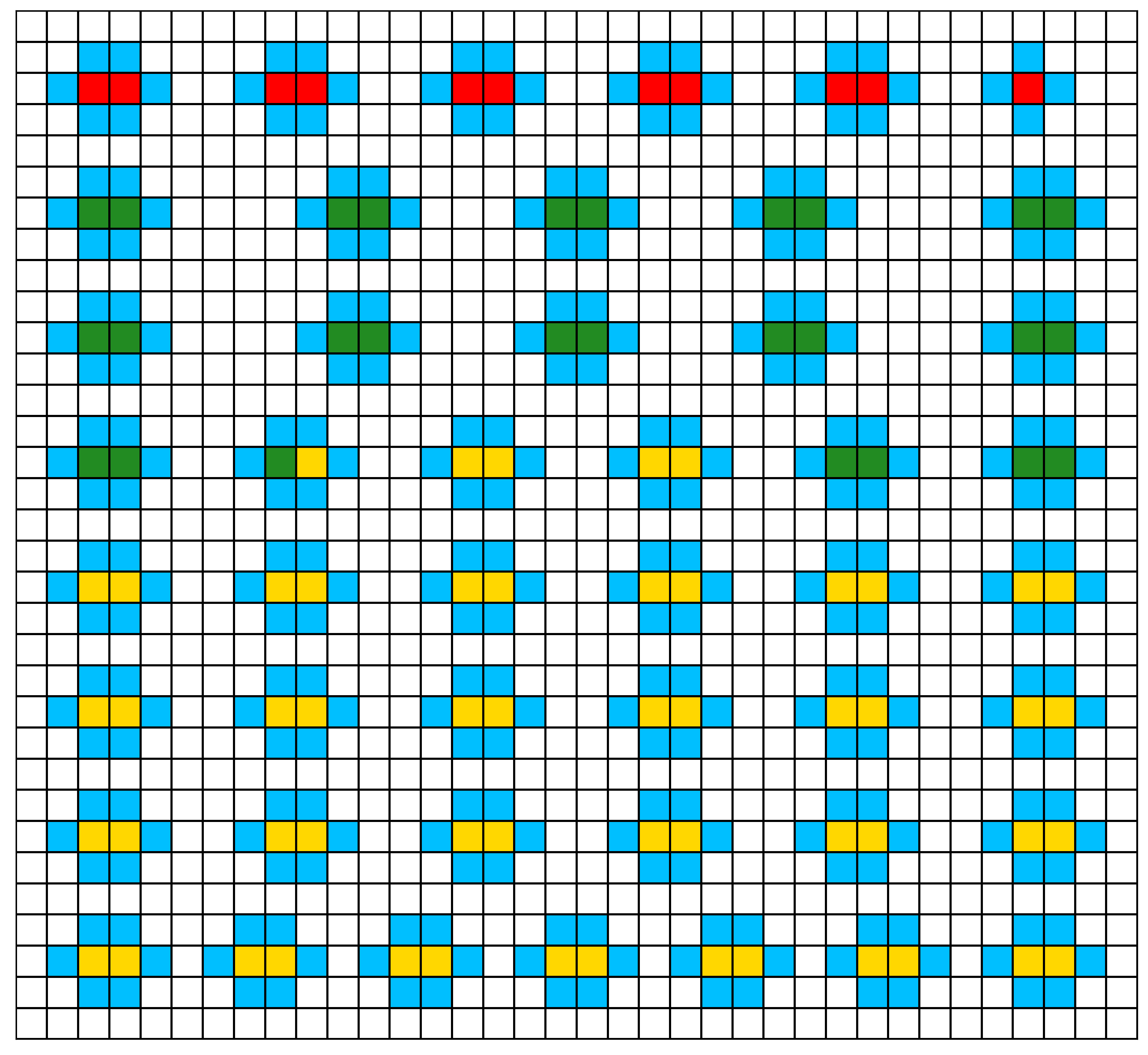}    
    \linebreak[4] %
    \vspace{5pt}
    \caption{Example manufacturing environment.}
      \label{fig:manufacture-exp-main}
\end{minipage}
\hfill
\begin{minipage}{.65\textwidth}
    \centering
    \nextfloat
    \begin{subfigure}[t]{0.47\textwidth}
        \includegraphics[width=1\textwidth]{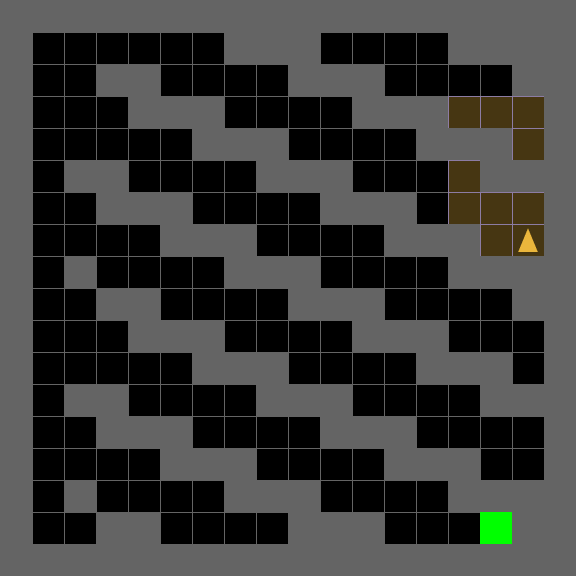}
        \caption{18 $\times$ 18.}
        \label{fig:maze-small}
    \end{subfigure}
    \begin{subfigure}[t]{0.47\textwidth}
        \includegraphics[width=1\textwidth]{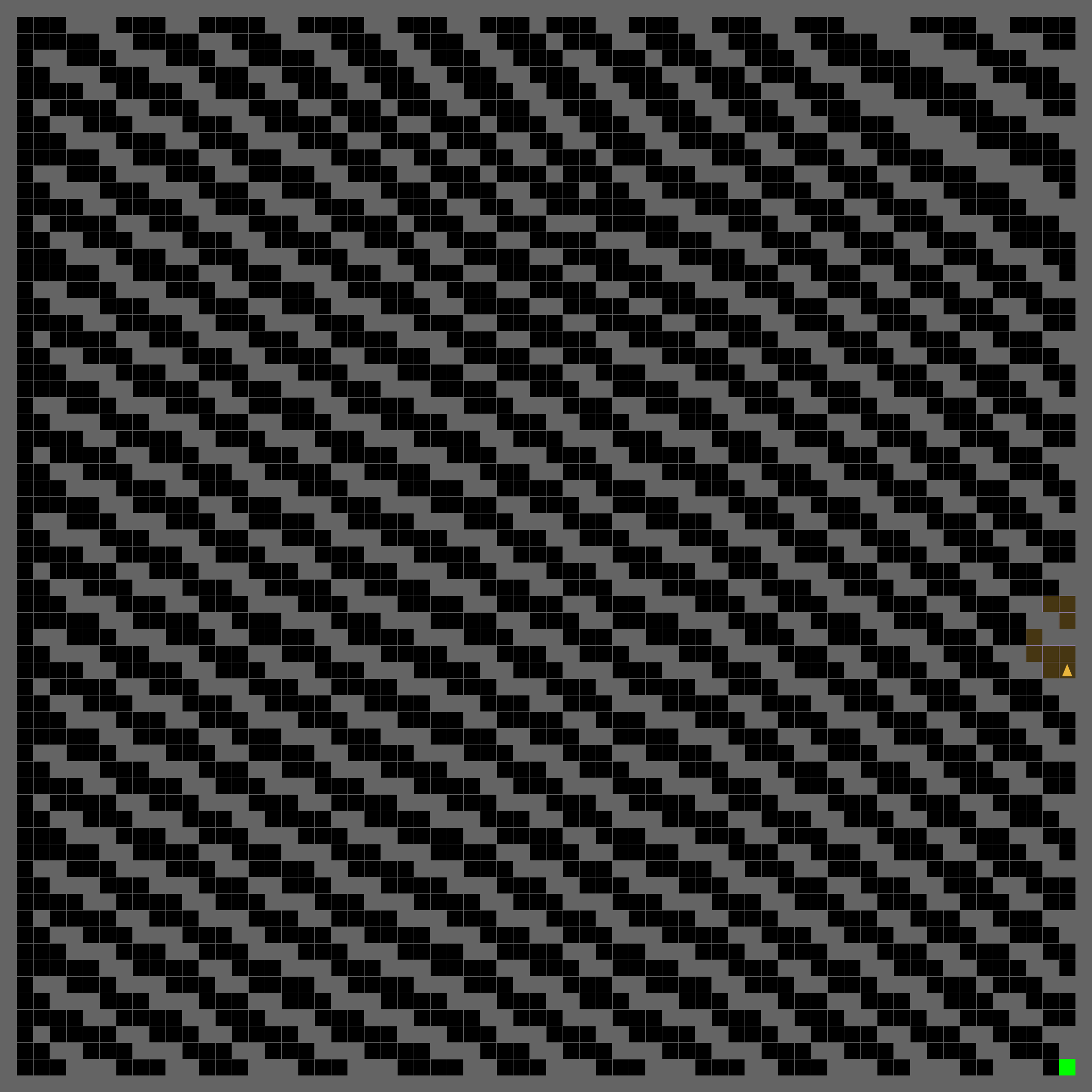}
        \caption{66 $\times$ 66.}
        \label{fig:maze-xxlarge}
    \end{subfigure}
    \caption{Maze environments with similar patterns of different sizes generated by the same NCA generator.}
    \label{fig:maze-envs-pattern}
\end{minipage}
\end{figure}

\begin{figure}[!t]

\end{figure}

In the maze domain, we show that it is possible to scale a single-agent RL policy to larger environments with similar regularized patterns. We use the NCA generator trained from 18 $\times$ 18 maze environments to generate a small 18 $\times$ 18 and a large 66 $\times$ 66 maze environment, shown in \Cref{fig:maze-envs-pattern}. Running the ACCEL~\cite{Dennis2020EmergentCA} agent 100 times in these environments results in success rates of 90\% and 93\%, respectively. The high success rate comes from the fact that the similar local observation space of the RL policy enables the agent to make the correct local decision and eventually arrive at the goal.

For comparison, we generate 100 baseline 66 $\times$ 66 maze environments by restricting the wall counts to be the same and the path lengths to range between 80\% to 120\% of the one in \Cref{fig:maze-xxlarge}, to maintain similar difficulty levels. \Cref{fig:xxlarge-maze-baseline} in \Cref{appen:baseline-env} shows two example baseline environments. Testing the ACCEL agent in these environments yields an average success rate of 22\% (standard error: 3\%), significantly lower than that achieved in the NCA-generated environments. This demonstrates the ability of our method to generate arbitrarily large environments with similar patterns in which the trained RL agent performs well.

\subsection{NCA Generation Time}
\label{appen:nca_gen_time}

\begin{table}[!t]
    \centering
    \begin{center}
    \resizebox{0.95\linewidth}{!}{
        \begin{tabular}{c|c||c|r|c||c|r|c}
        \toprule
                                         &      & \multicolumn{3}{c||}{Size $S$ with $N_a$ agents} & \multicolumn{3}{c}{Size $S_{eval}$ with $N_{a\_eval}$ agents}\\
        \hline
        Domain                           & Algorithm                 & $T_{NCA}$                & $T_{MILP}$       & $\Delta$                  & $T_{NCA}$                & $T_{MILP}$       & $\Delta$                  \\
        \hline
        \multirow{3}{*}{\shortstack{Warehouse (even)}} & CMA-MAE ($\alpha=0$) & 0.06 & 2.72 & 0.94 & 0.12  & 228.80 & 0.95 \\
                                         & CMA-MAE ($\alpha=1$) & 0.06 & 2.97 & 0.86 & 0.13  & 9278.18 & 0.84 \\
                                         & CMA-MAE ($\alpha=5$) & 0.06 & 2.55 & 0.97 & 0.13 & 19.79 & 0.99\\
        \hline
        \multirow{3}{*}{\shortstack{Warehouse (uneven)}} & CMA-MAE ($\alpha=0$) & 0.06 & 2.41 & 0.85 & 0.13 & 1,376.62 & 0.85\\
                                           & CMA-MAE ($\alpha=1$) & 0.06 & 2.36 & 0.95 & 0.13 & 26.54 & 0.96\\
                                           & CMA-MAE ($\alpha=5$) & 0.06 & 13.42 & 0.84 & 0.13 & 15,656.04 & 0.86\\
        \hline
        \multirow{1}{*}{Manufacturing}       & CMA-MAE ($\alpha=5$, opt)  & 0.07  & 3.04 & 0.17 & 0.18 & 15,746.10 & 0.18\\
        \bottomrule
        \end{tabular}
    }
    \end{center}
    \caption{CPU runtime measured (in seconds) for the trained NCA generators to generate environments ($T_{NCA}$), and for the MILP solver to subsequently repair the generated environments ($T_{MILP}$) of size $S$ and $S_{eval}$. 
    $\Delta$ refers to the similarity score between the repaired and unrepaired environments. We measure the CPU runtime in machine (1) listed in \Cref{appen:compute} and compute constraints in \Cref{appen:implementation}.
    }
    \label{tab:numerical-nca-milp-runtime}
\end{table}

We show the total CPU runtime of generating and repairing the environments of sizes $S$ and $S_{eval}$ in \Cref{tab:numerical-nca-milp-runtime}. In both sizes, larger similarity scores are correlated with shorter MILP runtime. The similarity scores of the manufacturing environments are significantly lower because we use a larger normalization factor $P$ than that in the warehouse domains. Nevertheless, we note that even the longest combined runtime of the NCA generation and MILP repair ($15,746.10 + 0.18 = 15,746.28$ seconds for the manufacturing domain) on the environments of size $S_{eval}$ is significantly shorter than optimizing the environments of that size directly. This reduction in time is attributed to our method, which involves running the MILP repair $N_{eval}$ times with an environment size of $S$ while training the NCA generators, and then using the MILP solver only once with size $S_{eval}$ while generating the environments. Our method stands in contrast to the previous direct environment optimization method~\cite{zhangLayout23} which requires running the MILP repair with the size of $S_{eval}$ for $N_{eval}$ times.

\subsection{On the Benefit of QD Algorithms}

QD algorithms can train a diverse collection of NCA generators.
In \Cref{appen:add-result:qd-score-archive-cover}, we compare the QD-score and archive coverage of CMA-MAE + NCA and DSAGE. We also benchmark CMA-MAE against MAP-Elites~\cite{mouret2015illuminating,vassiliades2018isoline} to highlight the benefit of using CMA-MAE in training NCA generators. We observe that CMA-MAE + NCA has the best QD-score and archive coverage in all domains.

Furthermore, QD algorithms can train more scalable NCA generators than single-objective optimizers.
In \Cref{appen:add-result:cma-es} we compare CMA-MAE with a single-objective optimizer CMA-ES~\cite{hansen2016cmaes}. We observe that CMA-MAE is less prone to falling into local optima and thus better NCA generators.

\section{Limitations and Future Work}

We present a method to train a diverse collection of NCA generators that can generate environments of arbitrary sizes within a range defined by finite compute resources. In small environments of size $S$, our method generates environments that have competitive or better throughput compared to the state-of-the-art environment optimization method~\cite{zhangLayout23}. In large environments of size $S_{eval}$, where existing optimization methods are inapplicable, our generated environments achieve significantly higher throughput and better scalability compared to human-designed environments.

Our work is limited in many ways, yielding many directions for future work. First, our method is only arbitrarily scalable under a range determined by finite compute resources. For example, with the compute resource in \Cref{appen:compute,appen:implementation}, the upper bound of the range is about $S_{eval} = 101 \times 102$. Although this upper bound is significantly higher than previous environment optimization methods, future work can seek to further increase it.
Second, we focus on generating 4-neighbor grid-based environments with an underlying undirected movement graph. Future work can consider generating similar environments with directed movement graphs or irregular-shaped environments. 
We additionally discuss social impacts of our work in \Cref{appen:social-impact}.

\section*{Acknowledgements}

This work used Bridge-2 at Pittsburgh Supercomputing Center (PSC) through allocation CIS220115 from the Advanced Cyberinfrastructure Coordination Ecosystem: Services \& Support (ACCESS) program, which is supported by National Science Foundation grants \#2138259, \#2138286, \#2138307, \#2137603, and \#2138296.
In addition, this work was supported by the
CMU Manufacturing Futures Institute, made possible by the
Richard King Mellon Foundation, and was partially supported by the NSF CAREER Award (\#2145077).

\bibliographystyle{named}
\bibliography{neurips_2023}

\newpage
\appendix

\section{NCA Generation Process} \label{appen:nca-process}

\begin{figure}[!t]
    \centering
    \begin{subfigure}{0.19\textwidth}
        \includegraphics[width=1\textwidth]{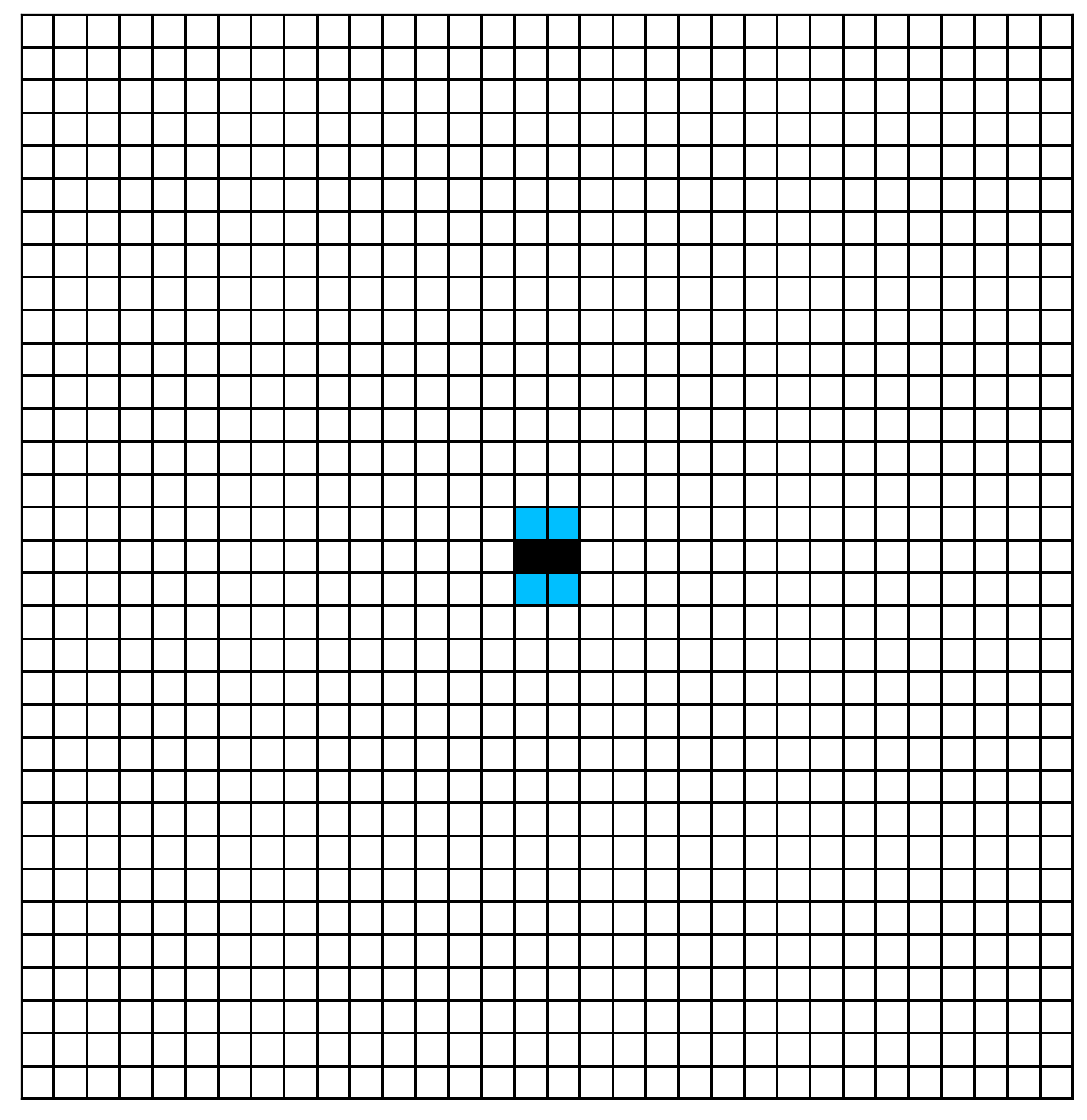}
        \caption{Iteration 0}
        \label{fig:nca-seed}
    \end{subfigure}
    \hfill
    \begin{subfigure}{0.19\textwidth}
        \includegraphics[width=1\textwidth]{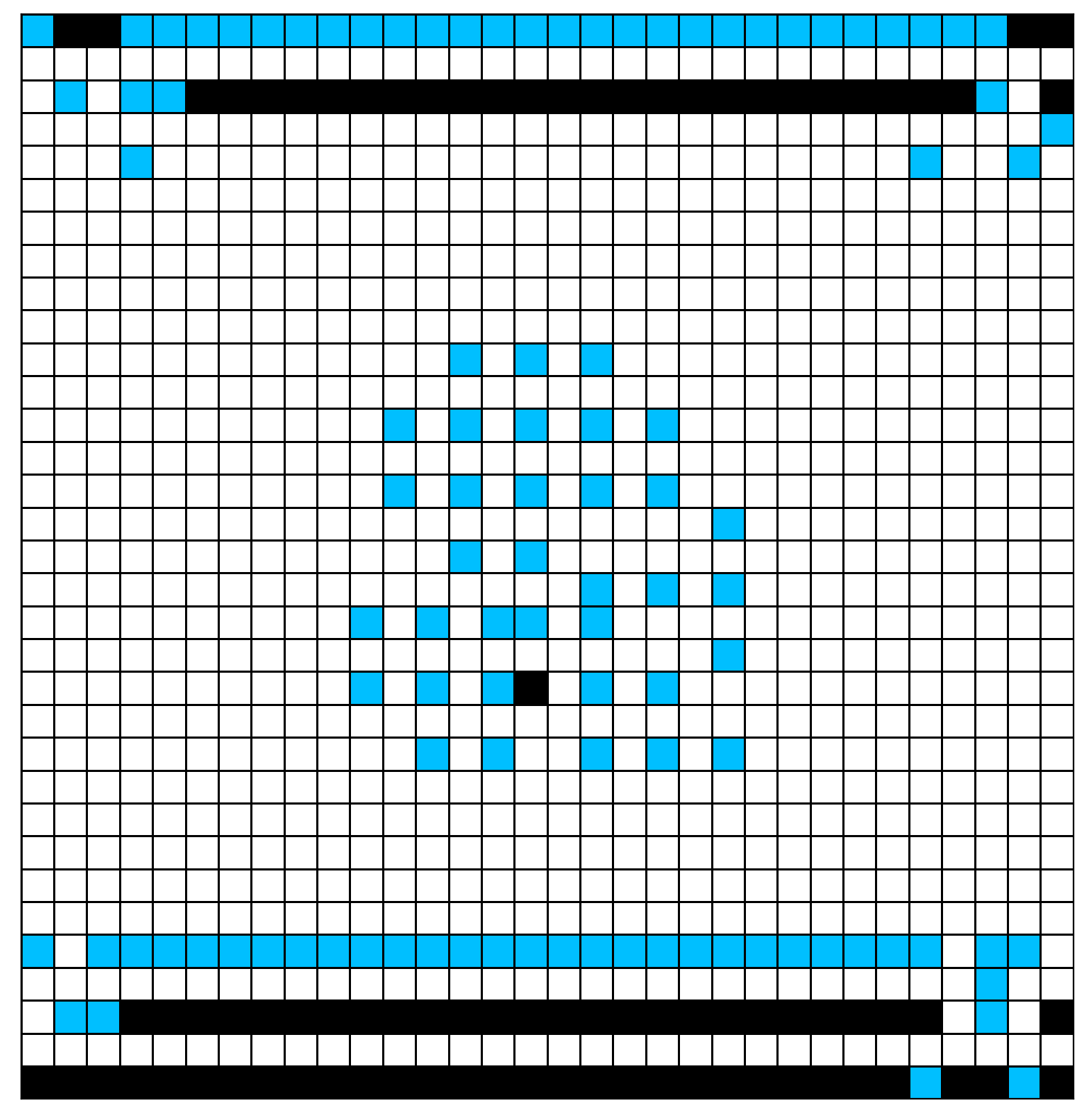}
        \caption{Iteration 5}
        \label{fig:nca-iter5}
    \end{subfigure}
    \hfill
    \begin{subfigure}{0.19\textwidth}
        \includegraphics[width=1\textwidth]{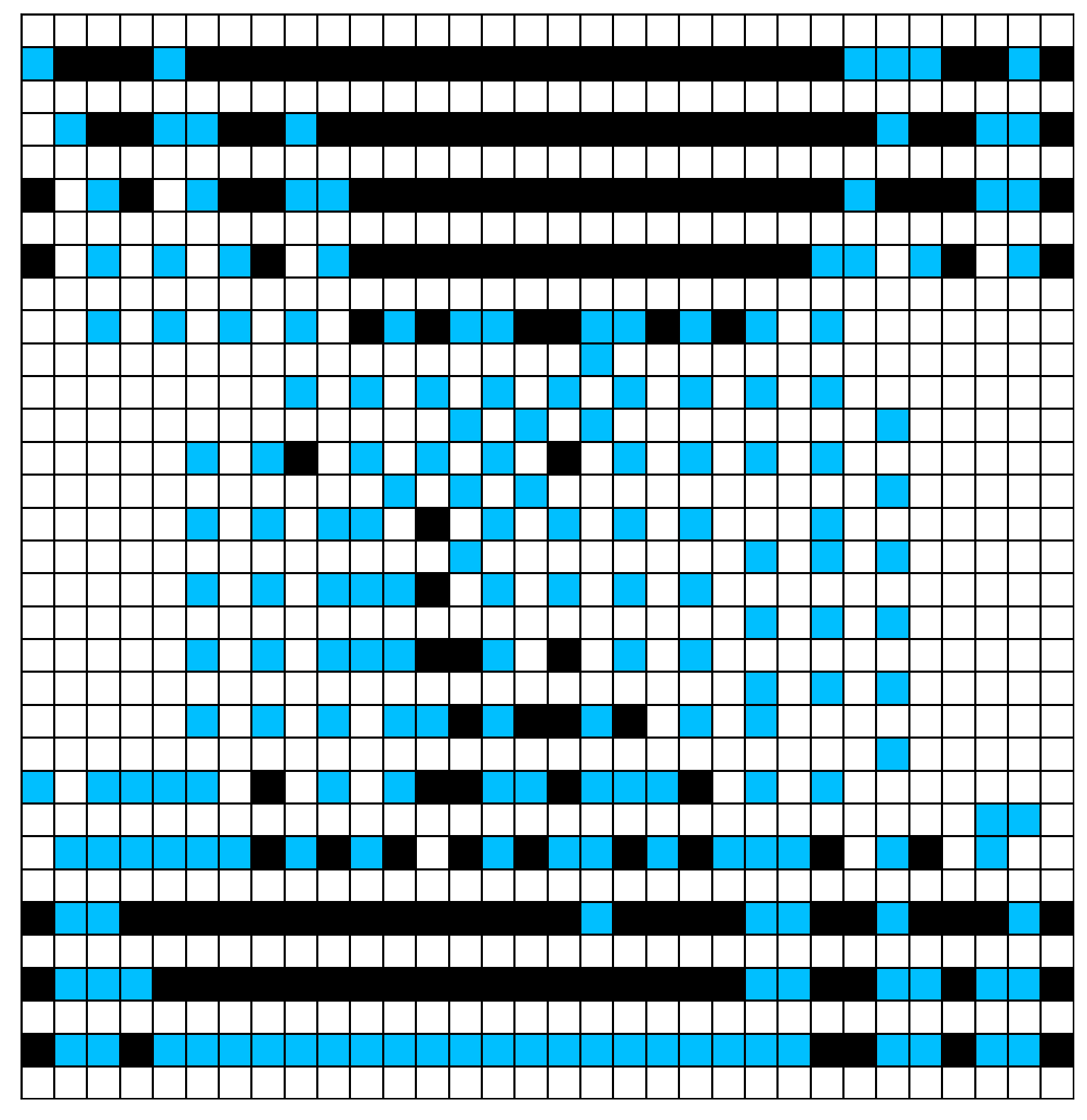}
        \caption{Iteration 10}
        \label{fig:nca-iter10}
    \end{subfigure}
    \hfill
    \begin{subfigure}{0.19\textwidth}
        \includegraphics[width=1\textwidth]{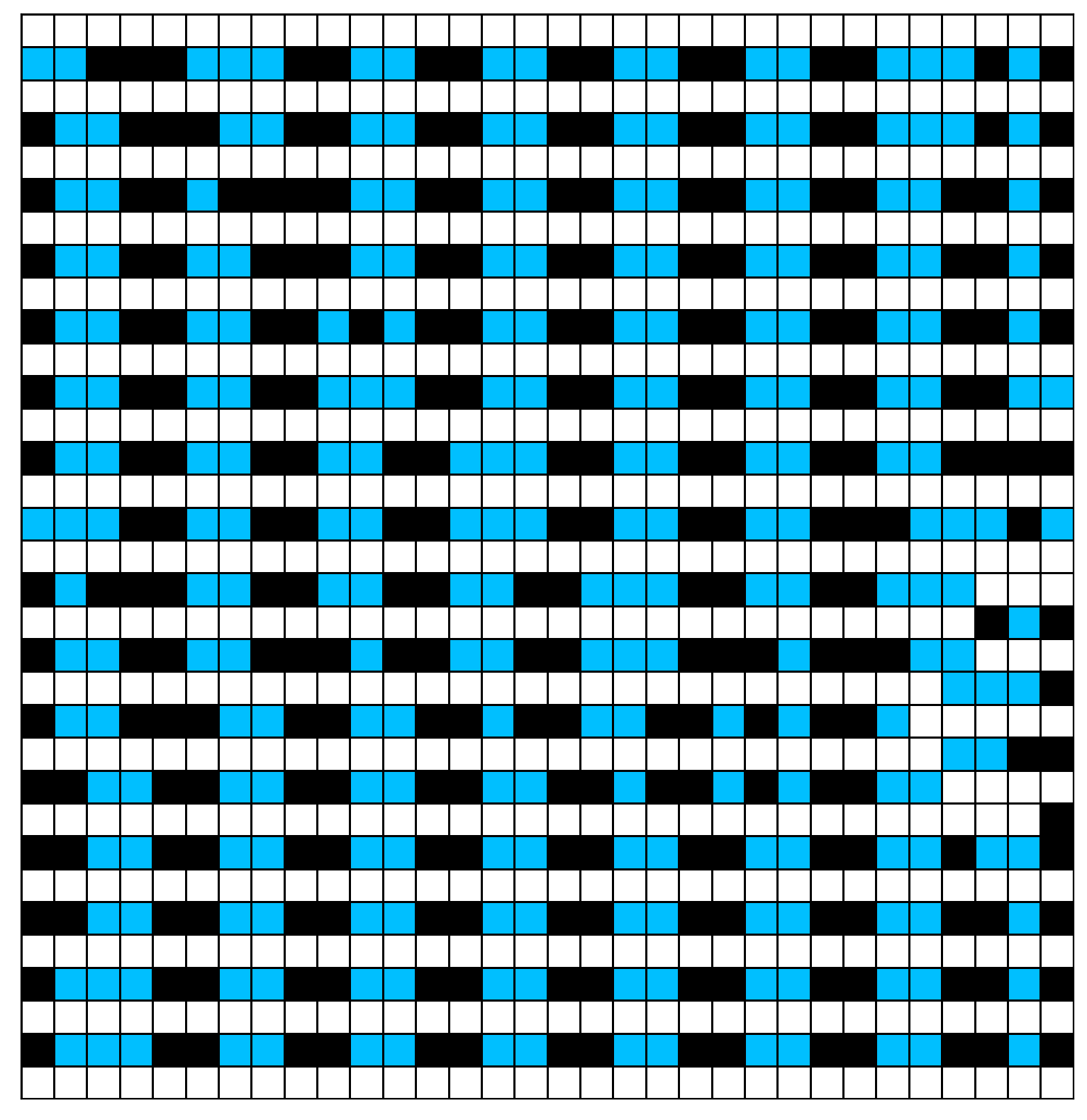}
        \caption{Iteration 30}
        \label{fig:nca-iter30}
    \end{subfigure}
    \hfill
    \begin{subfigure}{0.19\textwidth}
        \includegraphics[width=1\textwidth]{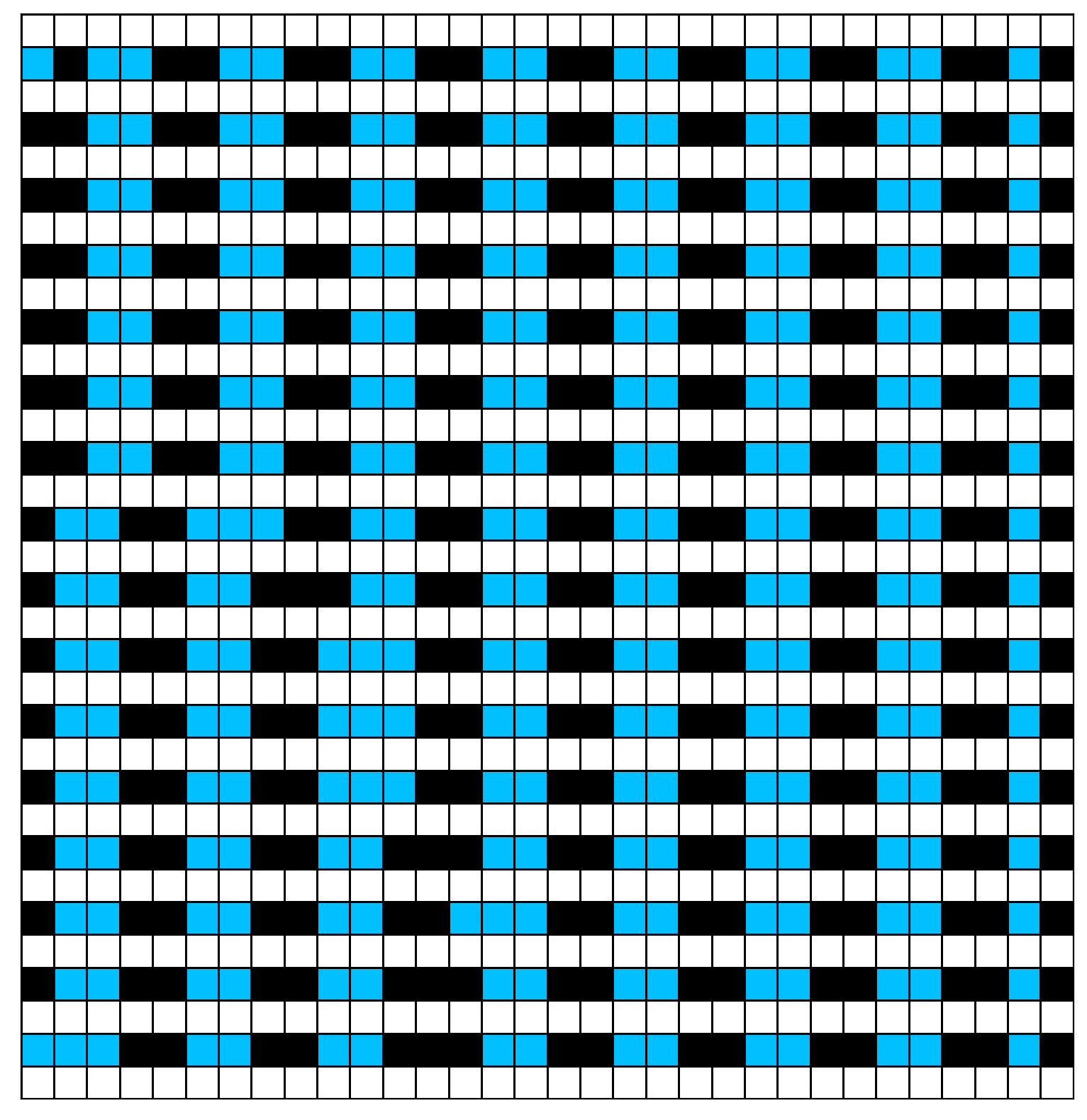}
        \caption{Iteration 50}
        \label{fig:nca-iter50}
    \end{subfigure}
    \caption{Example NCA environment generation process with 50 iterations, which starts from a fixed initial environment (\Cref{fig:nca-seed}) and iteratively generates the rest of the environments (\Cref{fig:nca-iter5,fig:nca-iter10,fig:nca-iter30,fig:nca-iter50}).}
    \label{fig:nca_process}
\end{figure}

\Cref{fig:nca_process} shows an example NCA generation process with 50 iterations.

\begin{figure}[!t]
    \centering
    \begin{subfigure}[t]{0.33\textwidth}
        \includegraphics[width=1\textwidth]{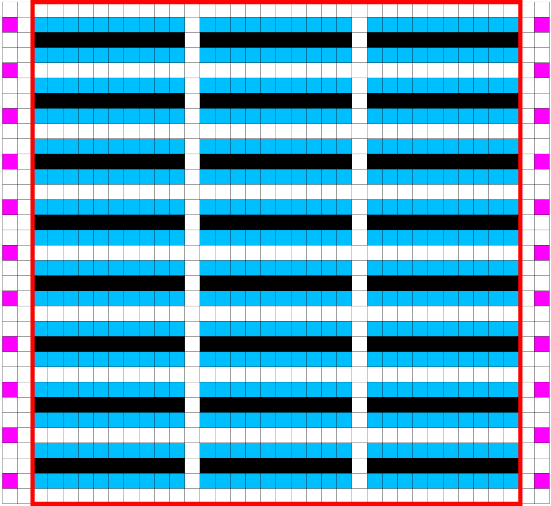}
        \caption{Warehouse}
        \label{fig:warehouse_exp}
    \end{subfigure}
    \hfill
    \begin{subfigure}[t]{0.33\textwidth}
        \includegraphics[width=1\textwidth]{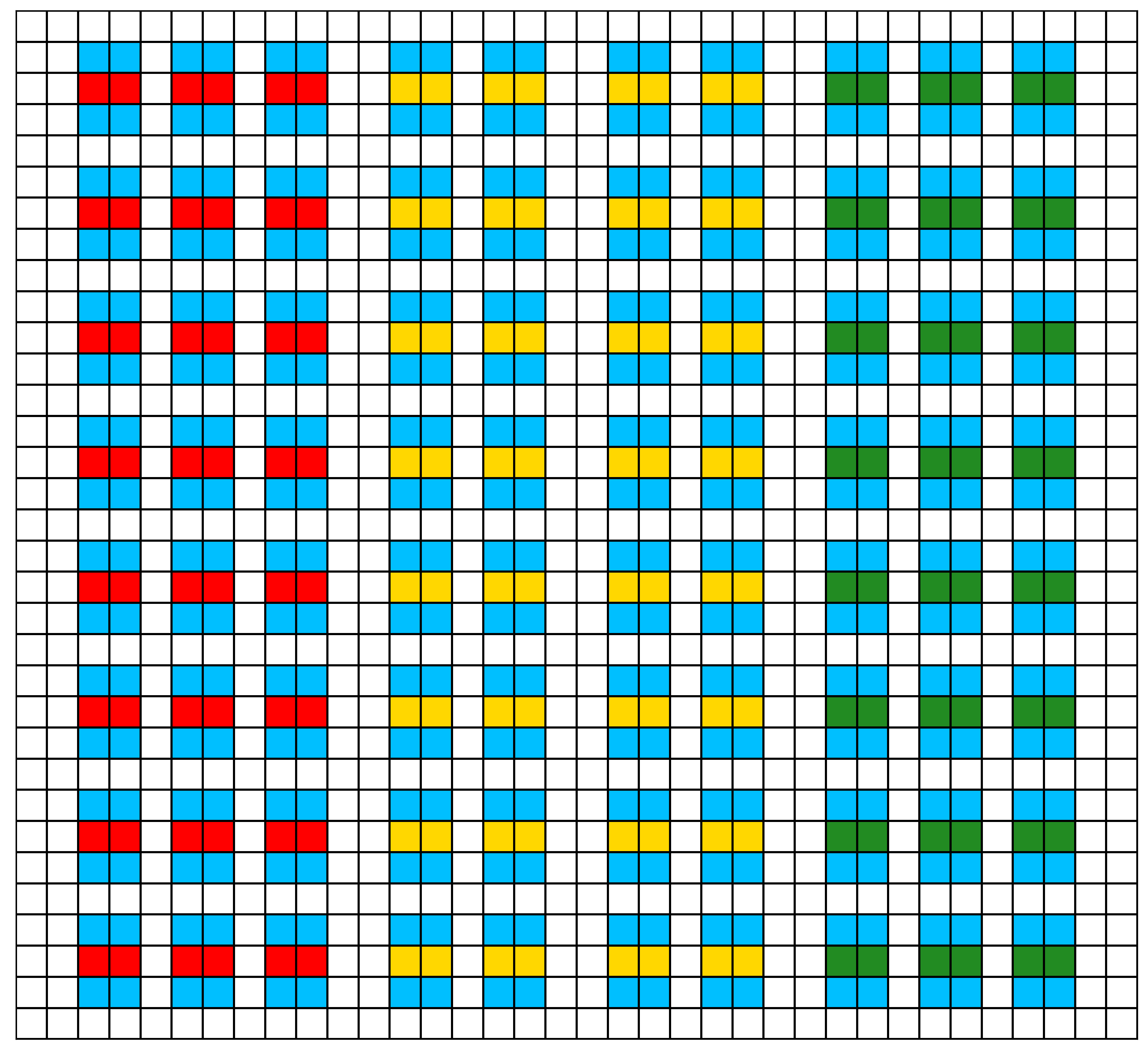}
        \caption{Manufacturing}
        \label{fig:manufacture_exp}
    \end{subfigure}
    \hfill
    \begin{subfigure}[t]{0.305\textwidth}
        \includegraphics[width=1\textwidth]{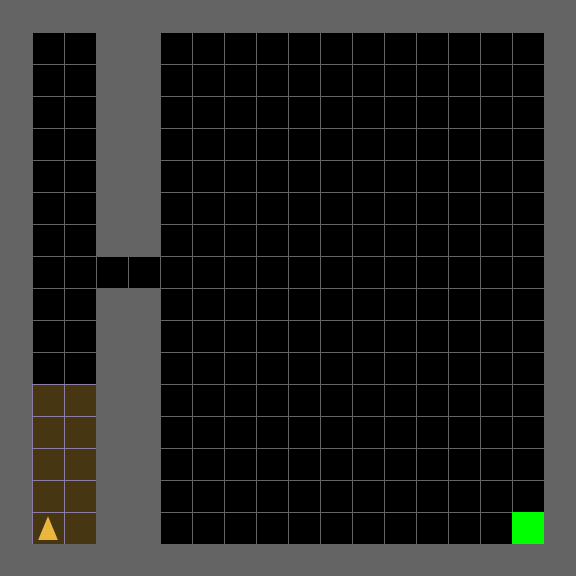}
        \caption{Maze}
        \label{fig:maze_exp}
    \end{subfigure}
    \caption{Example environments of the warehouse, manufacturing, and maze domains.}
\end{figure}

\section{Domain Details} \label{appen:domain}

\subsection{Warehouse}

\noindent \textbf{Environment.} We use the workstation scenario from previous work~\cite{zhangLayout23}. \Cref{fig:warehouse_exp} shows an example. A warehouse environment has four tile types. Black tiles are shelves for storing goods and also serve as obstacles. Blue tiles are endpoints where the agents park and interact with shelves. White tiles are empty spaces. Pink tiles are workstations where agents interact with human staff. Agents can traverse all non-black tiles. The goal location (= task) of each agent alternates between randomly selected workstations and endpoints. Furthermore, the NCA generator only generates the \textit{storage area} inside the red box in \Cref{fig:warehouse_exp}, which contains only black, blue, and white tiles. The \textit{non-storage area} outside the red box is kept unchanged. This is because the locations of the workstations are usually fixed along the borders of the real-world warehouses.

\noindent \textbf{Domain-Specific Constraints.} The constraints of the warehouse environment include: (1) all non-black tiles are connected so that the agents can reach all endpoints and workstations, (2) each black tile is adjacent to at least one blue tile and vice versa, ensuring that agents can interact with shelves via endpoints, (3) the number of black tiles is equal to a predefined number ($N_s$ or $N_{s\_eval}$ in \Cref{tab:search-space}) so that all generated warehouse environments have the same storage capability, and (4) the non-storage area is kept unchanged. The detailed MILP formulation of the constraints is included in the previous paper~\cite{zhangLayout23}.

\noindent \textbf{Agent-based Simulator.} Following previous work~\cite{zhangLayout23}, we use Rolling-Horizon Collision Resolution (RHCR)~\cite{Li2020LifelongMP}, a state-of-the-art lifelong MAPF planner, in our simulations. In lifelong MAPF, agents are constantly assigned new tasks once they finish their previous ones. At every $h$ timesteps, RHCR plans collision-free paths for all agents for the next $w$ timesteps ($w \geq h$). Following the design recommendation from the previous work~\cite{Li2020LifelongMP}, we use $w=10$ and $h=5$. We set the initial locations of the agents uniformly at random from the non-shelf tiles. We use two different task assigners, leading to two variations of the warehouse domain. First, we adopt the task assigner from the previous work~\cite{zhangLayout23} that assigns the next task (workstation or endpoint) uniformly at random. We refer to this variant as \textit{warehouse (even)}. Second, we extend the previous task assigner by asking it to select the next workstation with uneven probabilities. Specifically, the workstations on the left border are 5 times more likely to be selected than those on the right. The endpoints remain being evenly selected. We refer to this variant as \textit{warehouse (uneven)}, which is more challenging than warehouse (even) because the more frequently visited left border workstations could make the simulation to be congested.

\subsection{Manufacturing}

\noindent \textbf{Environment.} \Cref{fig:manufacture_exp} shows an example of a manufacturing environment. Each manufacturing environment has five tile types. The blue (endpoint) and white (empty space) tiles are the same as the warehouse environments. The red, green, and yellow tiles represent three types of workstations. Agents can traverse only blue and white tiles. The task of each agent is to go to an endpoint adjacent to a red, green, and yellow workstation in order and stay there for $t_r$, $t_g$, and $t_y$ timesteps, respectively. 
The NCA generator generates the entire environment. Since the manufacturing domain has multiple types of workstations, one of the challenges of generating high-performance manufacturing environments is to find a reasonable ratio of the number of different workstations.

\noindent \textbf{Domain-Specific Constraints.} In line with the warehouse domain, we constrain the manufacturing environments such that (1) all blue and white tiles are connected so that the robots can reach all endpoints, (2) each blue tile is adjacent to at least one of red, green, or yellow tiles and vice versa so that the robots can interact with the workstations via endpoints, (3) there is at least one red, one green, and one yellow tile to support the cyclical nature of manufacturing tasks, where a complete cycle involves sequential visits to endpoints next to red, green, and yellow workstations.

\noindent \textbf{Agent-based Simulator.} The agent-based simulator is the same as the one used in the warehouse domain except that (1) we ask each agent to stay for $t_r = 2$, $t_g = 5$, and $t_y = 10$ timesteps after arriving at their goal endpoints next to red, green, and yellow workstations, respectively, and (2) the task assigner assigns the next task to an endpoint, chosen uniformly at random, near the red, green, and yellow workstations in order.

\subsection{Maze}

\noindent \textbf{Environment.} We use the maze domain from previous works~\cite{Bhatt2022DeepSA,gym_minigrid,Dennis2020EmergentCA,ParkerHolder2022EvolvingCW}. \Cref{fig:maze_exp} shows an example maze environment. A maze environment has two tile types: wall (gray) and empty space (black). The yellow triangle represents the agent, and the green square is the goal location. We omit the agent and the goal location when generating the environments.

\noindent \textbf{Domain-Specific Constraints.} We do not have any domain-specific constraints for the maze domain. 

\noindent \textbf{Agent-based Simulator.} We use a trained ACCEL agent~\cite{Dennis2020EmergentCA}. The observation space is a 5 $\times$ 5 grid in front of it. The agent can turn left or right in its current tile, move forward to the adjacent tile, or stay in the current tile. We assign the start and goal locations of the agent to the first and last tile of the longest shortest path in the environment.
We limit the time horizon to 2 times the number of tiles in the environment, which is 648 for environments of size $S$, and 8,712 for environments of size $S_{eval}$, respectively. We stop the simulation early if the agent reaches the goal.

\section{Experiment Details} \label{appen:exp_setup}

\begin{figure}[!t]
    \centering
    \includegraphics[width=1\textwidth]{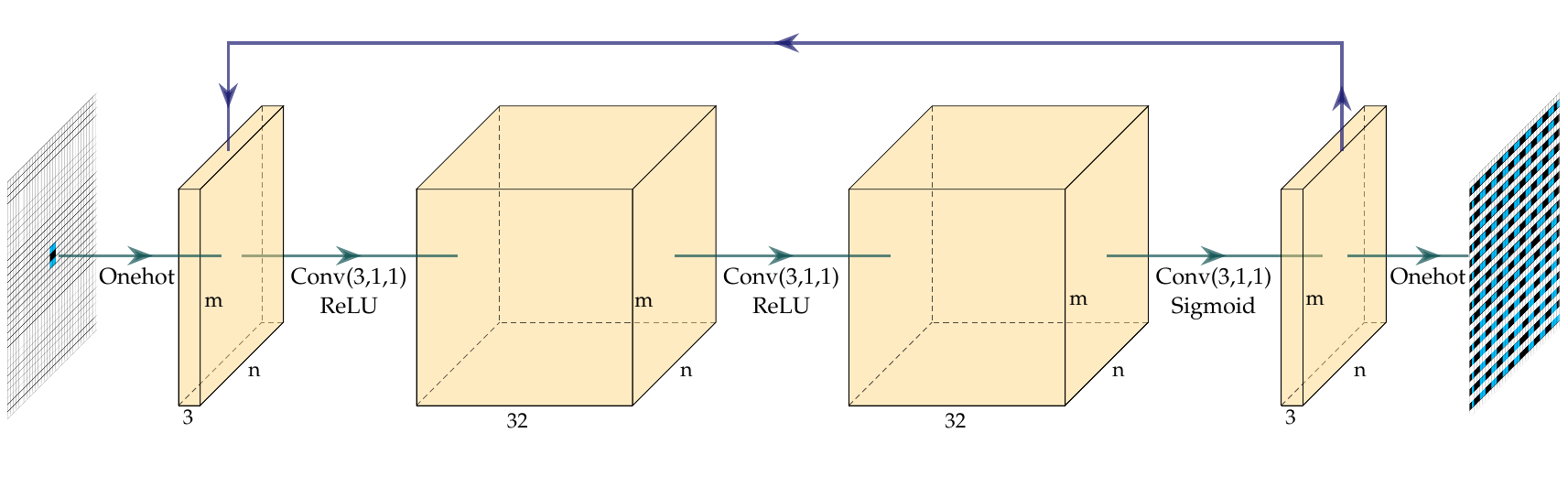}
    \caption{Architecture of the NCA generator. Starting with a fixed onehot-encoded initial environment, the generator transforms an initial environment iteratively to generate the final environment.}
    \label{fig:nca_model_arch}
\end{figure}

\subsection{NCA Setup}
\label{appen:nca_gen}

\subsubsection{NCA Generator Architecture}
\label{apen:nca_gen:nca_arch}

\Cref{fig:nca_model_arch} shows the model architecture of the NCA generator.
The generator has 3 convolutional layers of kernel size 3, stride 1 and padding 1. The first 2 convolutional layers have 32 output channels and are followed by a ReLU activation. The last layer has 3 output channels and is followed by a sigmoid function. This configuration guarantees that the width, height, and number of channels of the input and output tensors are the same.

\subsubsection{Initial Environments}

\begin{figure}[!t]
    \centering
    \begin{subfigure}[t]{0.295\textwidth}
        \includegraphics[width=1\textwidth]{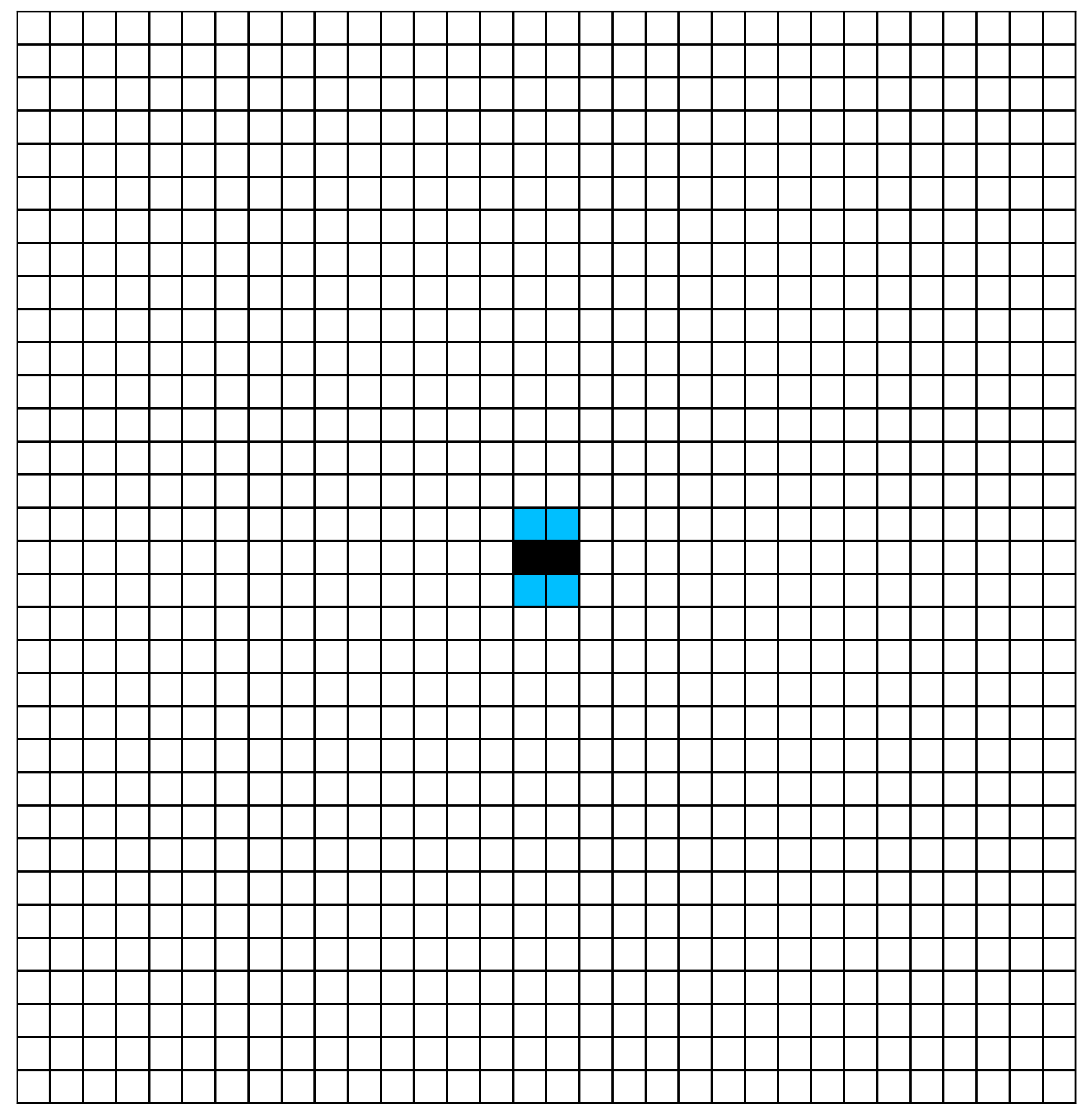}
        \caption{Warehouse}
        \label{fig:warehouse_seed}
    \end{subfigure}
    \hfill
    \begin{subfigure}[t]{0.33\textwidth}
        \includegraphics[width=1\textwidth]{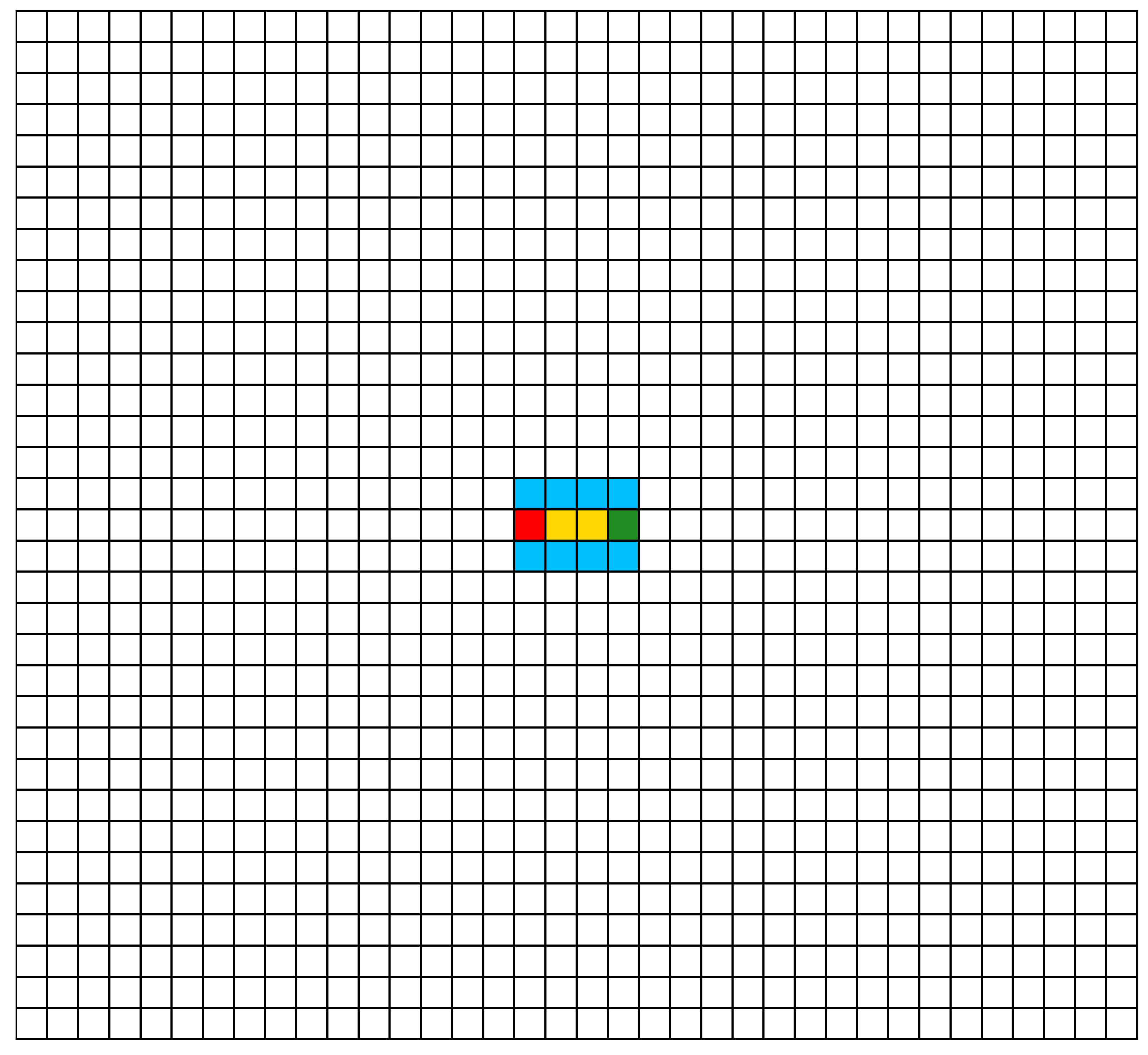}
        \caption{Manufacturing}
        \label{fig:manufacture_seed}
    \end{subfigure}
    \hfill
    \begin{subfigure}[t]{0.305\textwidth}
        \includegraphics[width=1\textwidth]{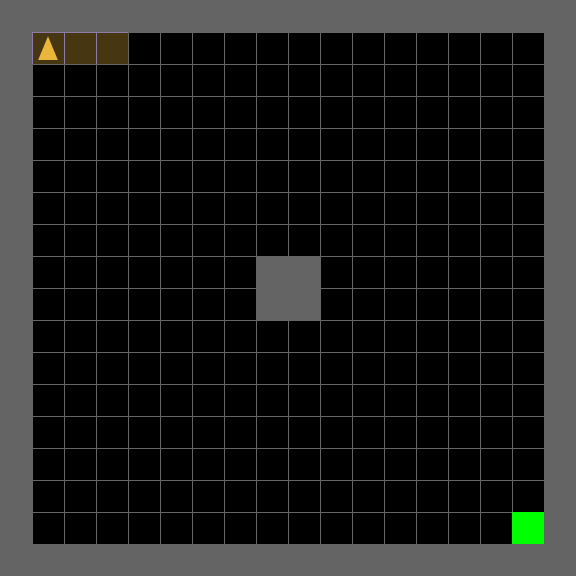}
        \caption{Maze}
        \label{fig:maze_seed}
    \end{subfigure}
    \caption{Initial environments of NCA generation.}
    \label{fig:nca_seed}
\end{figure}

\Cref{fig:nca_seed} shows the initial environments of size $S$ for all domains. They are characterized by a small central block of non-empty-spaces surrounded by empty spaces. The initial environments of size $S_{eval}$ maintains the same central blocks and are surrounded by more expansive empty spaces.

\subsubsection{Postprocessing Generated Environments}

After generating an environment and before repairing it, we may need to postprocess the environment to prepare for repairing.

\noindent \textbf{Warehouse.} In the warehouse domain, we generate a 32 $\times$ 33 and 98 $\times$ 101 storage area and add fixed non-storage area to the left and right side to form an environment of size $S = 36 \times 33$ and $S_{eval} = 102 \times 101$, respectively.

\noindent \textbf{Manufacturing.} In the manufacturing domain, we directly generate environments of size $S = 36 \times 33$ and $S_{eval} = 102 \times 101$.

\noindent \textbf{Maze.} In the maze domain, we generate a 16 $\times$ 16 environment and surround it with walls to create a $S = 18 \times 18$ environment. Similarly, we generate a 64 $\times$ 64 environment and surround it with walls to create a  $S_{eval} = 66 \times 66$  environment.

\subsection{QD Setup}
\label{appen:qd_setup}

\subsubsection{Objective}

\noindent \textbf{Warehouse.} We use $f_{res}$ and $f_{opt}$ in \Cref{sec:method} as the objective functions. We set $p_i = 1$ for all $i$, meaning that all tiles have the same weight while computing the similarity score. 

\noindent \textbf{Manufacturing.} We use $f_{res}$ and $f_{opt}$ in \Cref{sec:method} as the objective functions. We set $\alpha = 5$ as it results in the most scalable environments in the warehouse domain. Suppose $\mathbf{x}_i$ denotes the $i^{\text{th}}$ tile of the environment $\mathbf{x} \in \mathbf{X}$. For the objective functions, we let $p_i = 5$ if $(\mathbf{x}_{in})_i$ is a workstation, and $p_i = 1$ otherwise. Intuitively, we give more rewards to the environments in which workstations are not changed during MILP repair.

\noindent \textbf{Maze.} We use the same binary objective from the previous work~\cite{Bhatt2022DeepSA}.

\begin{figure}[!t]
    \centering
    \begin{minipage}[b]{0.2\textwidth}
        \centering
        Baseline
        \vspace{1.5cm}
    \end{minipage}
    \begin{subfigure}[t]{0.25\textwidth}
        \includegraphics[width=1\textwidth]{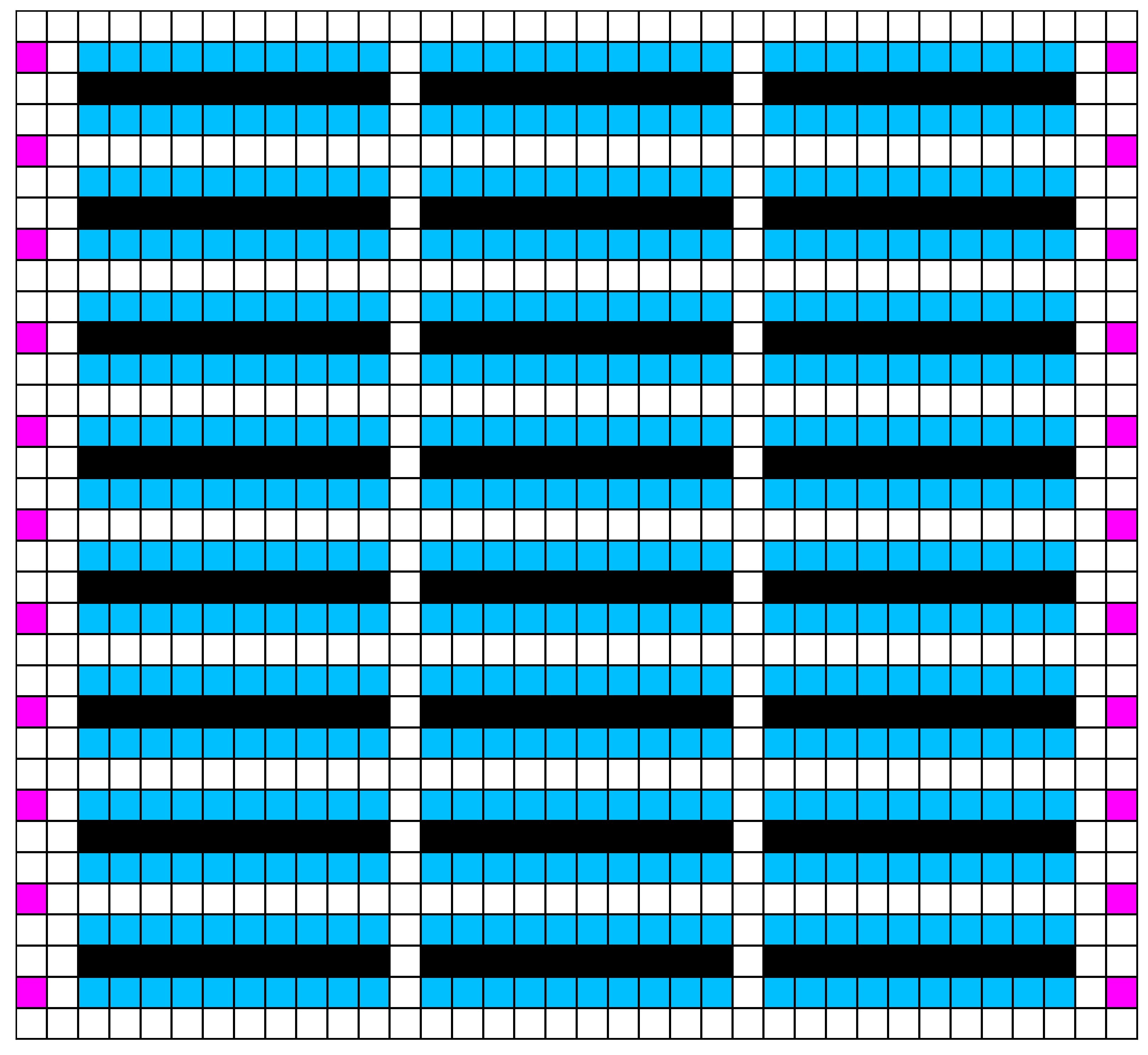}
        \caption{Human-designed}
        \label{fig:warehouse-large-human}
    \end{subfigure}
    \hfill
    \begin{subfigure}[t]{0.25\textwidth}
        \includegraphics[width=1\textwidth]{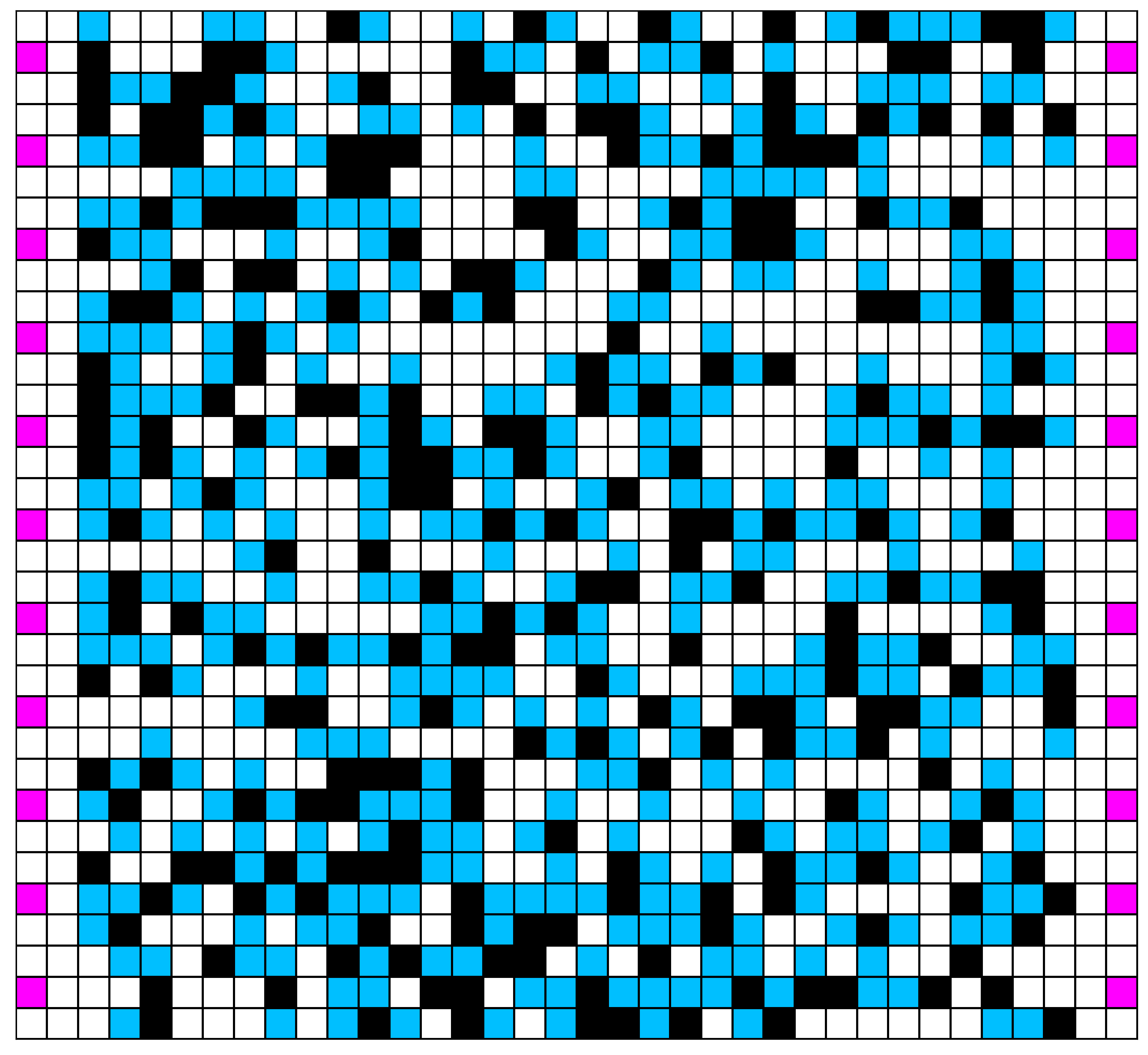}
        \caption{DSAGE (even)}
        \label{fig:warehouse-large-dsage-even}
    \end{subfigure}
    \hfill
    \begin{subfigure}[t]{0.25\textwidth}
        \includegraphics[width=1\textwidth]{maps/warehouse/kiva_large_200_agents_dsage_opt_entropy_throughput_hamming_uneven_a=5.png}
        \caption{DSAGE (uneven)}
        \label{fig:warehouse-large-dsage-uneven}
    \end{subfigure}\\
    \begin{minipage}[b]{0.20\textwidth}
        \centering
        CMA-MAE with\\
        warehouse (even)
        \vspace{1.3cm}
    \end{minipage}
    \begin{subfigure}[t]{0.25\textwidth}
        \includegraphics[width=1\textwidth]{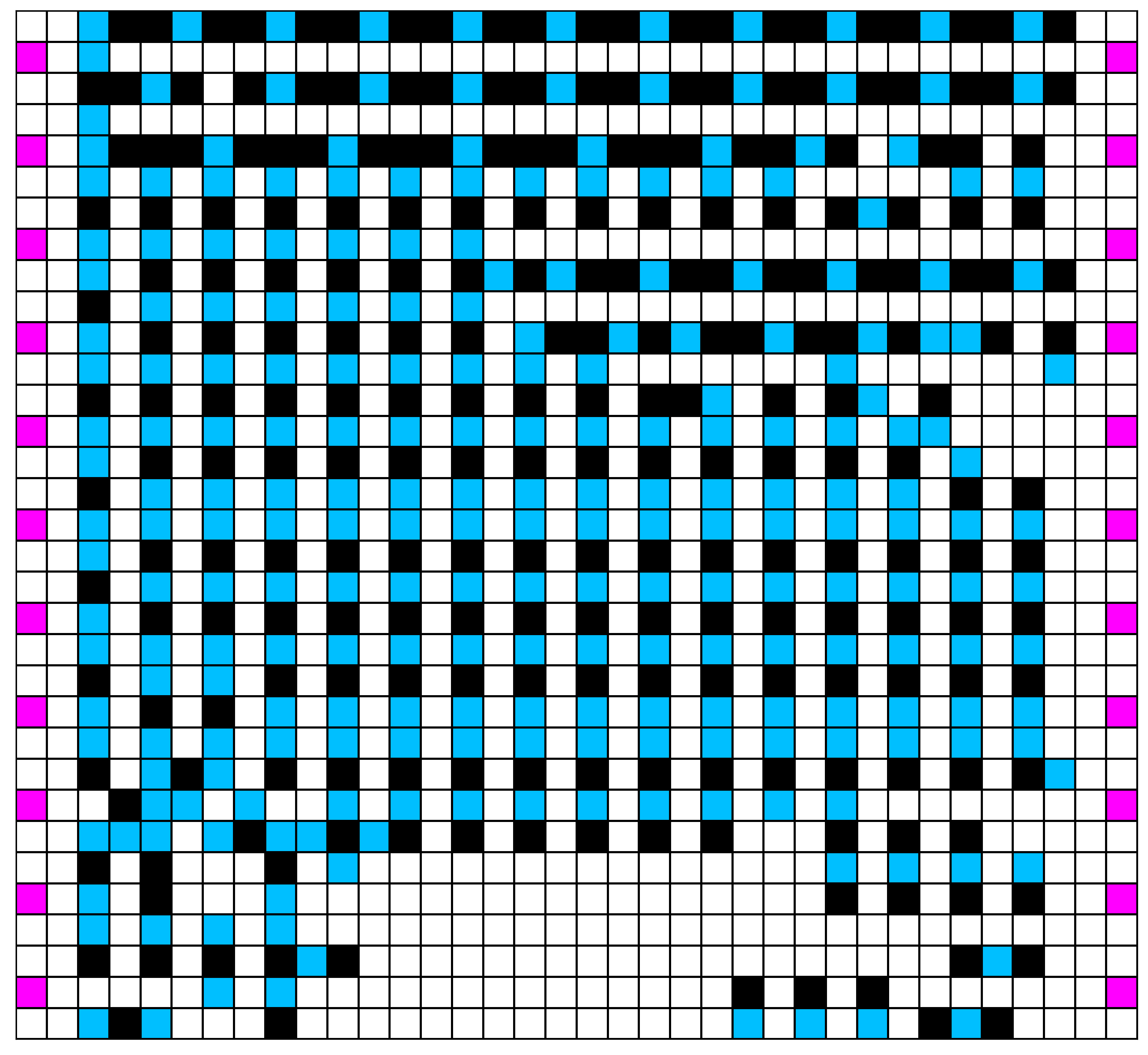}
        \caption{CMA-MAE ($\alpha = 0$)}
        \label{fig:warehouse-large-cma-mae-a=0-even}
    \end{subfigure}
    \hfill
    \begin{subfigure}[t]{0.25\textwidth}
        \includegraphics[width=1\textwidth]{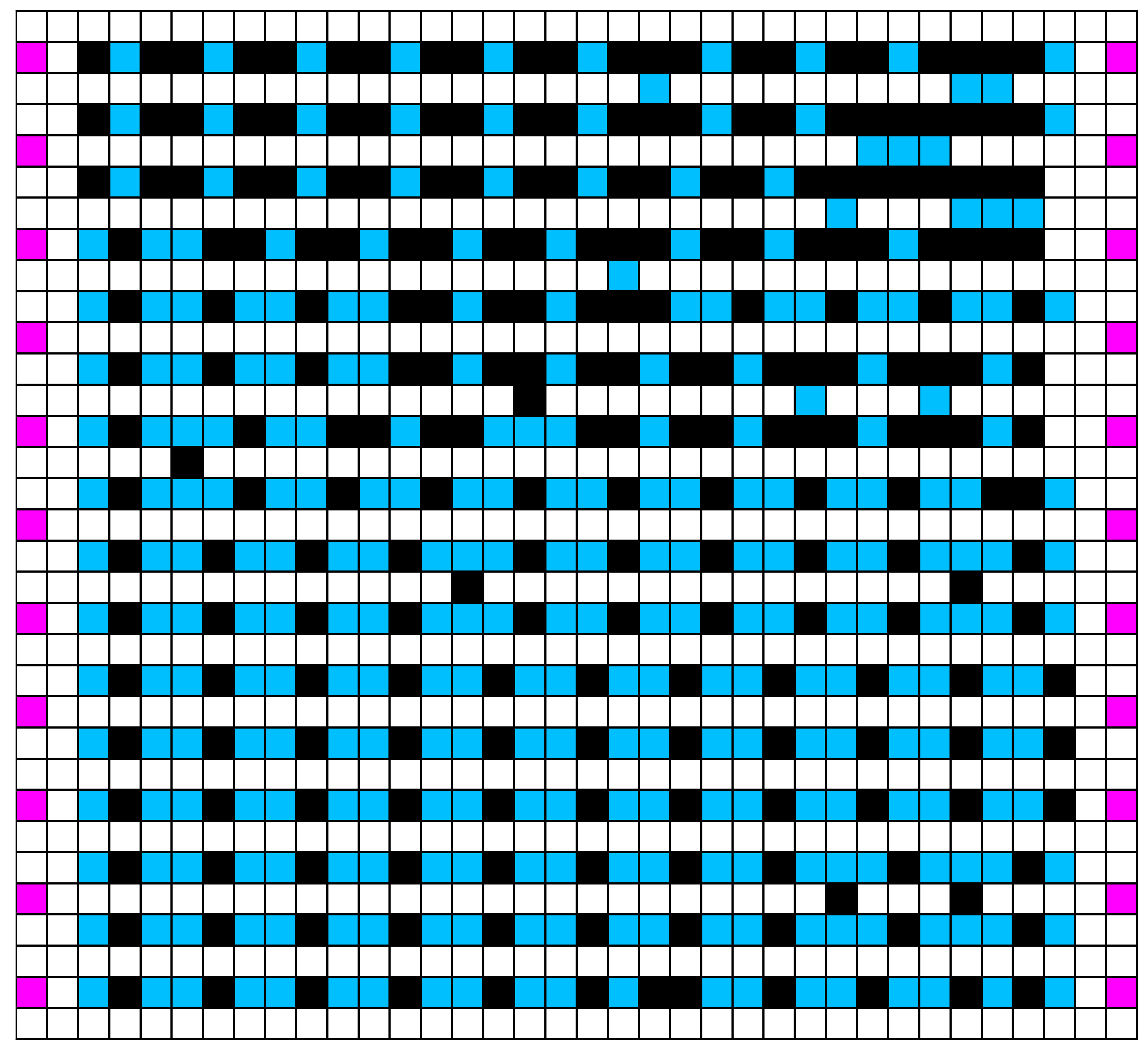}
        \caption{CMA-MAE ($\alpha = 1$)}
        \label{fig:warehouse-large-cma-mae-a=1-even}
    \end{subfigure}
    \hfill
    \begin{subfigure}[t]{0.25\textwidth}
        \includegraphics[width=1\textwidth]{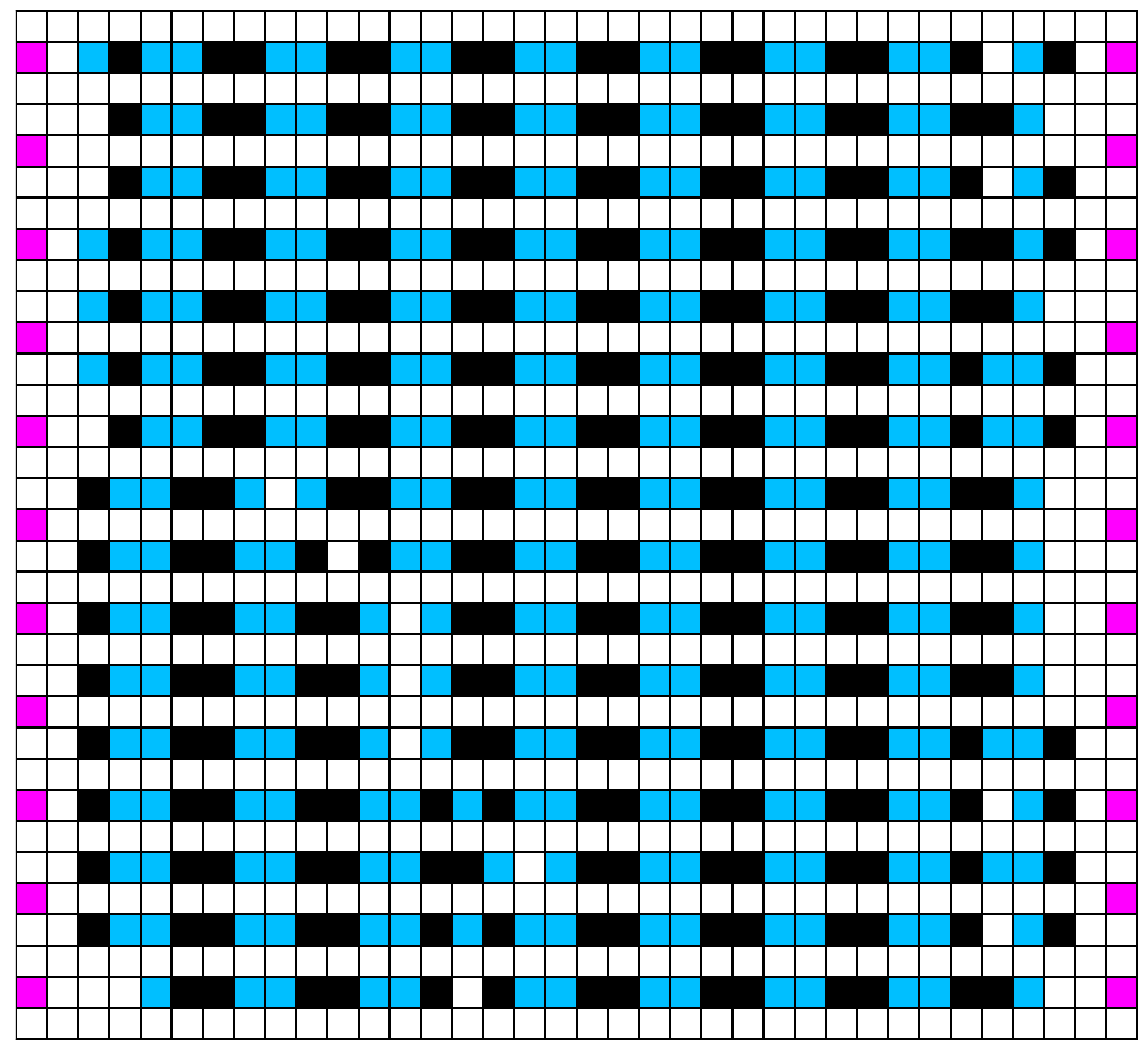}
        \caption{CMA-MAE ($\alpha = 5$)}
        \label{fig:warehouse-large-cma-mae-a=5-even}
    \end{subfigure}\\
    \begin{minipage}[b]{0.20\textwidth}
        \centering
        CMA-MAE with\\
        warehouse (uneven)
        \vspace{0.9cm}
    \end{minipage}
    \begin{subfigure}[t]{0.25\textwidth}
        \includegraphics[width=1\textwidth]{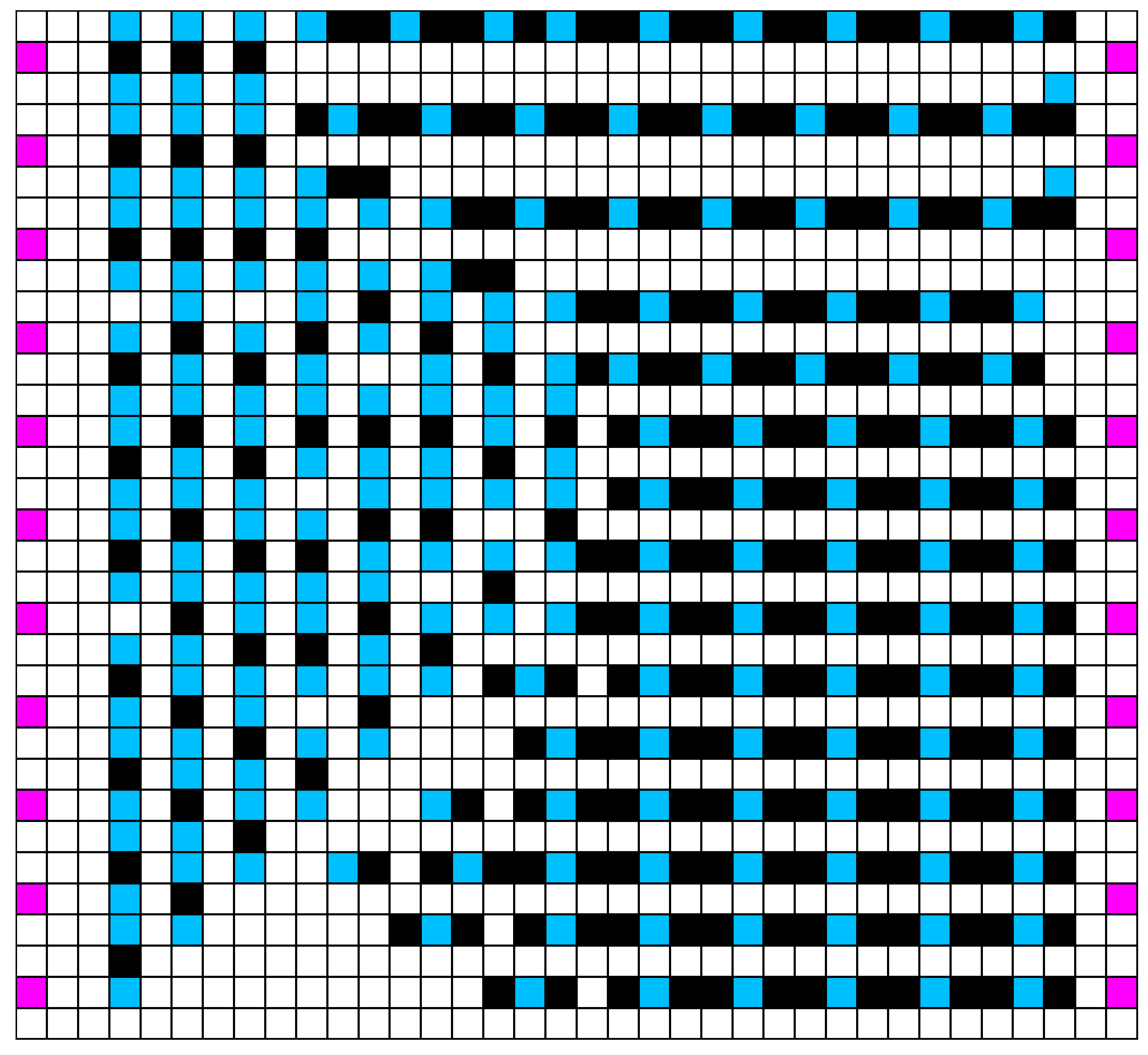}
        \caption{CMA-MAE ($\alpha = 0$)}
        \label{fig:warehouse-large-cma-mae-a=0-uneven}
    \end{subfigure}
    \hfill
    \begin{subfigure}[t]{0.25\textwidth}
        \includegraphics[width=1\textwidth]{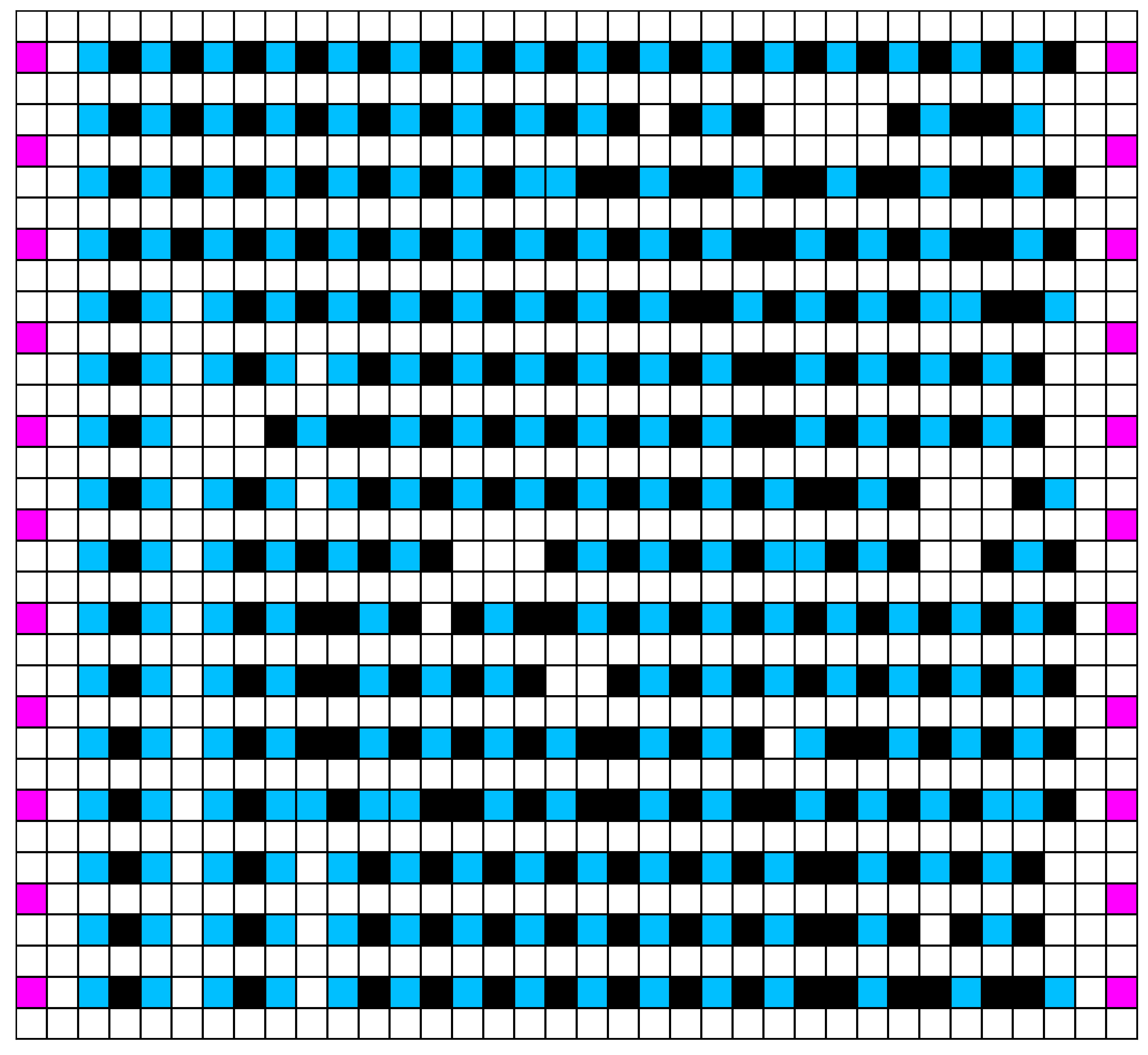}
        \caption{CMA-MAE ($\alpha = 1$)}
        \label{fig:warehouse-large-cma-mae-a=1-uneven}
    \end{subfigure}
    \hfill
    \begin{subfigure}[t]{0.25\textwidth}
        \includegraphics[width=1\textwidth]{maps/warehouse/kiva_large_200_agents_cma-mae_opt_entropy_uneven_throughput_hamming_a=5.png}
        \caption{CMA-MAE ($\alpha = 5$)}
        \label{fig:warehouse-large-cma-mae-a=5-uneven}
    \end{subfigure}
    \caption{Baseline and NCA-generated warehouse environments of size $S$.}
    \label{fig:warehouse-env-large}
\end{figure}

\begin{figure}[!t]
    \centering
    \begin{minipage}[b]{0.32\textwidth}
        \centering
        Baseline
        \begin{subfigure}[t]{1\textwidth}
            \centering
            \includegraphics[width=1\textwidth]{maps/manufacture/manufacture_large_93_stations.png}
            \caption{Human-designed\\ \ }
            \label{fig:manufacture-large-human}
        \end{subfigure}\\
        \begin{subfigure}[t]{1\textwidth}
            \centering
            \includegraphics[width=1\textwidth]{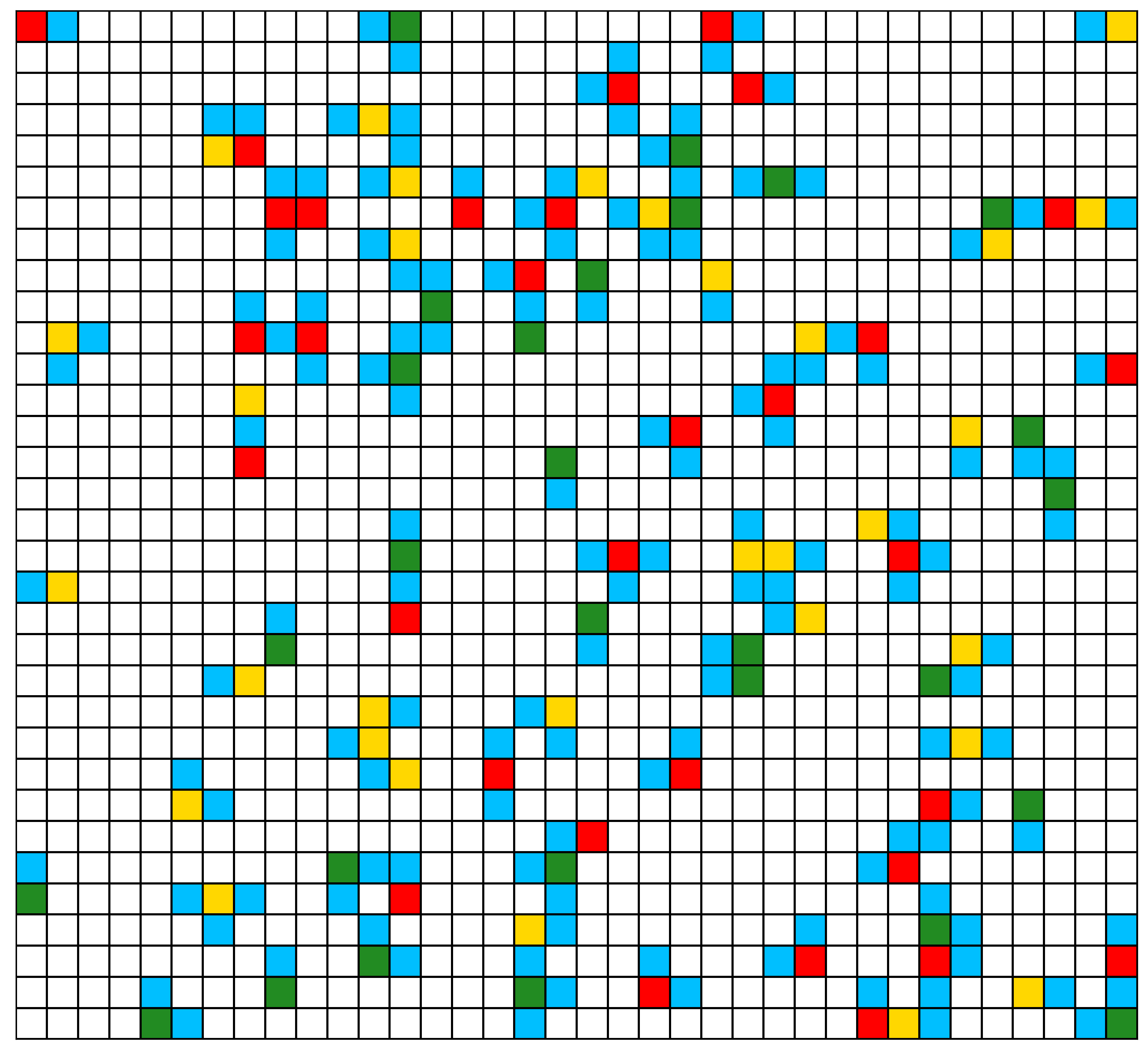}
            \caption{DSAGE}
            \label{fig:manufacture-large-dsage}
        \end{subfigure}
    \end{minipage}
    \begin{minipage}[b]{0.32\textwidth}
        \centering
        NCA-generated (Repaired)
        \begin{subfigure}[t]{1\textwidth}
            \includegraphics[width=1\textwidth]{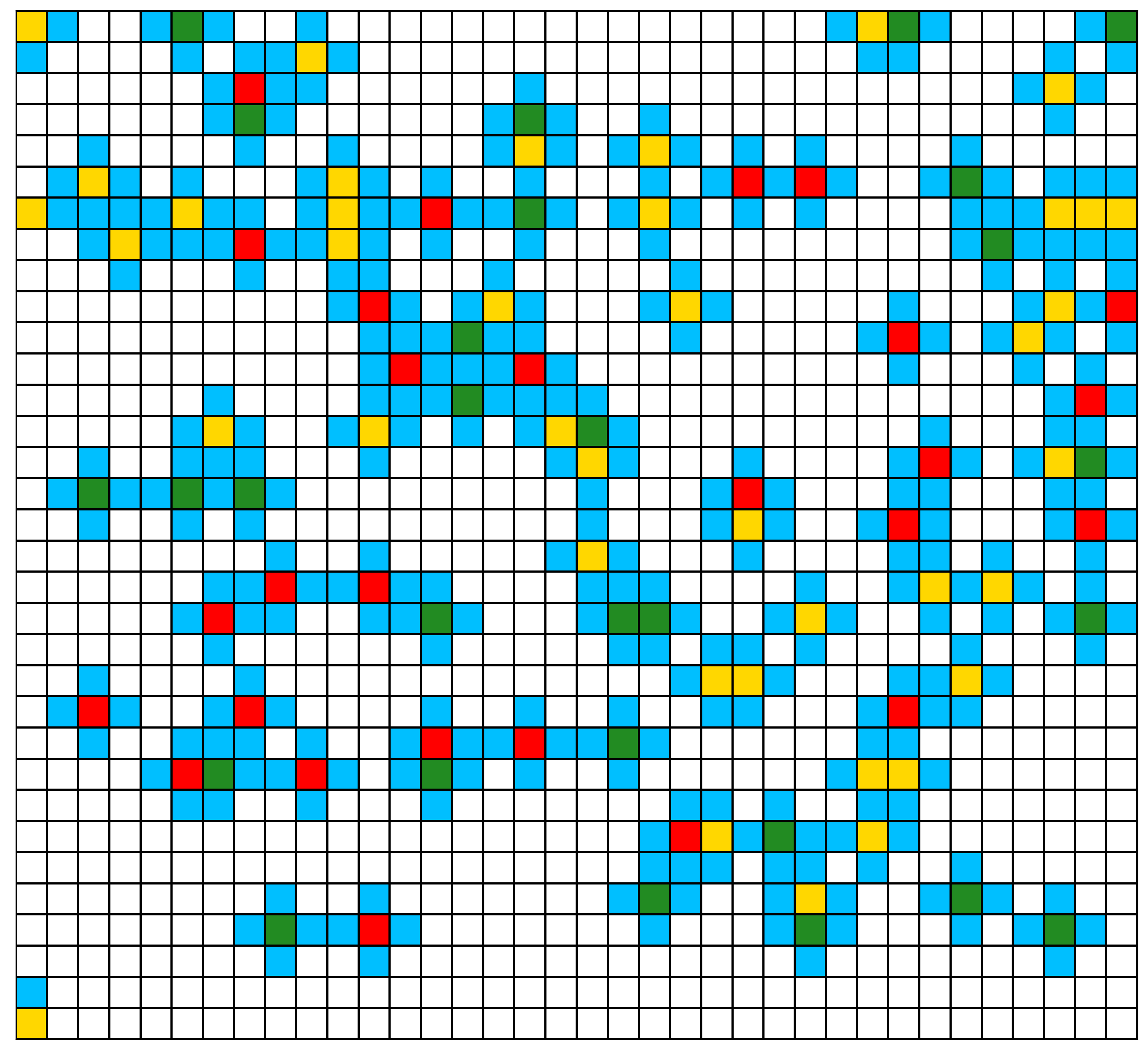}
            \caption{CMA-MAE ($\alpha = 5$, comp DSAGE)}
            \label{fig:manufacture-large-cma-mae-a=5-comp-dsage}
        \end{subfigure}
        \hfill
        \begin{subfigure}[t]{1\textwidth}
            \includegraphics[width=1\textwidth]{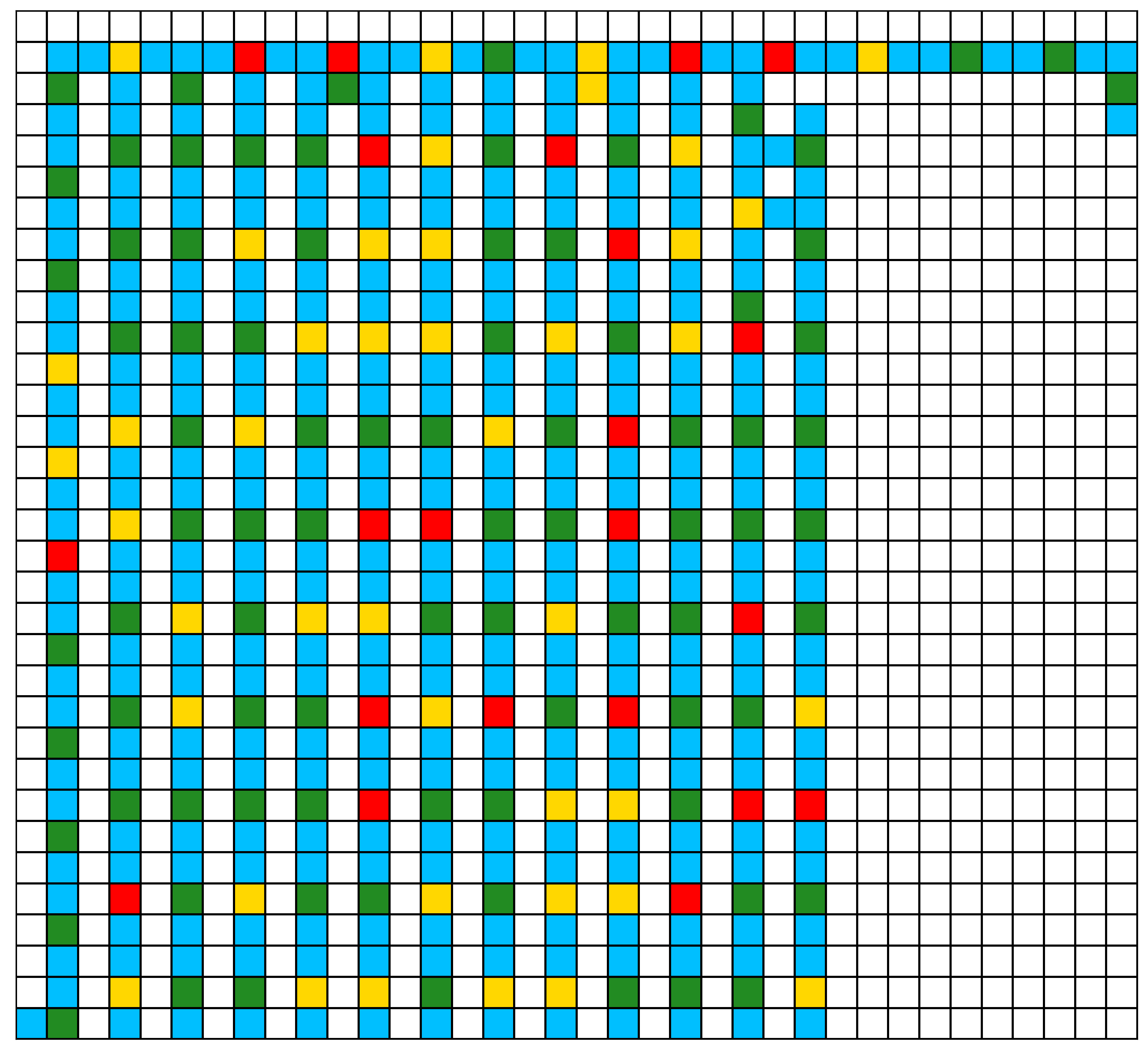}
            \caption{CMA-MAE ($\alpha = 5$, opt)}
            \label{fig:manufacture-large-cma-mae-a=5-opt}
        \end{subfigure}
    \end{minipage}
    \begin{minipage}[b]{0.32\textwidth}
        \centering
        NCA-generated (Unrepaired)
        \begin{subfigure}[t]{1\textwidth}
            \includegraphics[width=1\textwidth]{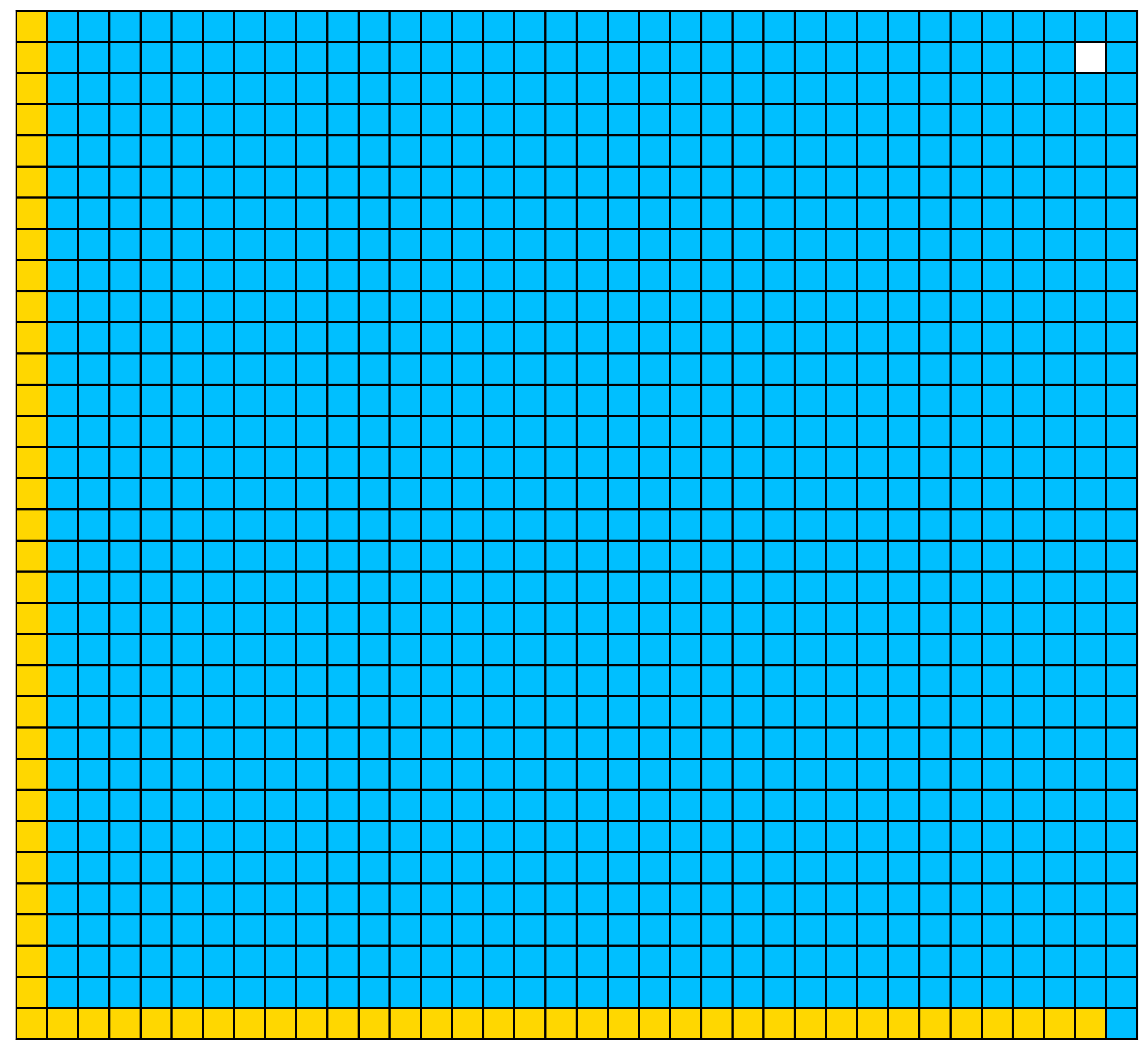}
            \caption{CMA-MAE ($\alpha = 5$, comp DSAGE)}
            \label{fig:manufacture-large-cma-mae-a=5-comp-dsage-unrepaired}
        \end{subfigure}
        \hfill
        \begin{subfigure}[t]{1\textwidth}
            \includegraphics[width=1\textwidth]{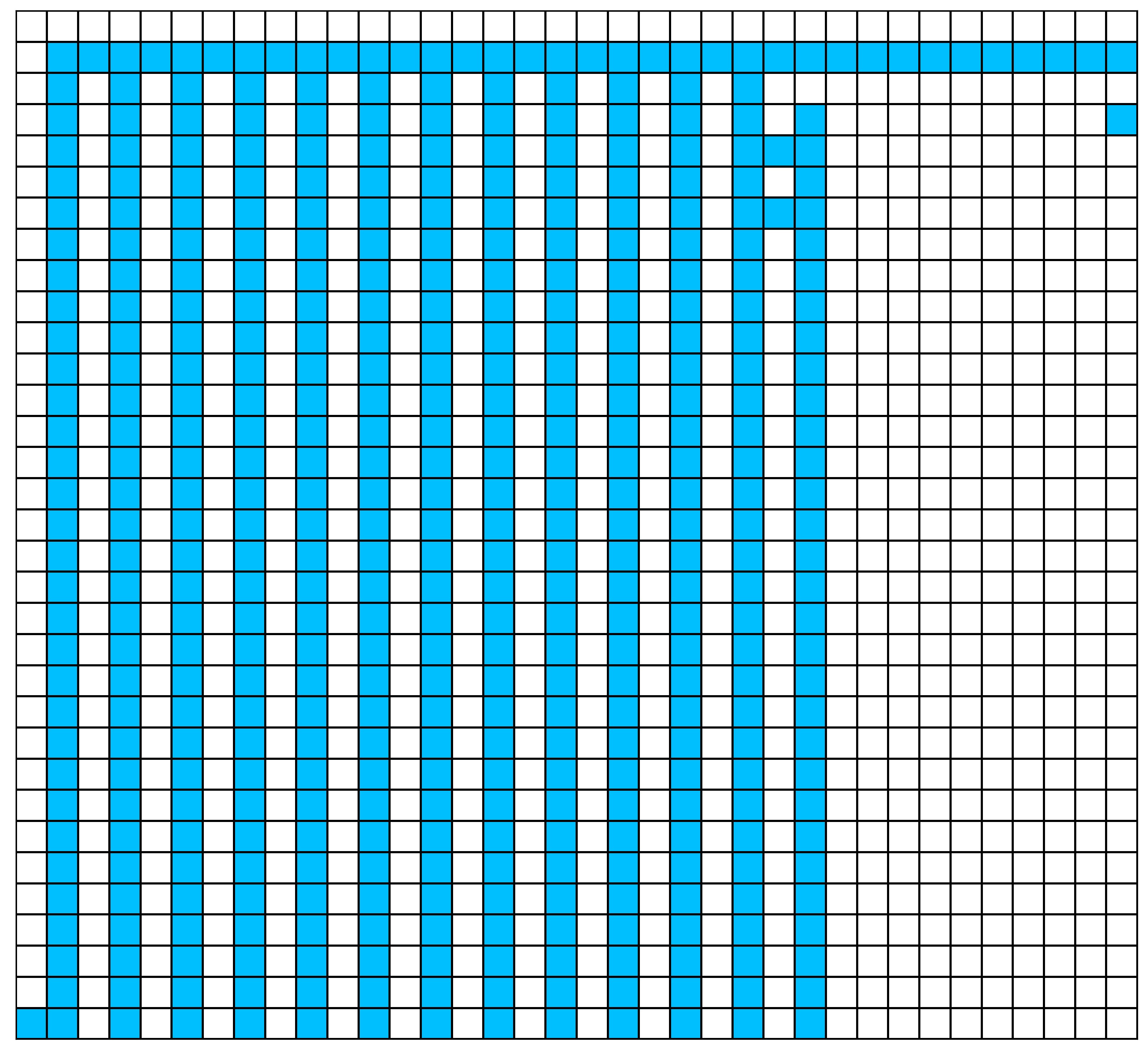}
            \caption{CMA-MAE ($\alpha = 5$, opt)}
            \label{fig:manufacture-large-cma-mae-a=5-opt-unrepaired}
        \end{subfigure}
    \end{minipage}
    \caption{Baseline and NCA-generated manufacturing environments of size $S$.}
    \label{fig:manufacture-env-large}
\end{figure}

\subsubsection{Archive Dimensions}

\noindent \textbf{Warehouse.} We use a 100 $\times$ 100 archive for the warehouse domain. We set the range of the connected shelf components to be $[140, 240]$, following previous work~\cite{zhangLayout23}, and the range of the environment entropy to be $[0, 1]$. In DSAGE, we follow previous work to downsample the archive to get a subset of elites~\cite{Bhatt2022DeepSA,zhangLayout23}. We set the downsampled archive dimension to be $[50, 25]$, with 50 on the dimension of the number of connected shelf components and 25 on the environment entropy dimension. We use an irregular dimension because DSAGE fails to diversify the environment entropy measure (introduced in \Cref{appen:add-result:qd-score-archive-cover}) and we need to use a higher downsampling resolution on the number of connected shelf components to sample a reasonable number of elites.

\noindent \textbf{Manufacturing.} We use a 100 $\times$ 100 archive for the manufacturing domain. We set the range of the number of workstations to be $[0, 600]$ to allow CMA-MAE to explore a wide range of environments with different number of workstations, and the range of the environment entropy to be $[0, 1]$. We use an irregular 25 $\times$ 10 downsampled archive with higher resolution on the number of workstations for the same reason as the warehouse domain

\noindent \textbf{Maze.} Following previous work~\cite{Bhatt2022DeepSA}, we use a 256 $\times$ 162 archive. We set the range of the number of walls to be $[0, 256]$, where 256 is the total number of tiles in the environments of size $S$ (excluding the appended walls around the border), and the range of average agent path length to be $[0, 648]$, where 648 is the time horizon of the agent-based simulation in environments of size $S$.

\subsubsection{CMA-MAE}

For the warehouse and manufacturing domains, we use an archive learning rate of 0.01, while, for the maze domain, we use 0.5. We use a higher archive learning rate for the maze domain because a higher learning rate promotes CMA-MAE to explore the measure space. Since the binary objective function of the maze domain (introduced in \Cref{sec:domain}) can be trivially optimized, we promote CMA-MAE to do more exploration than exploitation in the maze domain. In all domains, the initial batch of solutions is sampled from a multivariate Gaussian distribution with a mean of 0 and a standard deviation of 0.2.

\subsection{Compute Resource} \label{appen:compute}

We run our experiments on 4 local machines and 2 high performing clusters, namely 
(1) a local machine with a 16-core AMD Ryzen 9 5950X CPU, 32 GB of RAM, and a Nvidia RTX 3080 GPU, 
(2) a local machine with a 64-core AMD Ryzen Threadripper 3990X, 192 GB of RAM, and an Nvidia RTX 3090Ti GPU, 
(3) a local machine with a 64-core AMD Ryzen Threadripper 3990X, 64 GB of RAM, and an Nvidia RTX A6000 GPU, 
(4) a local machine with a 64-core AMD Ryzen Threadripper 3990X, 64 GB of RAM, and an Nvidia RTX 3090 GPU, 
(5) a high-performing clusters with a V100 GPU, 200 Xeon CPUs each with 4 GB of RAM, and numerous 32-core AMD EPYC 7513 CPUs, each with up to 248 GB of RAM, 
(6) a high-performing cluster with a V100 GPU and heterogeneous CPUs. We perform our experiments in different machines because our experiments are not sensitive to runtime and require massive parallelization. We measure all CPU runtimes in machine (1).

\begin{figure}
    \begin{minipage}[b]{\EvalEnvShowSize\textwidth}
        \centering
        Warehouse (even)
        \begin{subfigure}[t]{1\textwidth}
            \centering
            \includegraphics[width=1\textwidth]{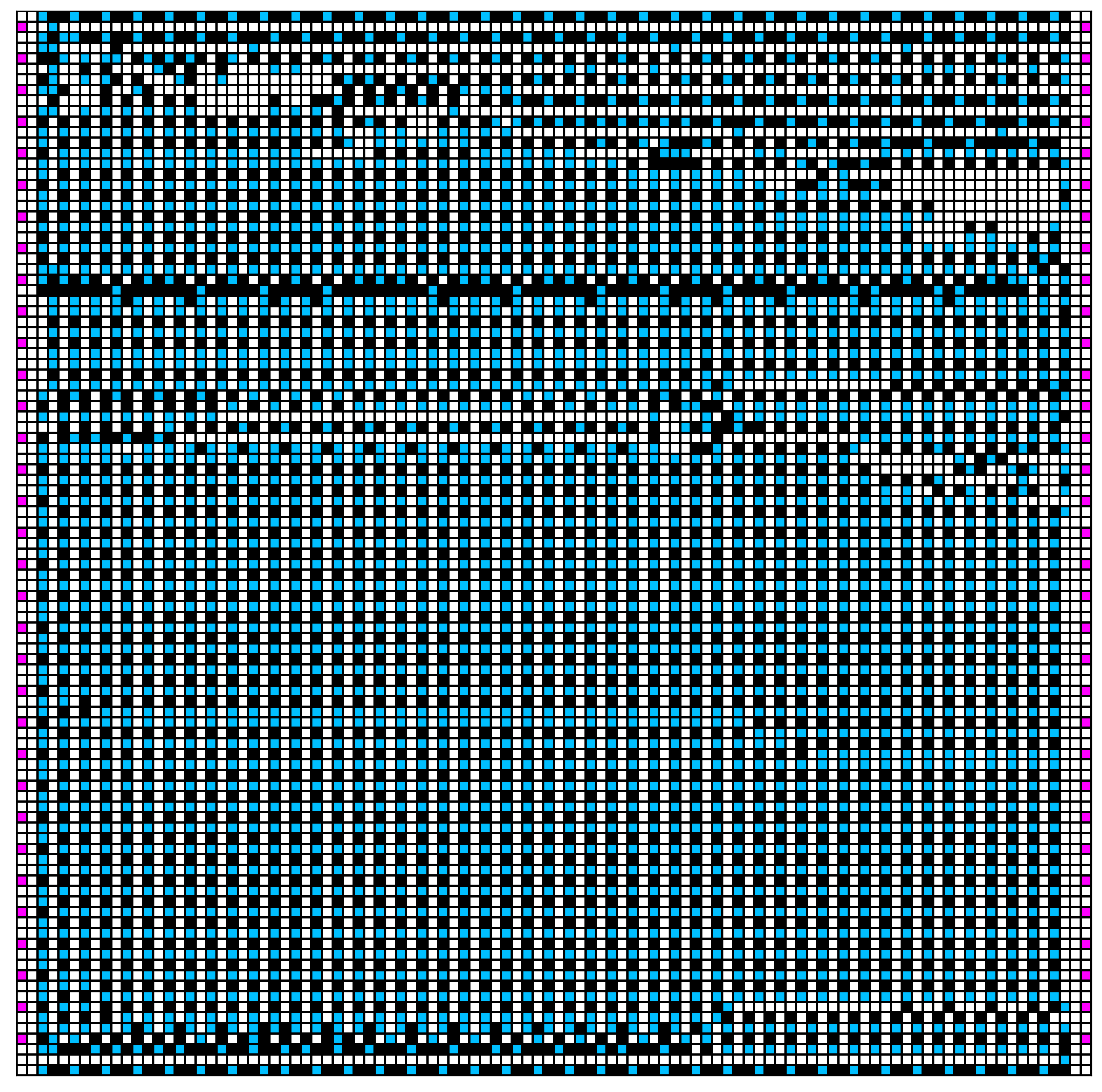}
            \caption{CMA-MAE ($\alpha = 0$)}
            \label{fig:warehouse-xxlarge-cma-mae-a=0-even}
        \end{subfigure}\\
        \begin{subfigure}[t]{1\textwidth}
            \centering
            \includegraphics[width=1\textwidth]{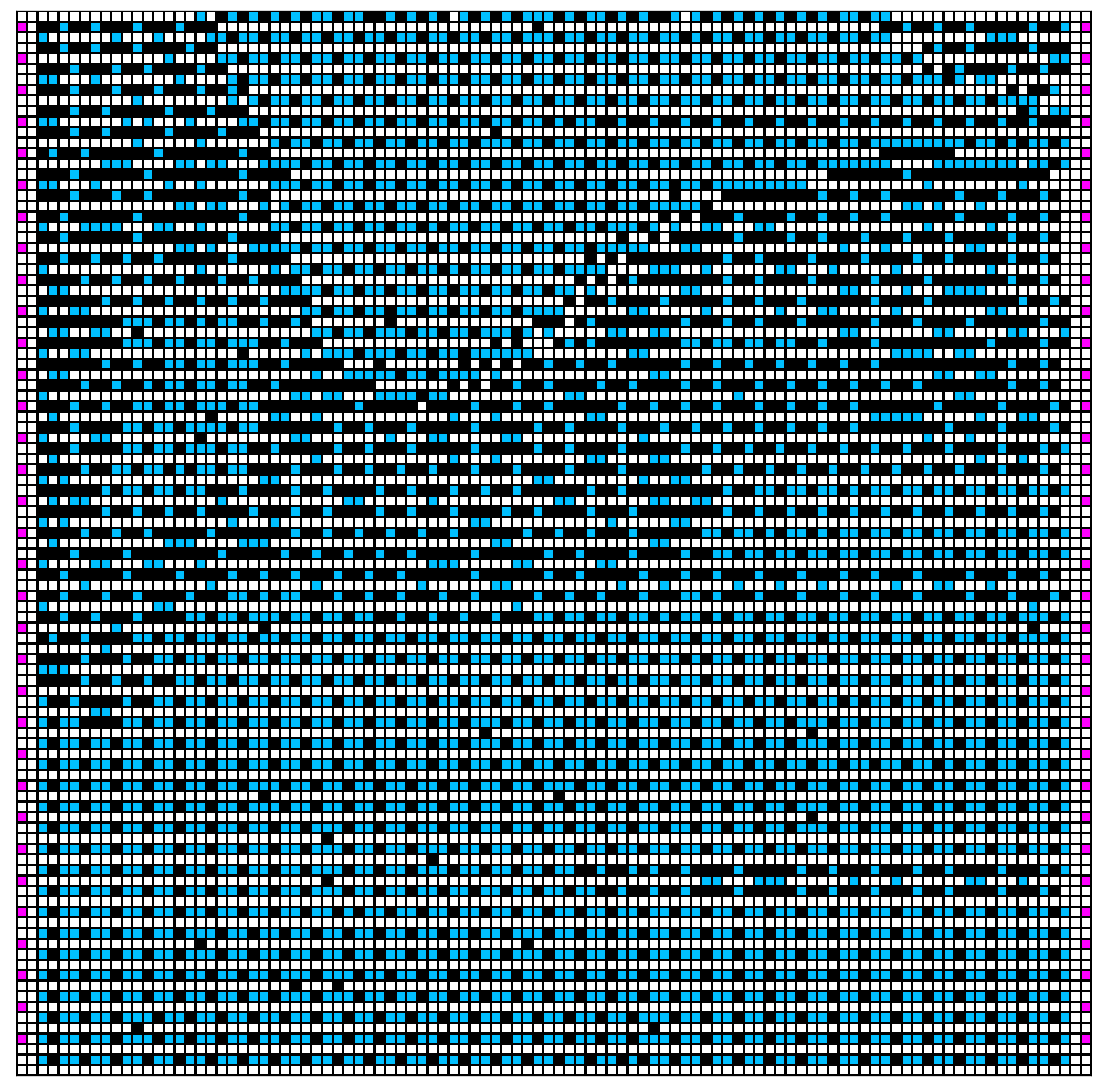}
            \caption{CMA-MAE ($\alpha = 1$)}
            \label{fig:warehouse-xxlarge-cma-mae-a=1-even}
        \end{subfigure}\\
        \begin{subfigure}[t]{1\textwidth}
            \centering
            \includegraphics[width=1\textwidth]{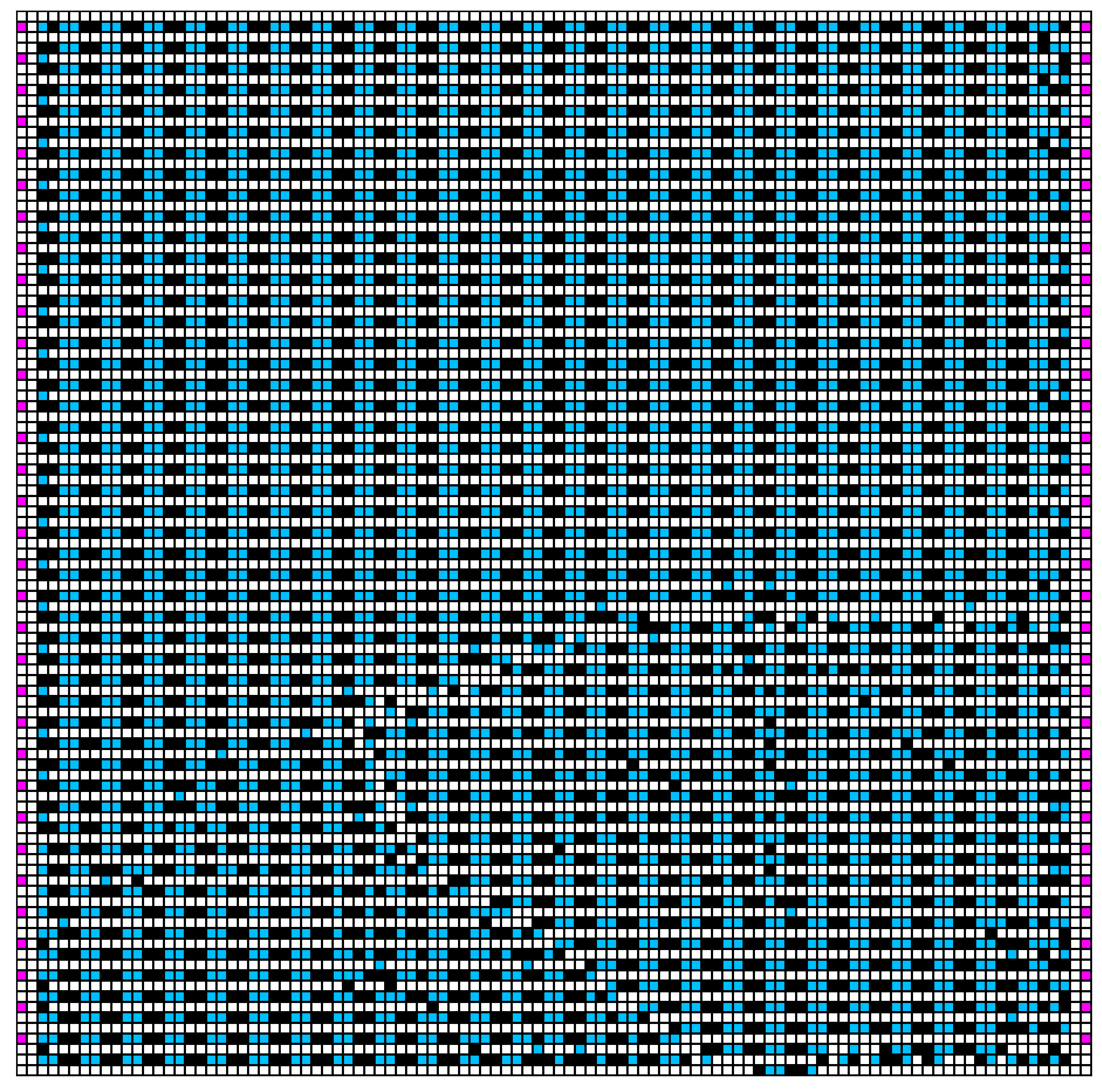}
            \caption{CMA-MAE ($\alpha = 5$)}
            \label{fig:warehouse-xxlarge-cma-mae-a=5-even}
        \end{subfigure}

    \end{minipage}
    \hfill
    \begin{minipage}[b]{\EvalEnvShowSize\textwidth}
        \centering
         Warehouse (uneven)
        \begin{subfigure}[t]{1\textwidth}
            \centering
            \includegraphics[width=1\textwidth]{maps/warehouse/kiva_xxlarge_cma-mae_global_opt_repaired_one_endpt_uneven_w_iter=200.png}
            \caption{CMA-MAE ($\alpha = 0$)}
            \label{fig:warehouse-xxlarge-cma-mae-a=0-uneven}
        \end{subfigure}\\
        \begin{subfigure}[t]{1\textwidth}
            \centering
            \includegraphics[width=1\textwidth]{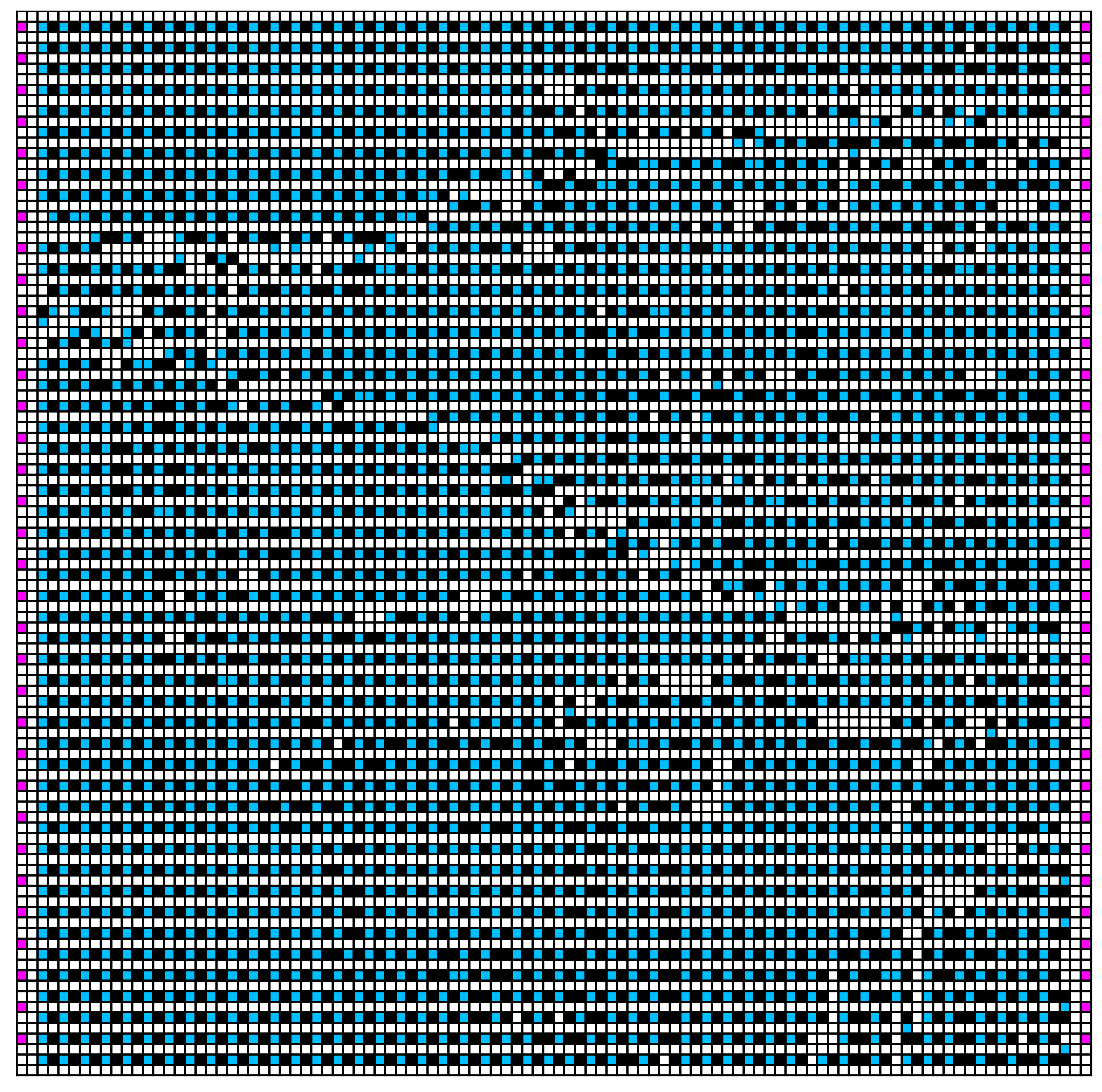}
            \caption{CMA-MAE ($\alpha = 1$)}
            \label{fig:warehouse-xxlarge-cma-mae-a=1-uneven}
        \end{subfigure}\\
        \begin{subfigure}[t]{1\textwidth}
            \centering
            \includegraphics[width=1\textwidth]{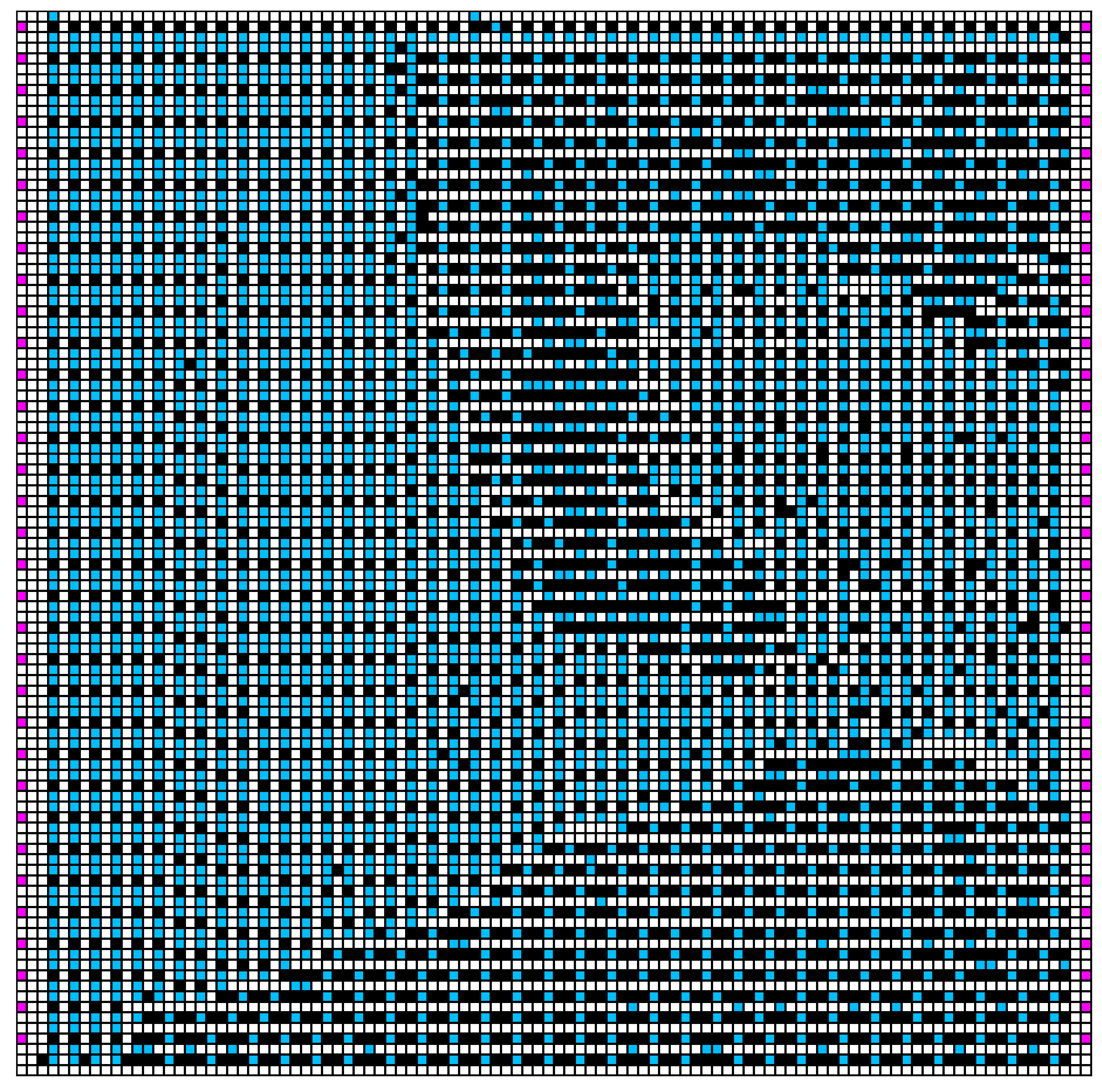}
            \caption{CMA-MAE ($\alpha = 5$)}
            \label{fig:warehouse-xxlarge-cma-mae-a=5-uneven}
        \end{subfigure}
    \end{minipage}
    \caption{NCA-generated warehouse environments of size $S_{eval}$.}
    \label{fig:warehouse-xxlarge-nca}
\end{figure}

\subsection{Implementation}
\label{appen:implementation}

We implement CMA-MAE with Pyribs~\cite{pyribs}, the NCA generators in PyTorch~\cite{Paszke2019PyTorchAI}, and the MILP solver with IBM's CPLEX library~\cite{ibm_cplex} in Python.

\noindent \textbf{Compute Constraint of MILP.} In the CPLEX MILP solver, we constrain the maximum number of threads to 1 and 28 for environments of size $S$ and $S_{eval}$, respectively. We set the deterministic runtime limit to be $6e4$ ticks for size $S$ and $3.6e6$ ticks for size $S_{eval}$ in CPLEX. The deterministic runtime limit in CPLEX is a metric that provides a consistent measure on the ``effort'' made by the MILP solver, measured in ``ticks'', a CPLEX abstract unit.

\section{Baseline and NCA-generated Environments}
\label{appen:env}

\subsection{NCA-generated Environments} \label{appen:opt-env}

\begin{figure}
    \centering
    \begin{subfigure}[t]{\EvalEnvShowSize\textwidth}
        \includegraphics[width=1\textwidth]{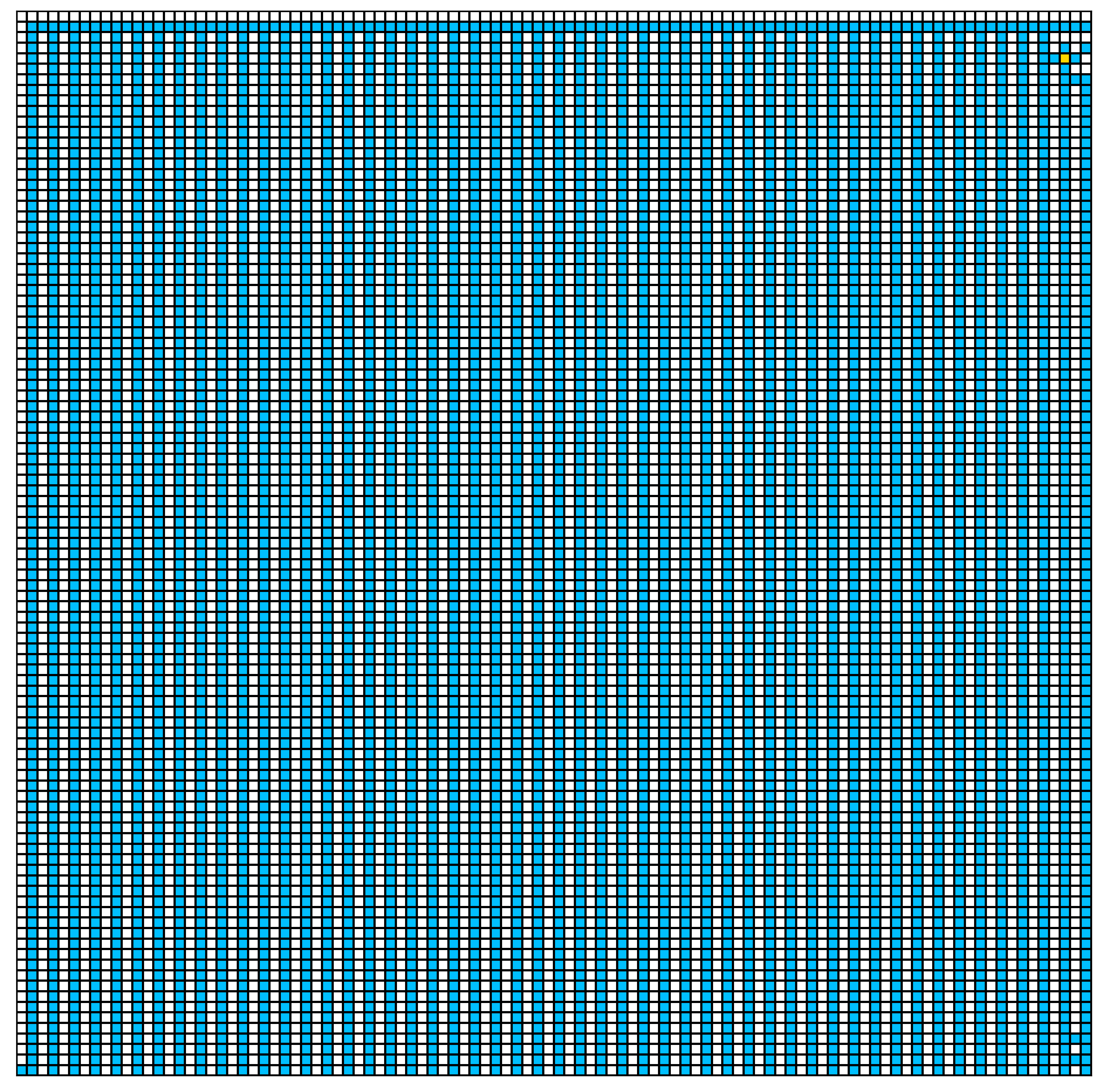}
        \caption{Unrepaired.}
        \label{fig:manufacture-xxlarge-nca-unrepaired}
    \end{subfigure}
    \hfill
    \begin{subfigure}[t]{\EvalEnvShowSize\textwidth}
        \includegraphics[width=1\textwidth]{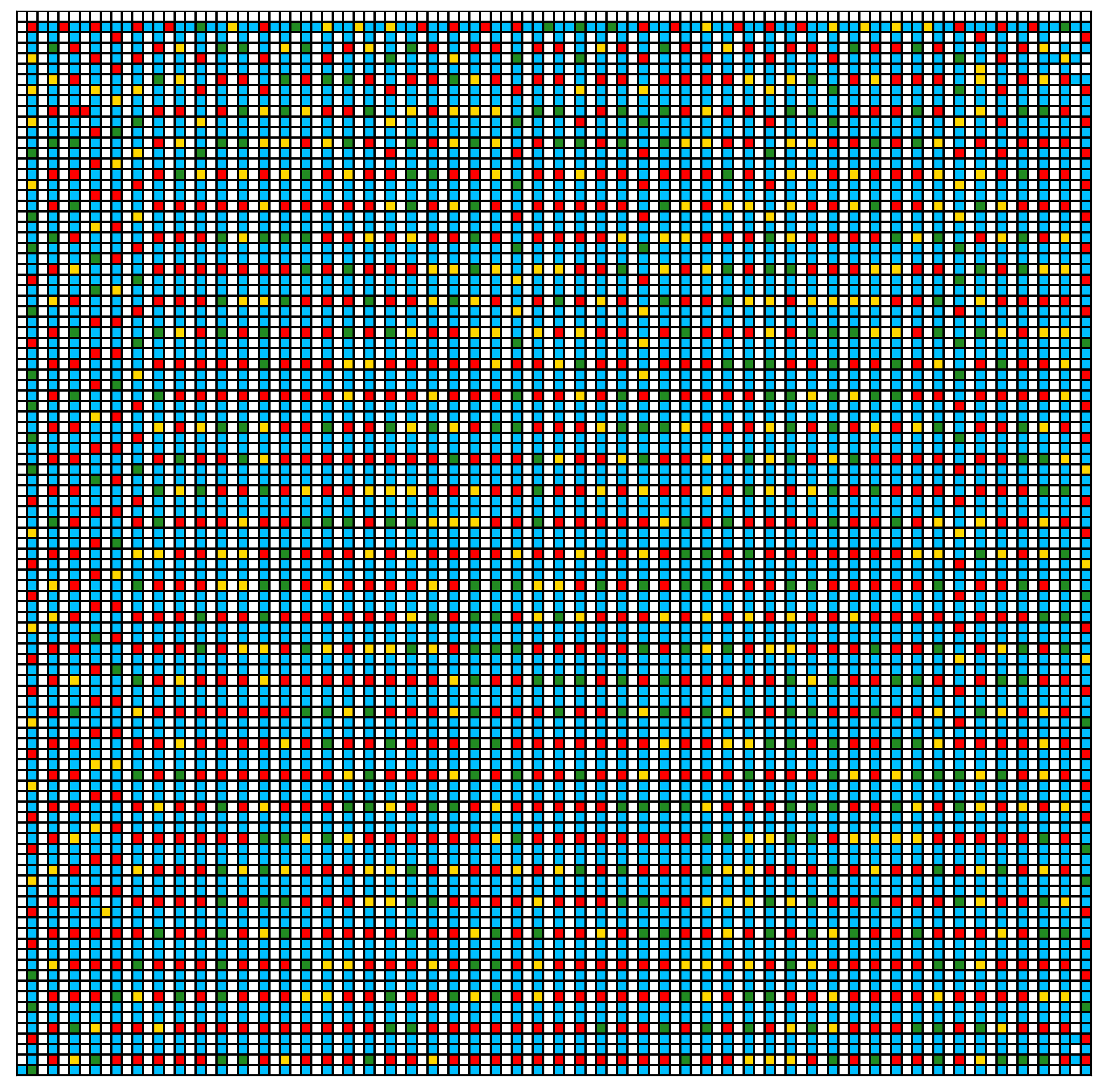}
        \caption{Repaired.}
        \label{fig:manufacture-xxlarge-nca-repaired}
    \end{subfigure}
    \caption{NCA-generated manufacturing environment of size $S_{eval}$ with CMA-MAE ($\alpha = 5$).}
    \label{fig:manufacture-xxlarge-nca}
\end{figure}

\begin{figure}
    \begin{subfigure}[t]{\EvalEnvShowSize\textwidth}
        \centering
        \includegraphics[width=1\textwidth]{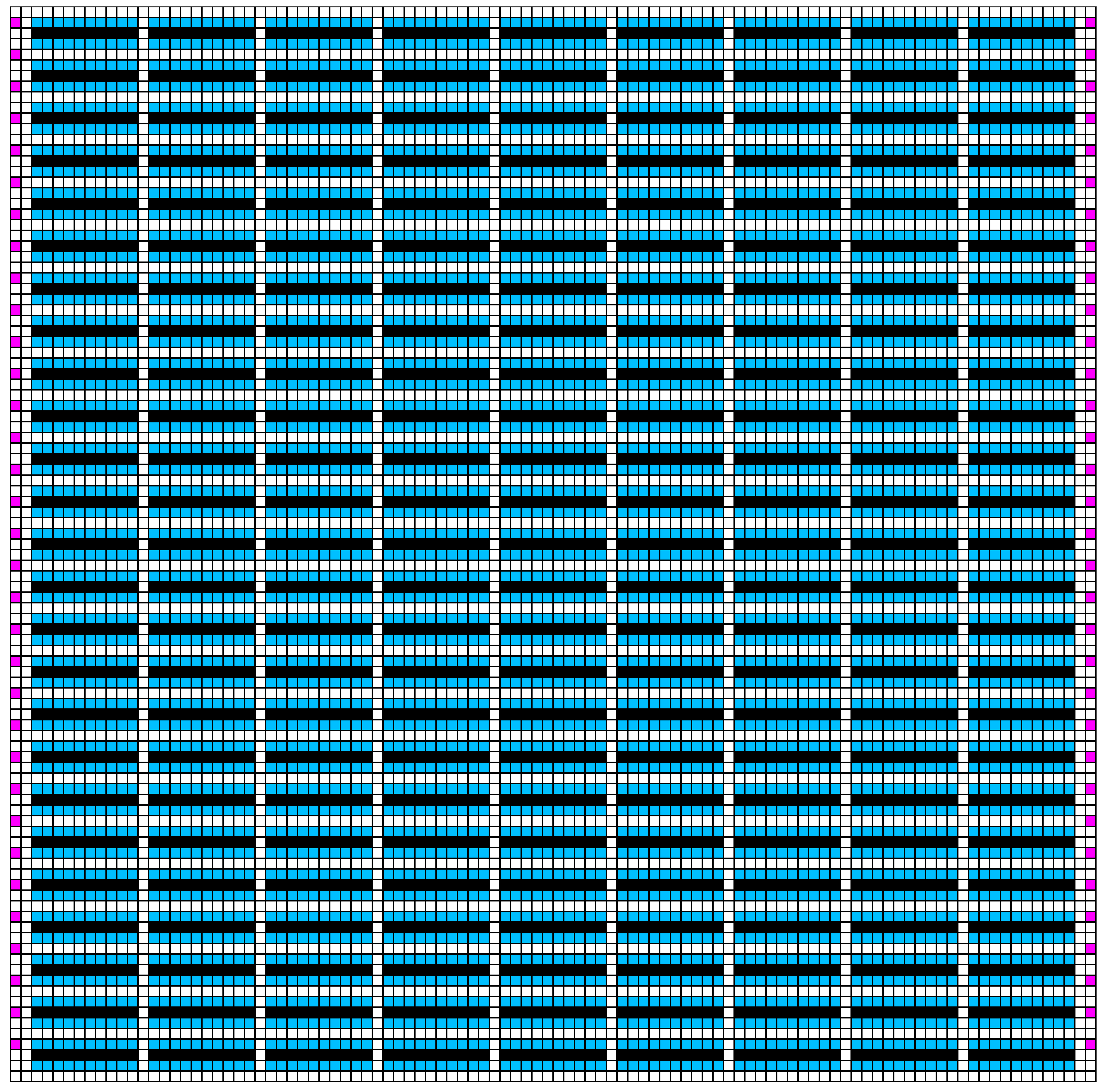}
        \caption{Warehouse}
        \label{fig:warehouse-xxlarge-human}
    \end{subfigure}
    \hfill
    \begin{subfigure}[t]{\EvalEnvShowSize\textwidth}
        \centering
        \includegraphics[width=1\textwidth]{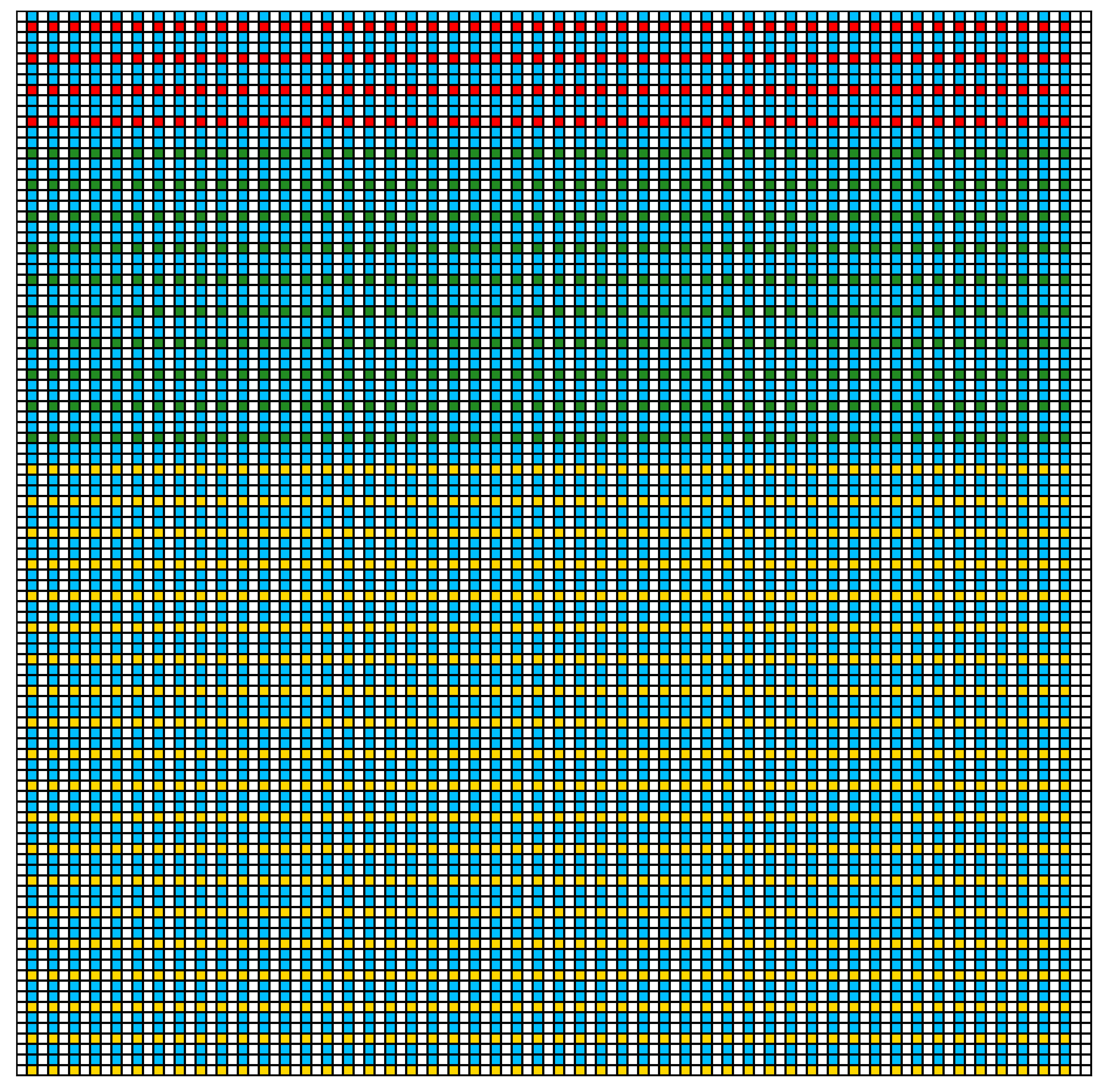}
        \caption{Manufacturing}
        \label{fig:manufacture-xxlarge-human}
    \end{subfigure}
    \caption{Human-designed warehouse and manufacturing environments of size $S_{eval}$.}
    \label{fig:xxlarge-human}
\end{figure}

\begin{figure}
    \begin{subfigure}[t]{\EvalEnvShowSize\textwidth}
        \centering
        \includegraphics[width=1\textwidth]{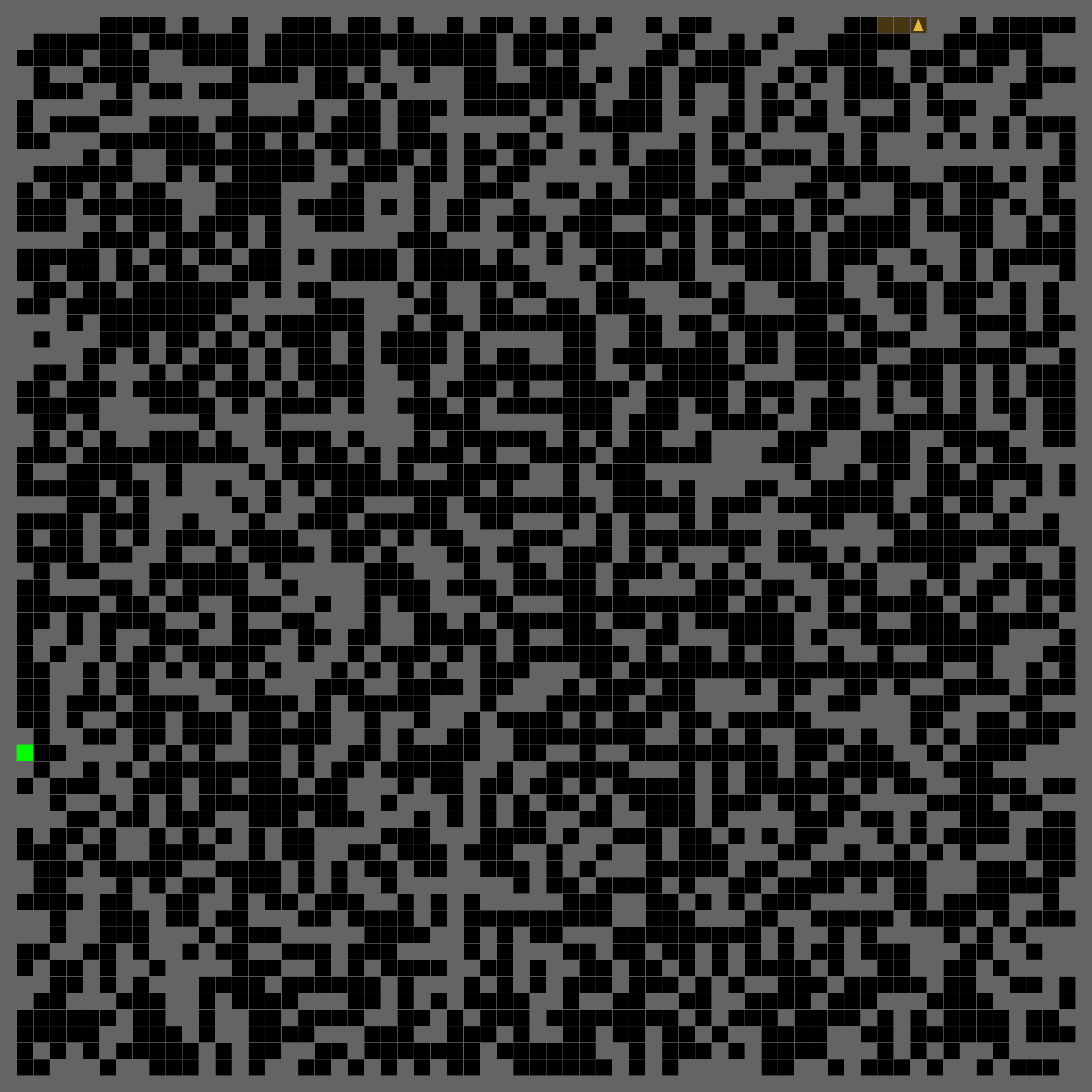}
    \end{subfigure}
    \hfill
    \begin{subfigure}[t]{\EvalEnvShowSize\textwidth}
        \centering
        \includegraphics[width=1\textwidth]{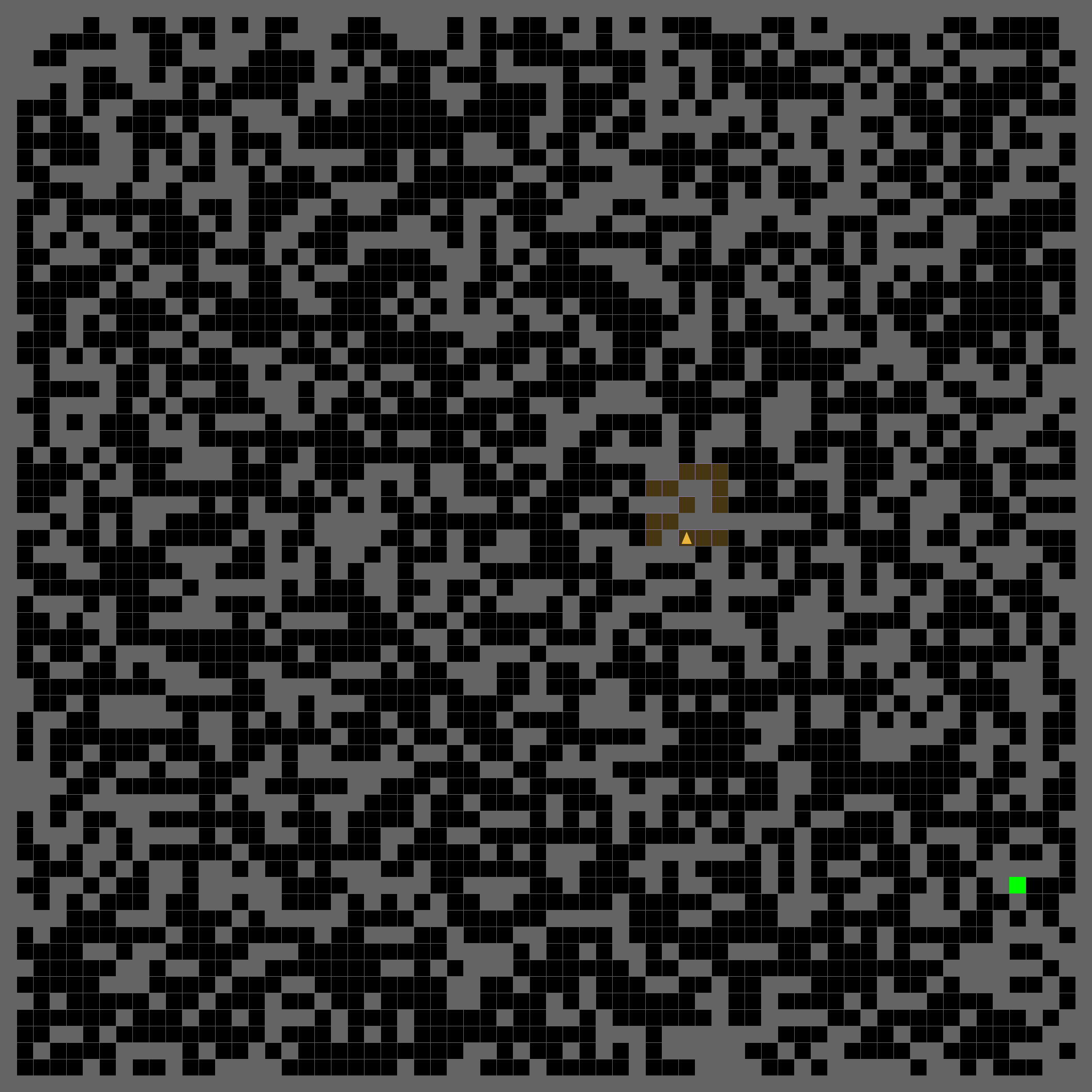}
    \end{subfigure}
    \caption{Baseline maze environments of size $S_{eval}$.}
    \label{fig:xxlarge-maze-baseline}
\end{figure}

\noindent \textbf{Warehouse.} \Cref{fig:warehouse-large-cma-mae-a=0-even,fig:warehouse-large-cma-mae-a=1-even,fig:warehouse-large-cma-mae-a=5-even,fig:warehouse-large-cma-mae-a=0-uneven,fig:warehouse-large-cma-mae-a=1-uneven,fig:warehouse-large-cma-mae-a=5-uneven} show the NCA-generated warehouse environments of size $S$. \Cref{fig:warehouse-xxlarge-nca} shows those of size $S_{eval}$. For the environments of warehouse (even), larger $\alpha$ values lead the optimal NCA generators in the result archives to generate environments with clearer patterns. Specifically, with $\alpha = 0$ (\Cref{fig:warehouse-large-cma-mae-a=0-even}), the environment has a large area of empty spaces at the bottom. With $\alpha = 1$ (\Cref{fig:warehouse-large-cma-mae-a=1-even}), the area of empty spaces is gone but the distribution of shelves is not even. With $\alpha = 5$ (\Cref{fig:warehouse-large-cma-mae-a=5-even}), the generated environment is mostly filled with repeated blocks of 1 $\times$ 2 shelves and the space in between the block is filled with endpoints. The trend is similar for environments of size $S_{eval}$ in the warehouse (even) domain (\Cref{fig:warehouse-xxlarge-cma-mae-a=0-even,fig:warehouse-xxlarge-cma-mae-a=1-even,fig:warehouse-xxlarge-cma-mae-a=5-even}). With $\alpha = 5$, the generated environment possesses the most regularized pattern.

For the warehouse (uneven) domain, larger $\alpha$ values do not always generate clearer patterns. Specifically, with size $S$, the environments generated with $\alpha = 0$ (\Cref{fig:warehouse-large-cma-mae-a=0-uneven}) and $\alpha = 5$ (\Cref{fig:warehouse-large-cma-mae-a=5-uneven}) have similar patterns. Both environments have fewer shelves on the left part of the environments and cluster more shelves on the right so that the agents have more traversable tiles to resolve conflicts near the more frequently visited workstations on the left border. Such pattern can also be found in size $S_{evals}$ (\Cref{fig:warehouse-xxlarge-cma-mae-a=0-uneven,fig:warehouse-xxlarge-cma-mae-a=5-uneven}). Notably, however, with size $S_{eval}$, the environment with $\alpha = 5$ has more traversable tiles near the workstations on the left border than that with $\alpha = 0$, resulting in higher success rate as shown in \Cref{tab:numerical-result} of \Cref{sec:result}.

\noindent \textbf{Manufacturing.} \Cref{fig:manufacture-large-cma-mae-a=5-opt,fig:manufacture-large-cma-mae-a=5-comp-dsage} show the NCA-generated environments of size $S$. The one generated by the optimal NCA generator (\Cref{fig:manufacture-large-cma-mae-a=5-opt}) possesses more regularized patterns than the one used to compare with DSAGE (\Cref{fig:manufacture-large-cma-mae-a=5-comp-dsage}). 
However, in their corresponding unrepaired environments (\Cref{fig:manufacture-large-cma-mae-a=5-comp-dsage-unrepaired,fig:manufacture-large-cma-mae-a=5-opt-unrepaired}), the NCA generators only generate empty spaces and endpoints and rely on the MILP solver to put the workstations in the environment. We conjecture that for the manufacturing domain, the number of endpoints is more important than the ratio between different types of workstations because more endpoints allow the agents to perform tasks in parallel around the workstations. Similarly, \Cref{fig:manufacture-xxlarge-nca} shows the unrepaired and repaired NCA-generated environments of size $S_{eval}$. The unrepaired environment (\Cref{fig:manufacture-xxlarge-nca-unrepaired}) has no workstations at all, while the repaired environment has randomly placed workstations by the MILP solver.

\subsection{Baseline Environments} \label{appen:baseline-env}

\noindent \textbf{Warehouse.} \Cref{fig:warehouse-large-human} shows the human-designed warehouse environment of size $S$ for both variants of warehouse domains. It is taken from the previous works~\cite{LiuAAMAS19,Li2020LifelongMP,zhangLayout23}, where 1 $\times$ 10 blocks of shelves are repeatedly placed, and each shelf is surrounded by two endpoints from the top and bottom of it. We scale this pattern to create the human-designed warehouse environment of size $S_{eval}$, shown in \Cref{fig:warehouse-xxlarge-human}. \Cref{fig:warehouse-large-dsage-even,fig:warehouse-large-dsage-uneven} shows the DSAGE-optimized warehouse environments of size $S$ for warehouse (even) and warehouse (uneven), respectively. Both of them have no clear regularized patterns.

\noindent \textbf{Manufacturing.} \Cref{fig:manufacture-large-human,fig:manufacture-xxlarge-human} show the human-designed manufacturing environments of size $S$ and $S_{eval}$, respectively. We create the human-designed environment of size $S$ (\Cref{fig:manufacture-large-human}) such that (1) the number of workstations mirrors that in the DSAGE-optimized environment so that approximately the same number of agents can operate in parallel around the workstations, and (2) the ratio of different types of stations approximates $t_r : t_g : t_y = 2 : 5 : 10$ so that the environment has more workstations that require agents to stay longer. 

To create these human-designed environments, we draw insights from some NCA-generated environments. Notably, the warehouse (even) environment with $\alpha = 5$ (\Cref{fig:warehouse-large-cma-mae-a=5-even}) has the best scalability (shown in \Cref{sec:result}). This informs the design of the human-designed manufacturing environment of size $S$, displayed in \Cref{fig:manufacture-large-human}, where we evenly distribute blocks of workstations and position endpoints around each for parallel agents operation.

In the size of $S_{eval}$, the NCA-generated manufacturing environment (\Cref{fig:manufacture-xxlarge-nca-repaired}) features standalone workstations surrounded by endpoints. We imitate this pattern to create the human-designed manufacturing environment of size $S_{eval}$, shown in \Cref{fig:manufacture-xxlarge-human}. Similar to the human-designed environment of size $S$, we additionally let the ratio of different types of workstations to approximate $t_r : t_g : t_y = 2 : 5 : 10$ and the total number of workstations are similar.

Given these human-designed manufacturing environments have NCA-inspired designs, we anticipate them to outperform their counterparts in the warehouse domain, compared to the NCA-generated ones.

\noindent \textbf{Maze.} \Cref{fig:xxlarge-maze-baseline} shows two example baseline environments using the method described \Cref{sec:result}. Compared to the NCA-generated maze environment of size $S_{eval}$ shown in \Cref{fig:maze-xxlarge}, both of them have no clear regularized patterns. Running the ACCEL agent in these two environments for 100 times results in success rates of 0\% and 8\%, respectively.

\begin{figure}[!t]
    \centering
    \begin{minipage}[b]{0.13\textwidth}
        \centering
        \ 
    \end{minipage}
    \hfill
    \begin{minipage}[b]{0.27\textwidth}
        \centering
        $\alpha=0$
    \end{minipage}
    \hfill
    \begin{minipage}[b]{0.27\textwidth}
        \centering
        $\alpha=1$
    \end{minipage}
    \hfill
    \begin{minipage}[b]{0.27\textwidth}
        \centering
        $\alpha=5$
    \end{minipage}\\
    \begin{minipage}[b]{0.13\textwidth}
        \centering
        Warehouse (even)
        \vspace{1.4cm}
    \end{minipage}
    \hfill
    \begin{subfigure}[t]{0.27\textwidth}
        \includegraphics[width=1\textwidth]{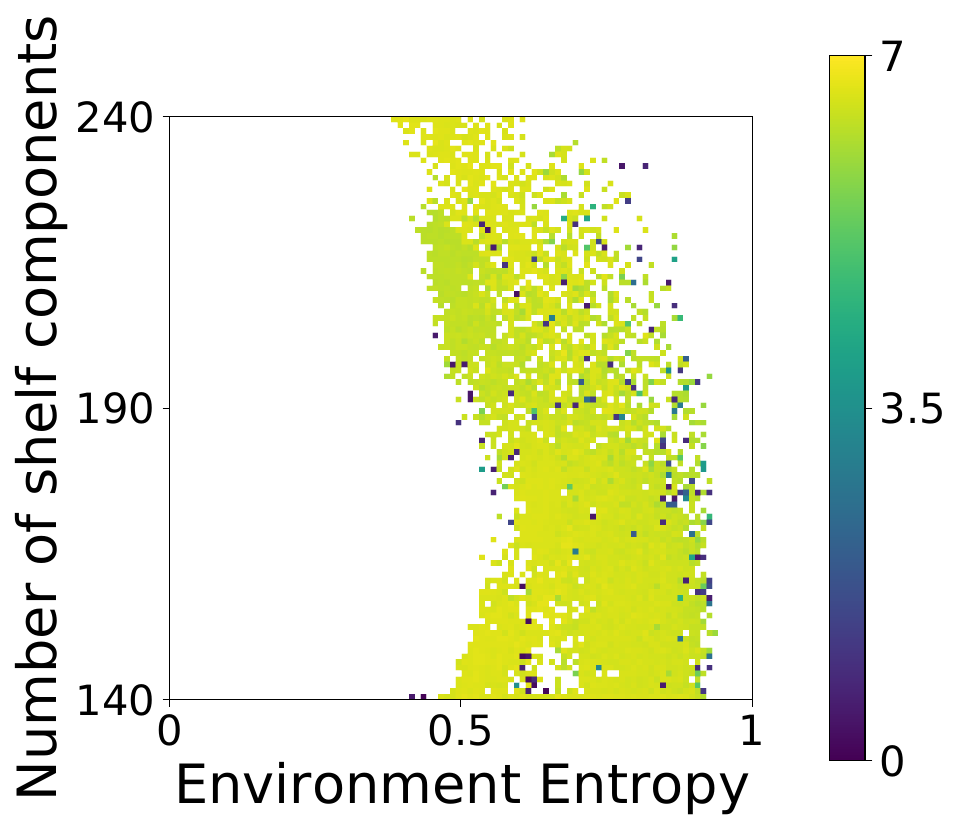}
    \end{subfigure}
    \hfill
    \begin{subfigure}[t]{0.27\textwidth}
        \includegraphics[width=1\textwidth]{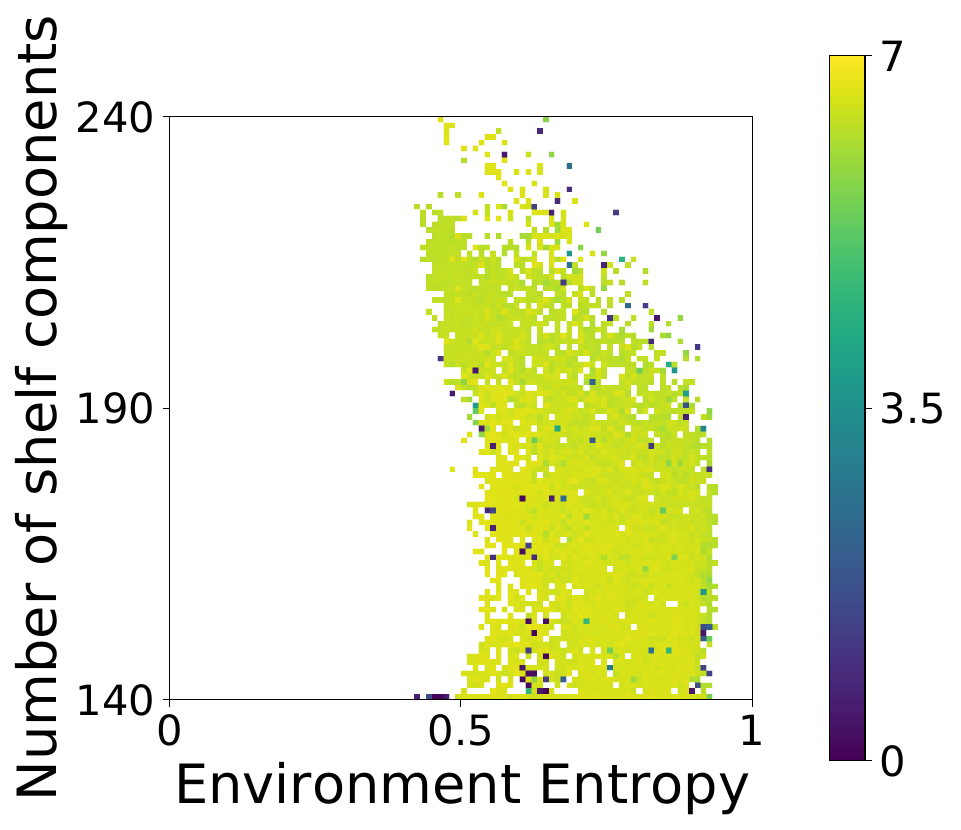}
    \end{subfigure}
    \hfill
    \begin{subfigure}[t]{0.27\textwidth}
        \includegraphics[width=1\textwidth]{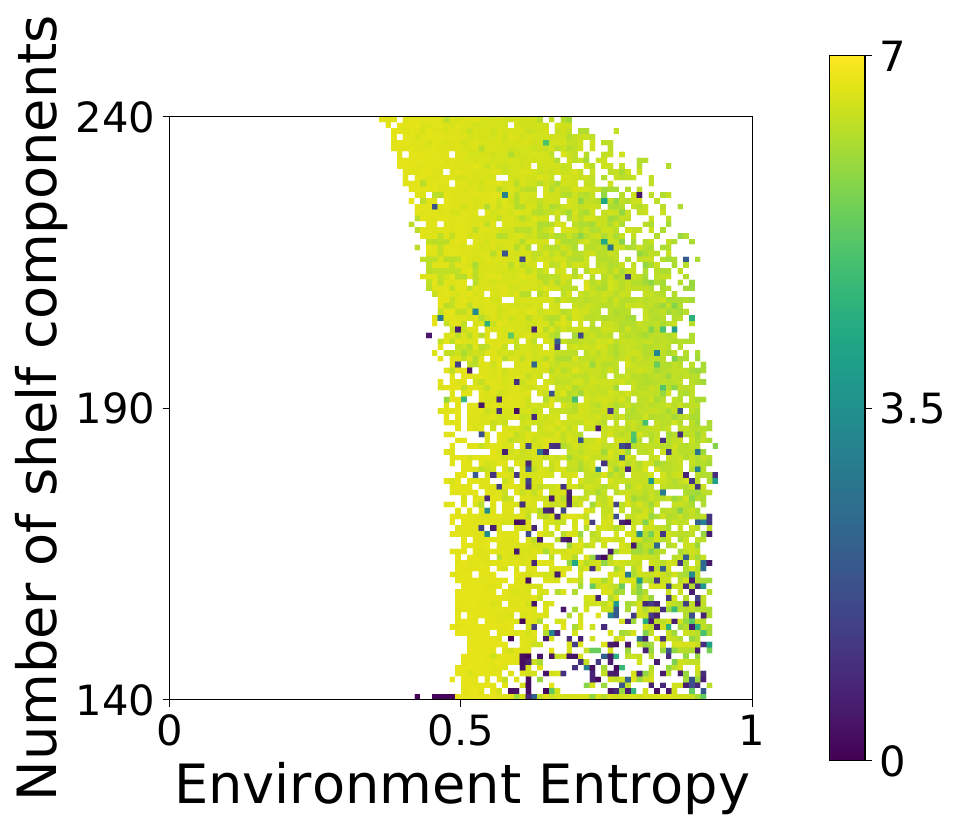}
    \end{subfigure}\\
    \begin{minipage}[b]{0.13\textwidth}
        \centering
        Warehouse (uneven)
        \vspace{1.4cm}
    \end{minipage}
    \hfill
    \begin{subfigure}[t]{0.27\textwidth}
        \includegraphics[width=1\textwidth]{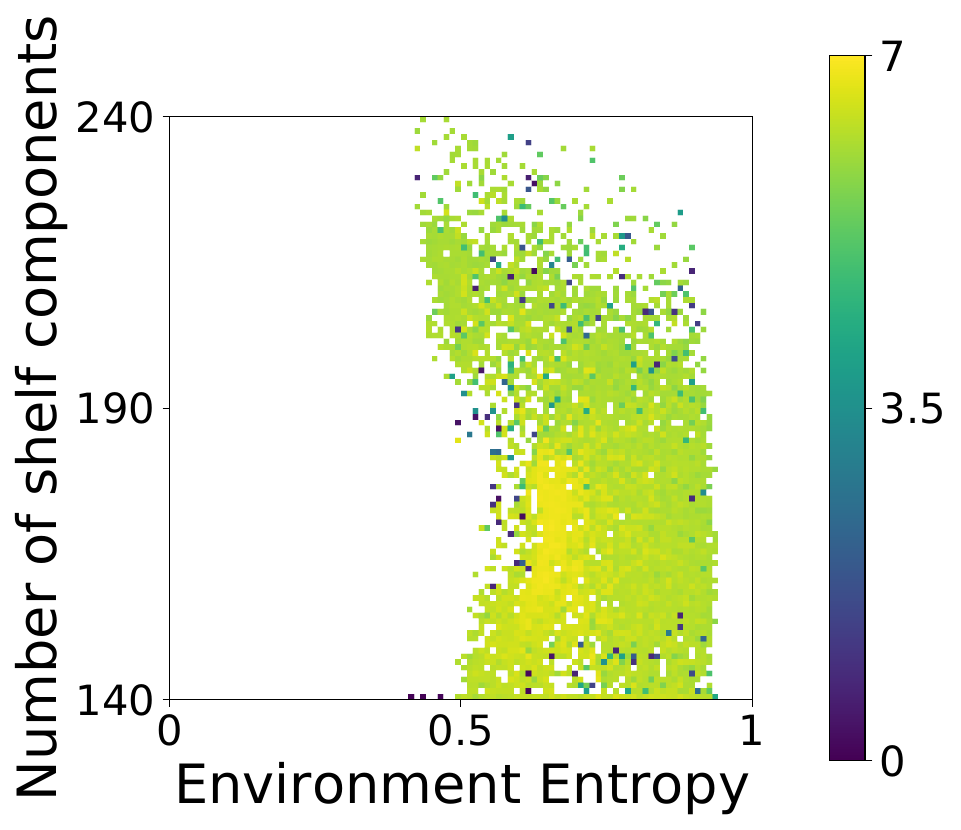}
    \end{subfigure}
    \hfill
    \begin{subfigure}[t]{0.27\textwidth}
        \includegraphics[width=1\textwidth]{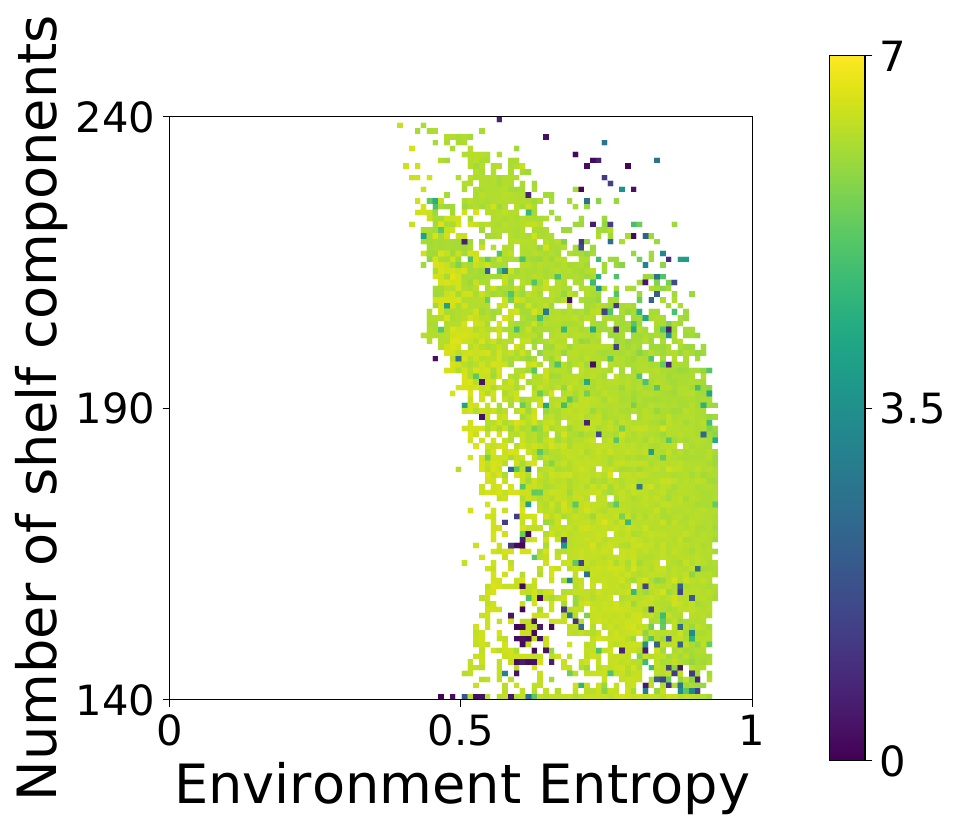}
    \end{subfigure}
    \hfill
    \begin{subfigure}[t]{0.27\textwidth}
        \includegraphics[width=1\textwidth]{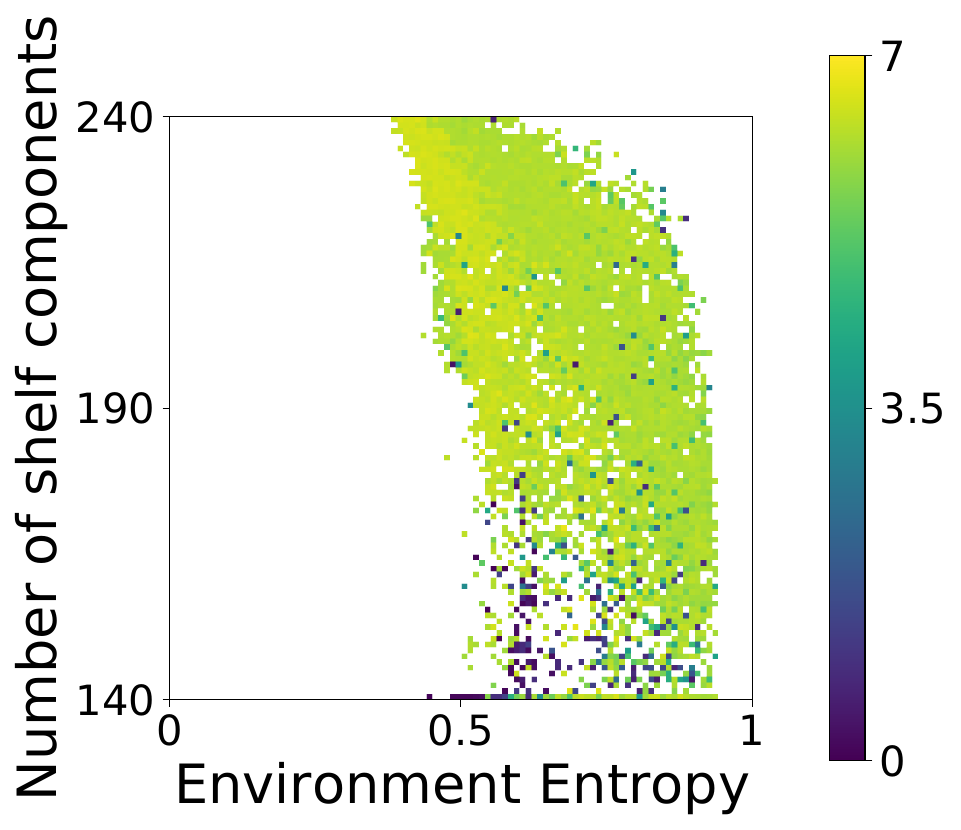}
    \end{subfigure}\\
    \caption{Example result archives of CMA-MAE + NCA in warehouse domains with different $\alpha$ values.}
    \label{fig:archive-a-vals}
\end{figure}

\section{Additional Results} \label{appen:add-result}

In this section, we include the following additional results:
(1) we compare the result archives with different $\alpha$ values to discuss the effect of $\alpha$ on patterns,
(2) we show the number of finished tasks over time in the NCA-generated and baseline environments of size $S$ in the warehouse and manufacturing domains, analyzing the occurrence of congestion in these environments, 
(3) we show the scalability of the trained NCA generators in environment sizes other than $S$ and $S_{eval}$,
(4) we compare our method with an additional baseline of tiling environments of size $S$ to create those of size $S_{eval}$,
(5) we show the QD-score and archive coverage of CMA-MAE + NCA comparing with MAP-Elites~\cite{mouret2015illuminating,vassiliades2018isoline} + NCA to demonstrate the advantage of CMA-MAE in terms of training a diverse collection of NCA generators, 
(6) we show the QD-score and archive coverage of CMA-MAE + NCA comparing with DSAGE with direct tile search~\cite{Bhatt2022DeepSA,zhangLayout23}  to demonstrate the advantage of NCA in terms of generating environments with regularized patterns, and
(7) we compare CMA-MAE with a derivative-free single-objective optimizer CMA-ES~\cite{hansen2016cmaes} to demonstrate that optimizing without diversifying measures can more easily lead to local optima.

\subsection{On the Effect of \texorpdfstring{$\alpha$}{a} on Patterns} \label{appen:alpha-val}

\Cref{fig:archive-a-vals} shows the result archives of warehouse domains with different values of $\alpha$. We observe that a larger $\alpha$ value is correlated with better archive coverage in the low environment entropy area in the archive. Since a lower value of environment entropy corresponds to clearer regularized patterns and thus a possibly more scalable NCA generator, a reasonably large $\alpha$ value can give users more freedom in choosing trained NCA generators from the result archive.

\subsection{Number of Finished Tasks Over Time}
\label{appen:add-result:task_thr_time}

\begin{figure}[!t]
    \centering
    \includegraphics[width=0.9\textwidth]{figs/large_stats_legend.pdf}
    \begin{subfigure}[t]{0.3\textwidth}
        \includegraphics[width=1\textwidth]{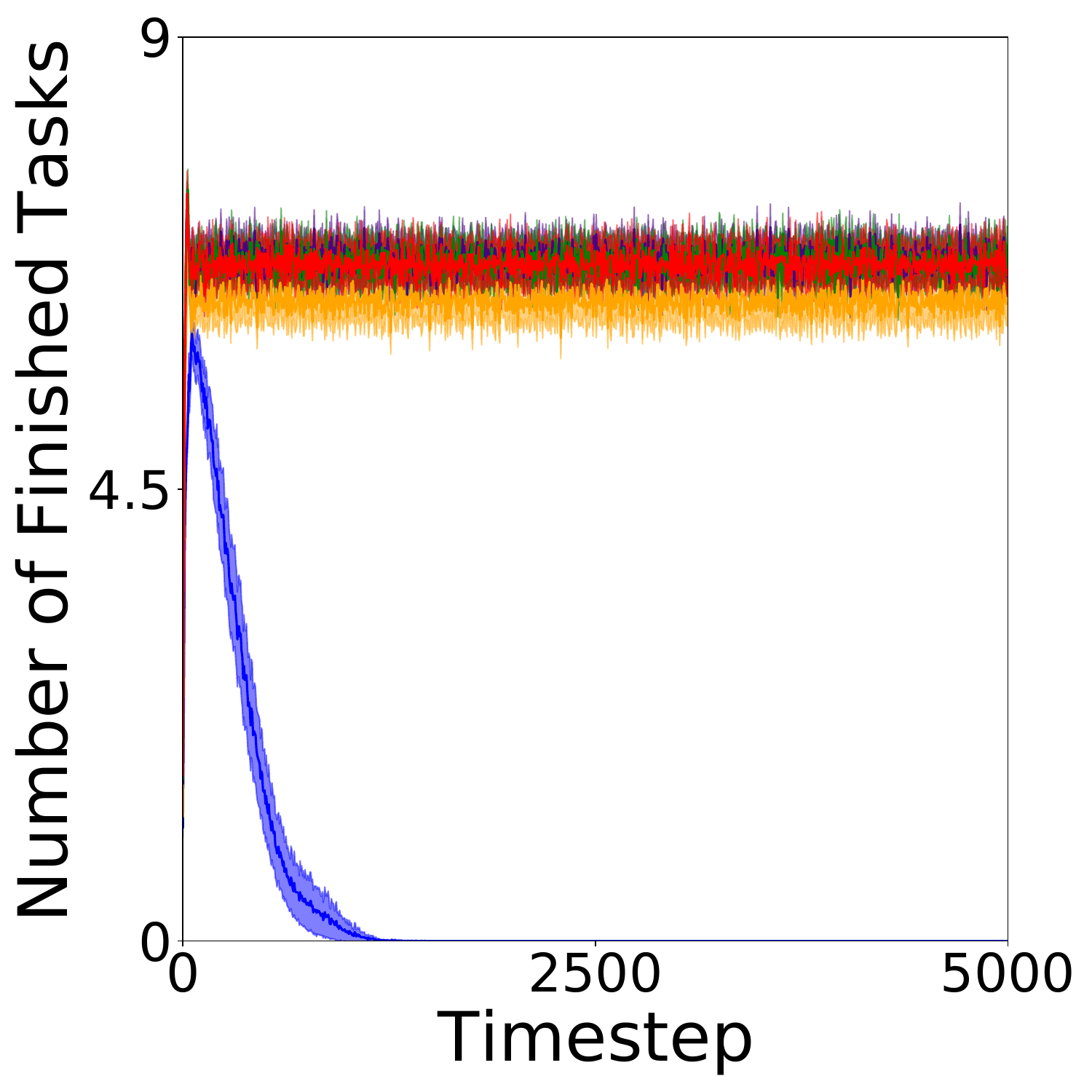}
        \caption{Warehouse (even)}
        \label{fig:warehouse-large-even-thr_time}
    \end{subfigure}
    \begin{subfigure}[t]{0.3\textwidth}
        \includegraphics[width=1\textwidth]{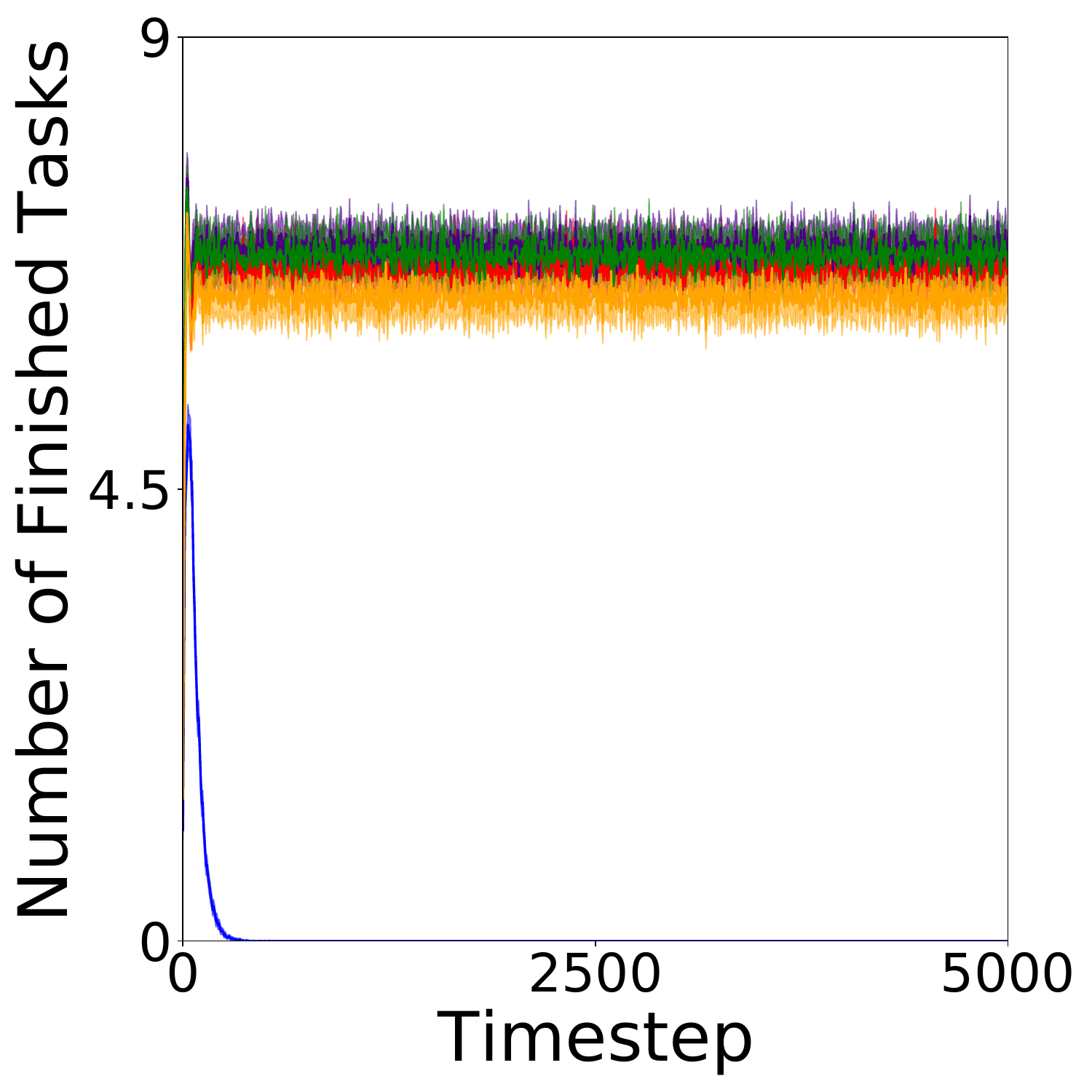}
        \caption{Warehouse (uneven)}
        \label{fig:warehouse-large-uneven-thr_time}
    \end{subfigure}
    \begin{subfigure}[t]{0.3\textwidth}
        \includegraphics[width=1\textwidth]{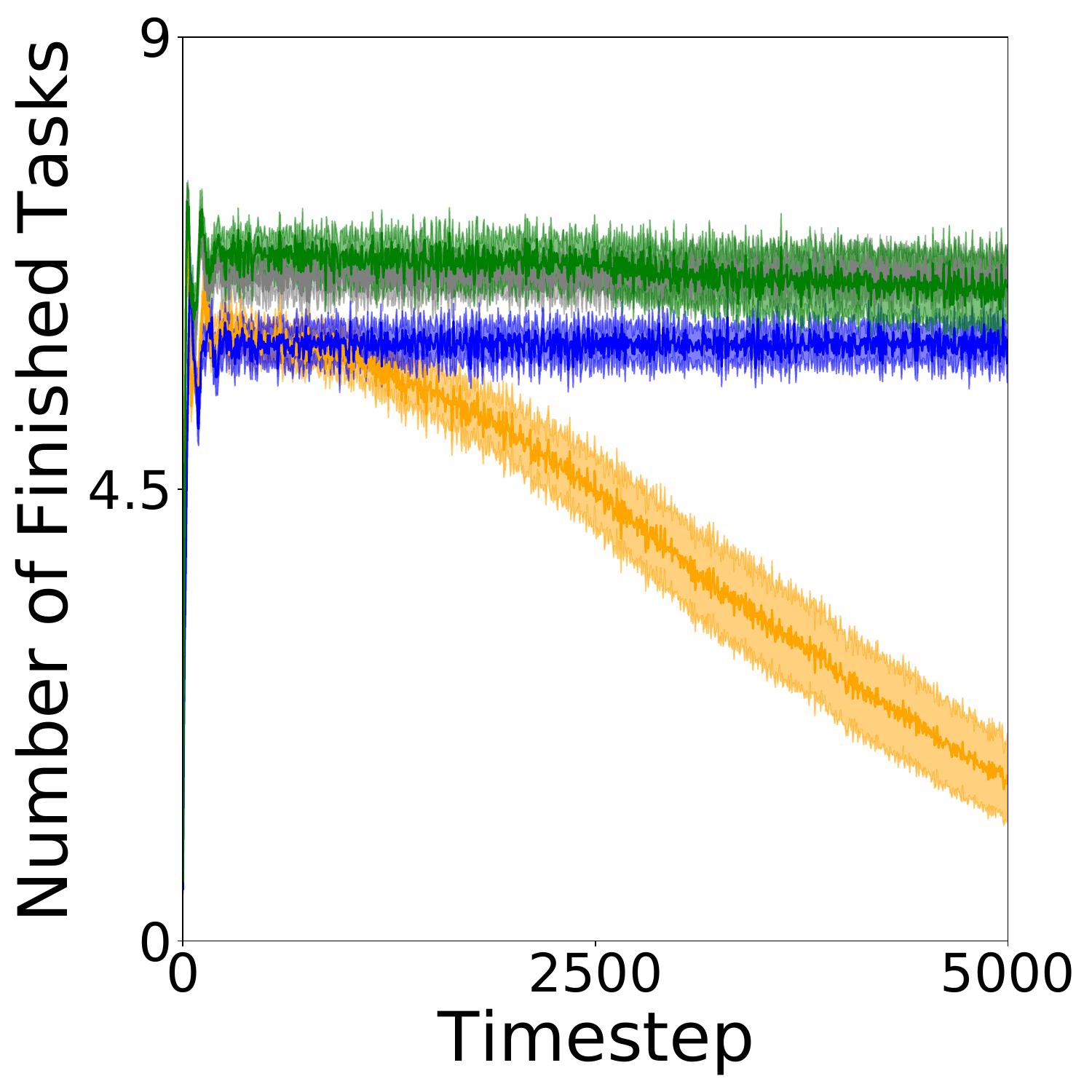}
        \caption{Manufacturing}
        \label{fig:manufacture-large-thr_time}
    \end{subfigure}
    \caption{Number of finished tasks per timestep with $N_a = 200$ agents. The solid lines are the average while the shaded area shows the 95\% confidence interval.}
    \label{fig:large-thr-time}
\end{figure}

To understand the gap in the performance of the environments in the warehouse and manufacturing domains, we run 100 simulations with $N_a$ agents in all of the environments of size $S$ and plot the number of finished tasks at each timestep. We do not stop the simulation even if the agents encounter congestion. \Cref{fig:large-thr-time} shows the number of finished tasks over 5,000 timesteps. All NCA-generated environments maintain a stable number of finished tasks throughout the simulation.

\begin{figure}[!t]
    \centering
    \includegraphics[width=0.65\textwidth]{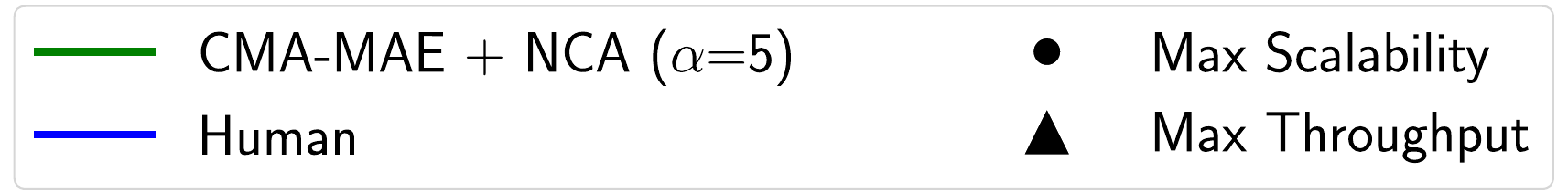}\\
    \begin{subfigure}[t]{0.337\textwidth}
        \includegraphics[width=1\textwidth]{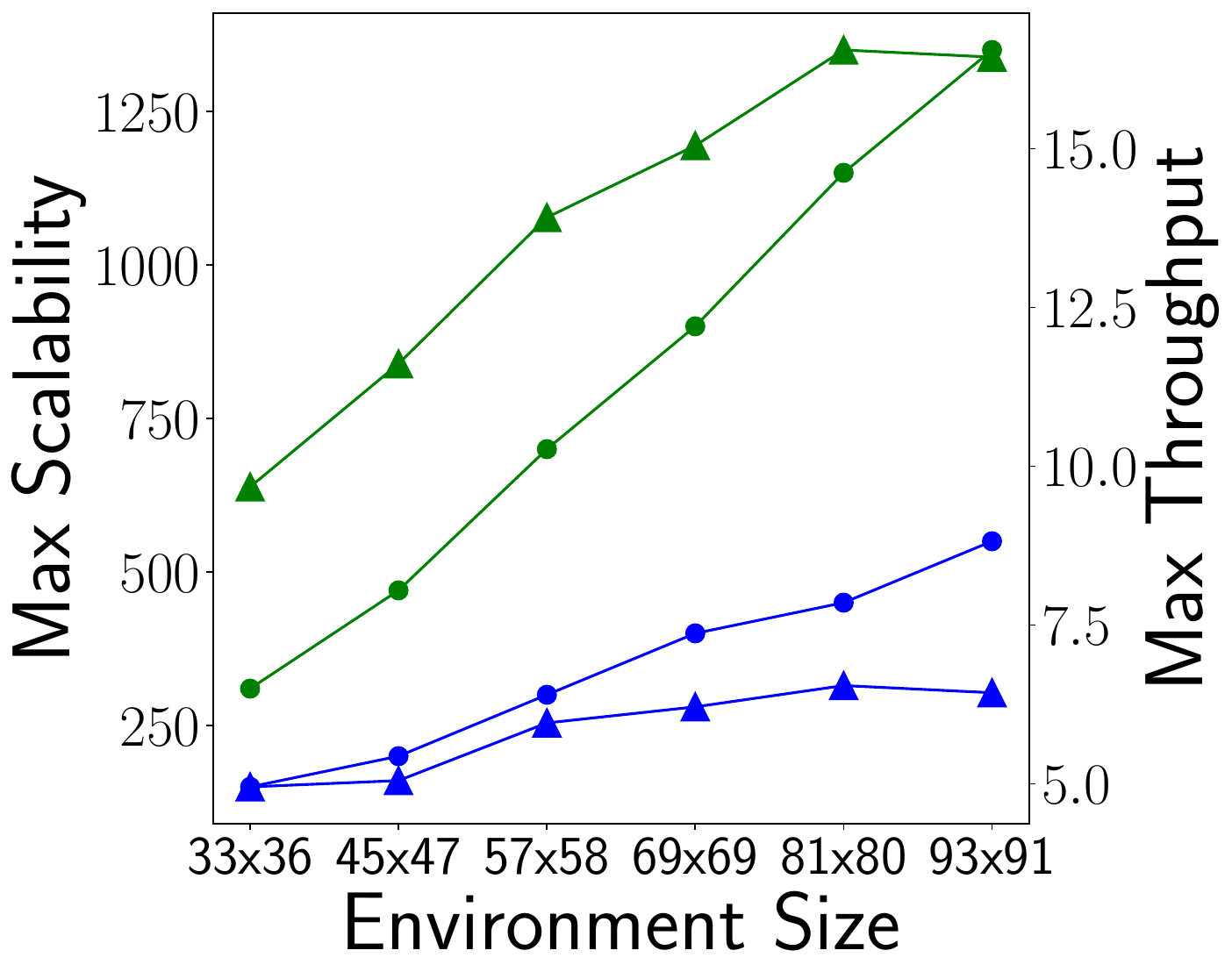}
        \caption{Warehouse (even)}
    \end{subfigure}
    \hfill
    \begin{subfigure}[t]{0.32\textwidth}
        \includegraphics[width=1\textwidth]{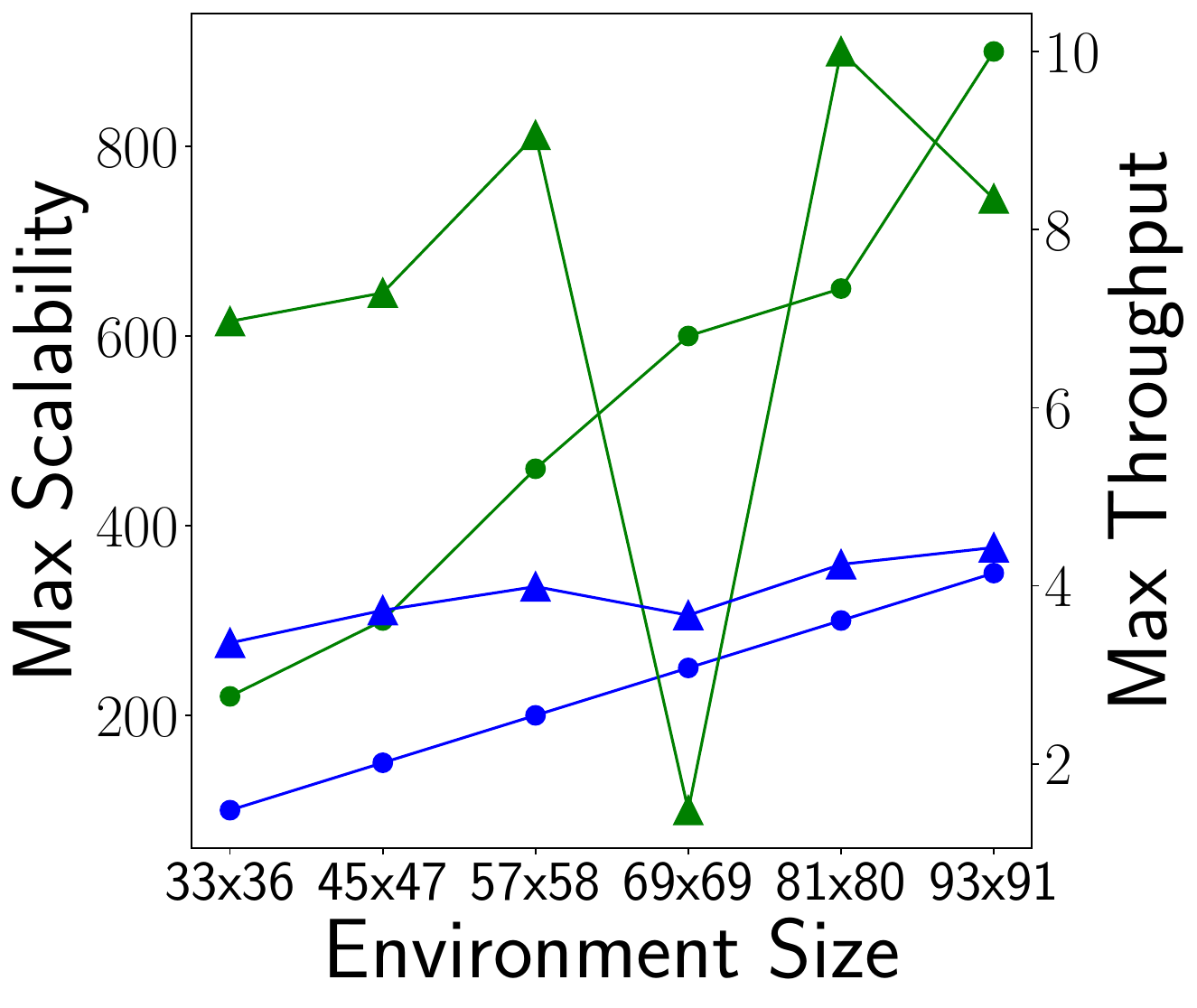}
        \caption{Warehouse (uneven)}
    \end{subfigure}
    \hfill
    \begin{subfigure}[t]{0.33\textwidth}
        \includegraphics[width=1\textwidth]{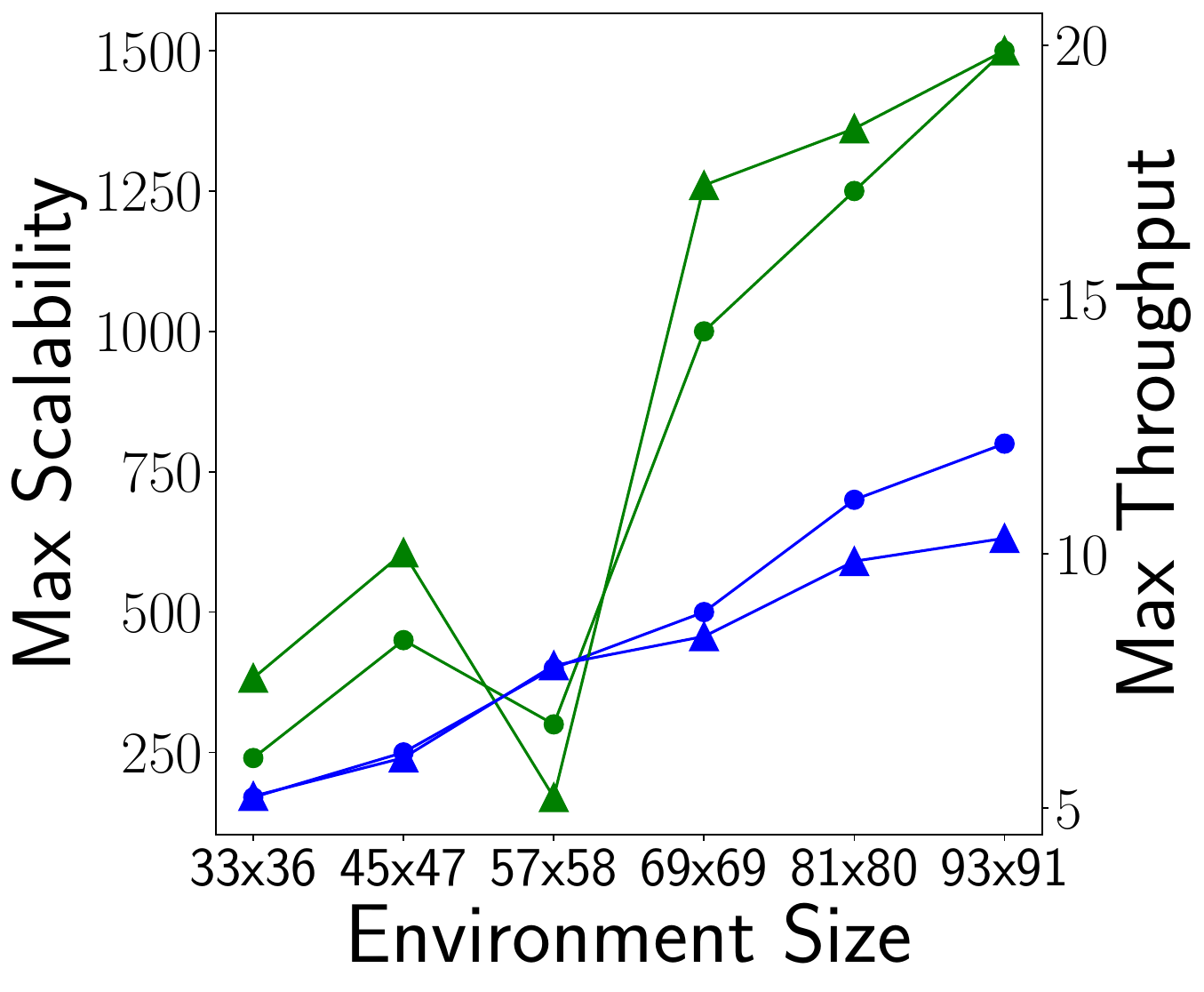}
        \caption{Manufacturing}
    \end{subfigure}
    \caption{Scalability of NCA-generated environments in the warehouse (even), warehouse (uneven), and manufacturing domains. To determine maximum scalability, we incrementally add agents and run 50 simulations, each with 5000 timesteps, for each count, monitoring average throughput. Once throughput drops, we identify the agent count corresponding to the peak mean throughput as the maximum scalability.}
    \label{fig:scalability}
\end{figure}

\subsection{Scalability in More Environment Sizes}
\label{appen:add-result:scale-env-size}

\begin{figure}
    \centering
    \begin{subfigure}[t]{\EvalEnvShowSize\textwidth}
        \centering
        \includegraphics[width=1\textwidth]{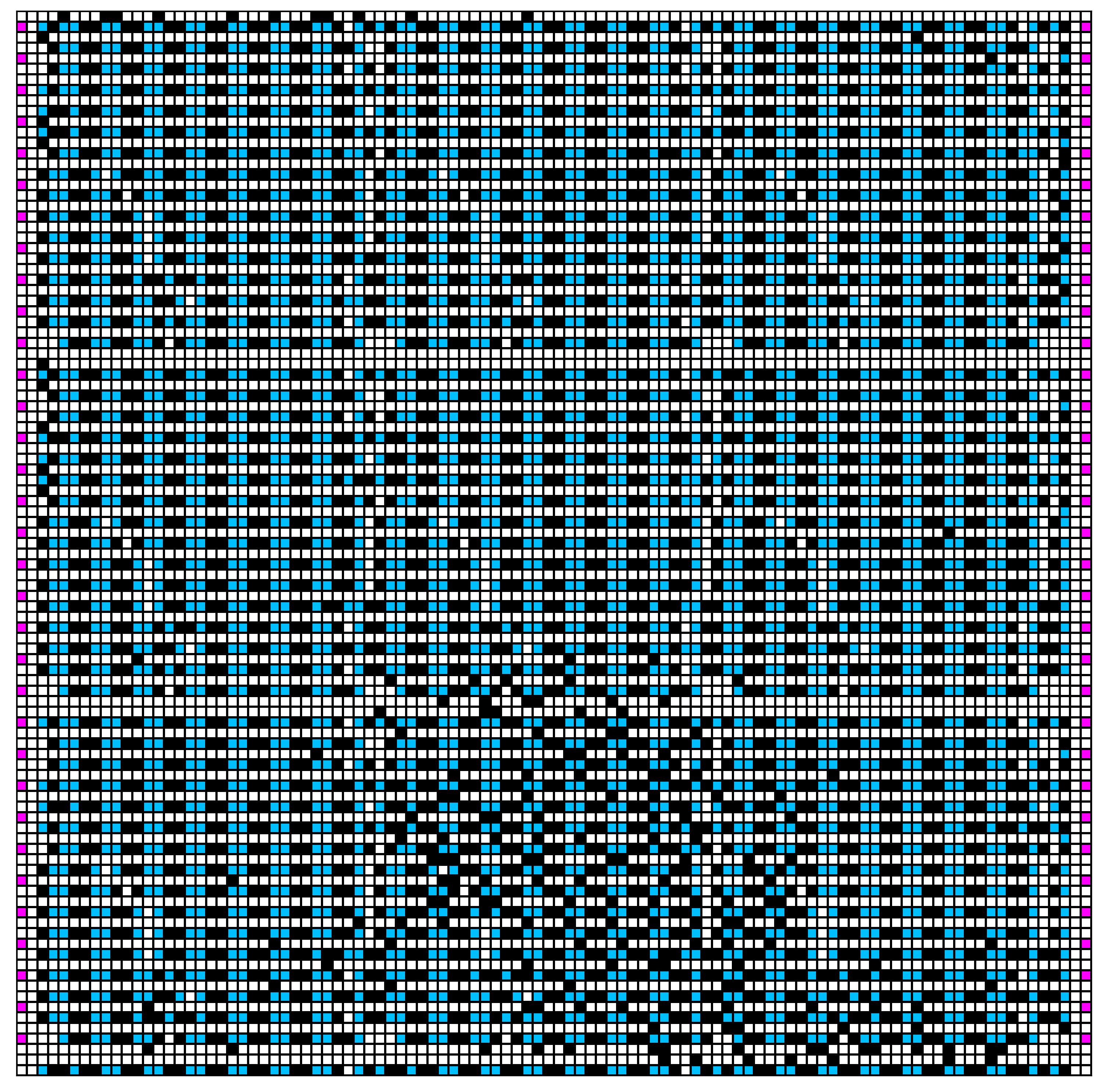}
        \caption{Warehouse (even) tiling environment}
        \label{fig:tile-baseline-warehouse-even}
    \end{subfigure}
    \hfill
    \begin{subfigure}[t]{\EvalEnvShowSize\textwidth}
        \centering
        \includegraphics[width=1\textwidth]{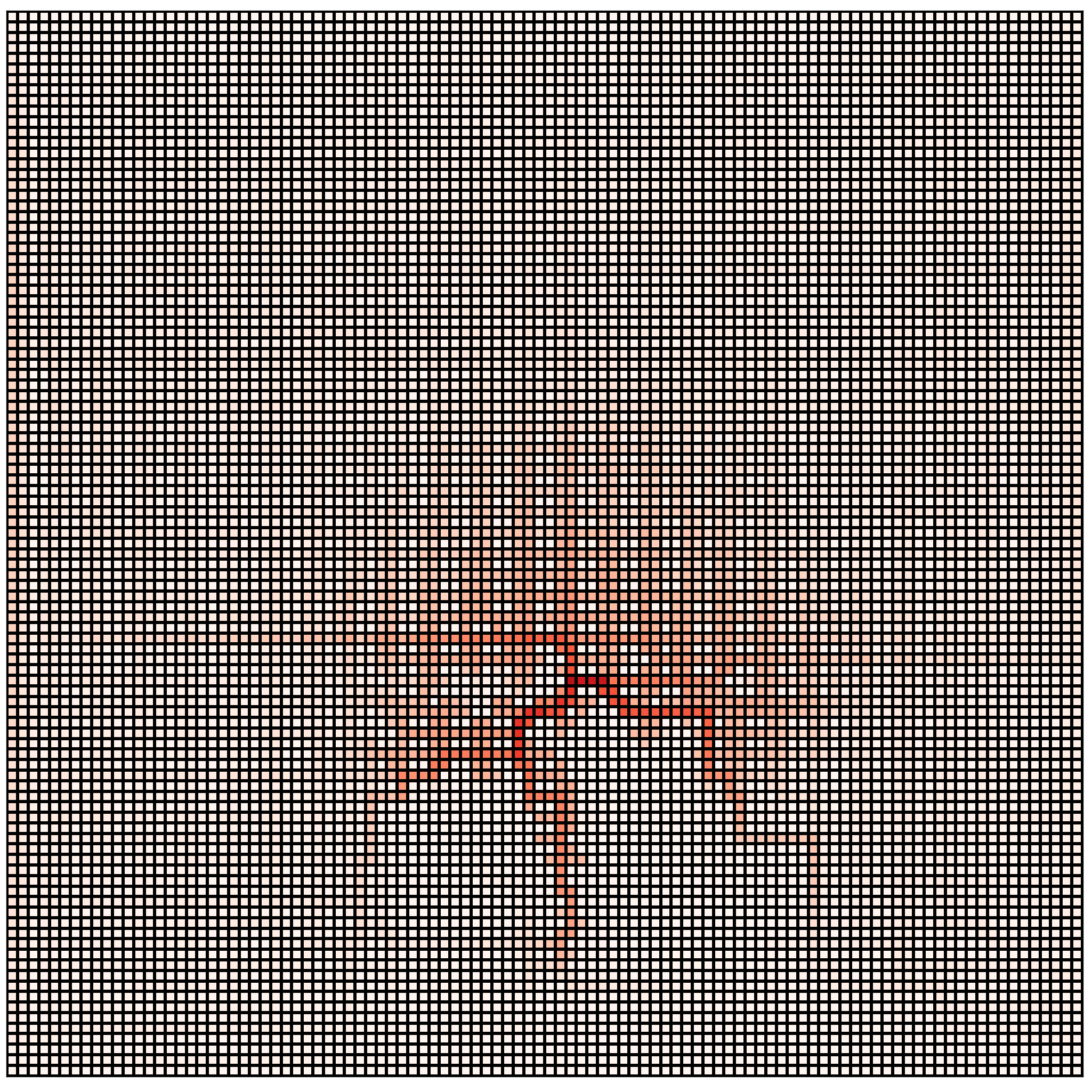}
        \caption{Warehouse (even) tile-usage map}
        \label{fig:tile-baseline-warehouse-even-tile-usage}
    \end{subfigure}\\
    \begin{subfigure}[t]{\EvalEnvShowSize\textwidth}
        \centering
        \includegraphics[width=1\textwidth]{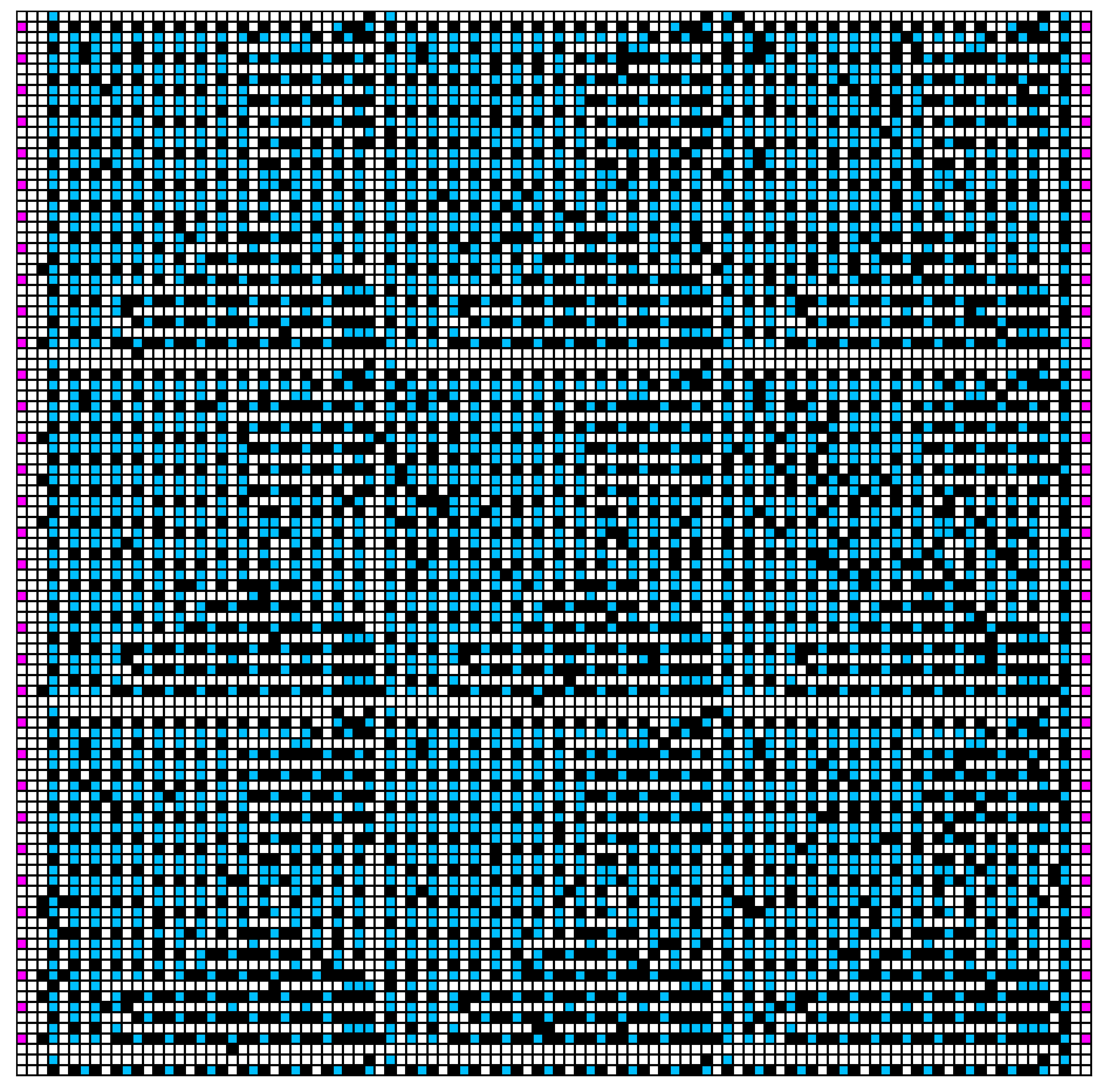}
        \caption{Warehouse (uneven) tiling environment}
        \label{fig:tile-baseline-warehouse-uneven}
    \end{subfigure}
    \hfill
    \begin{subfigure}[t]{\EvalEnvShowSize\textwidth}
        \centering
        \includegraphics[width=1\textwidth]{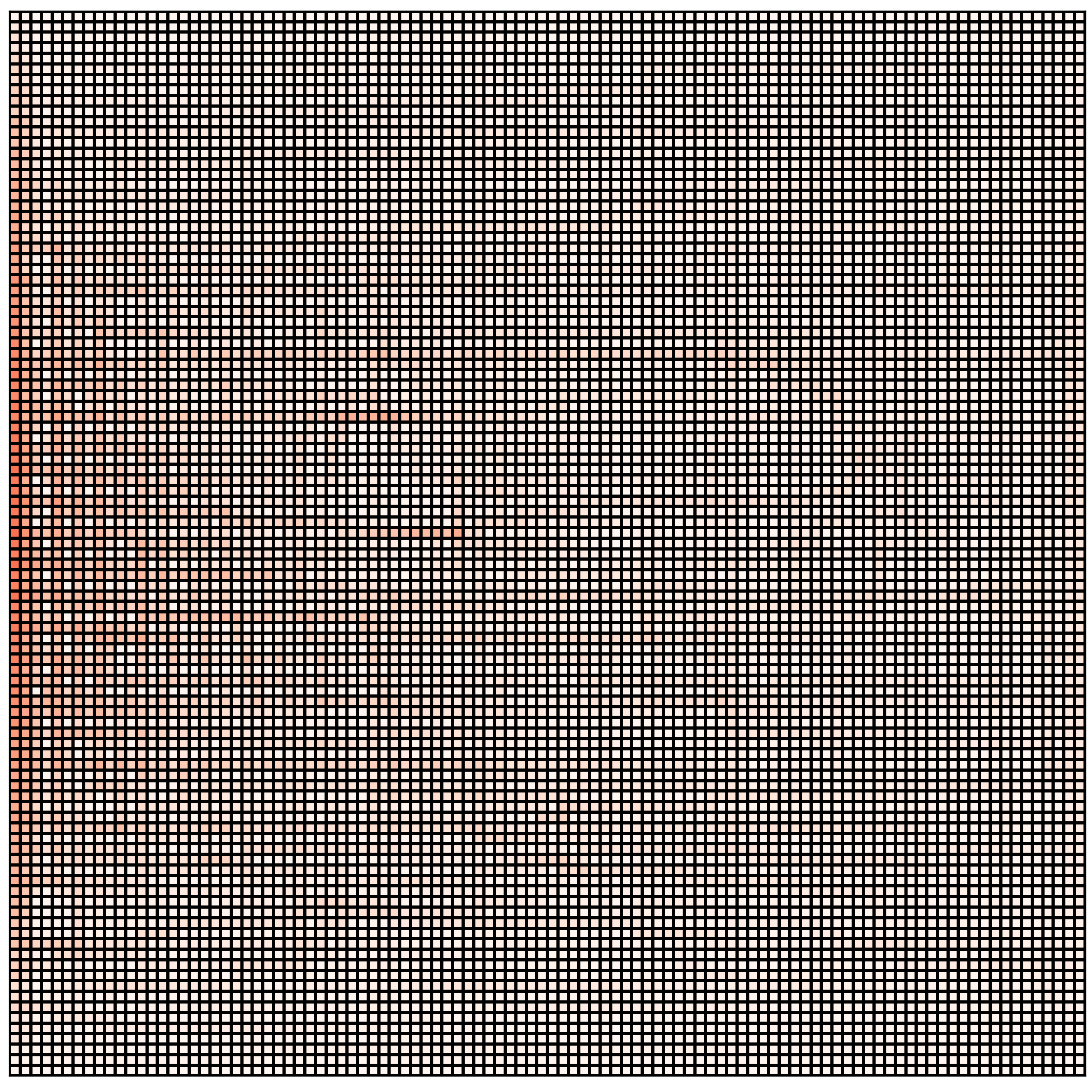}
        \caption{Warehouse (uneven) tile-usage map}
        \label{fig:tile-baseline-warehouse-uneven-tile-usage}
    \end{subfigure}\\
    \begin{subfigure}[t]{\EvalEnvShowSize\textwidth}
        \centering
        \includegraphics[width=1\textwidth]{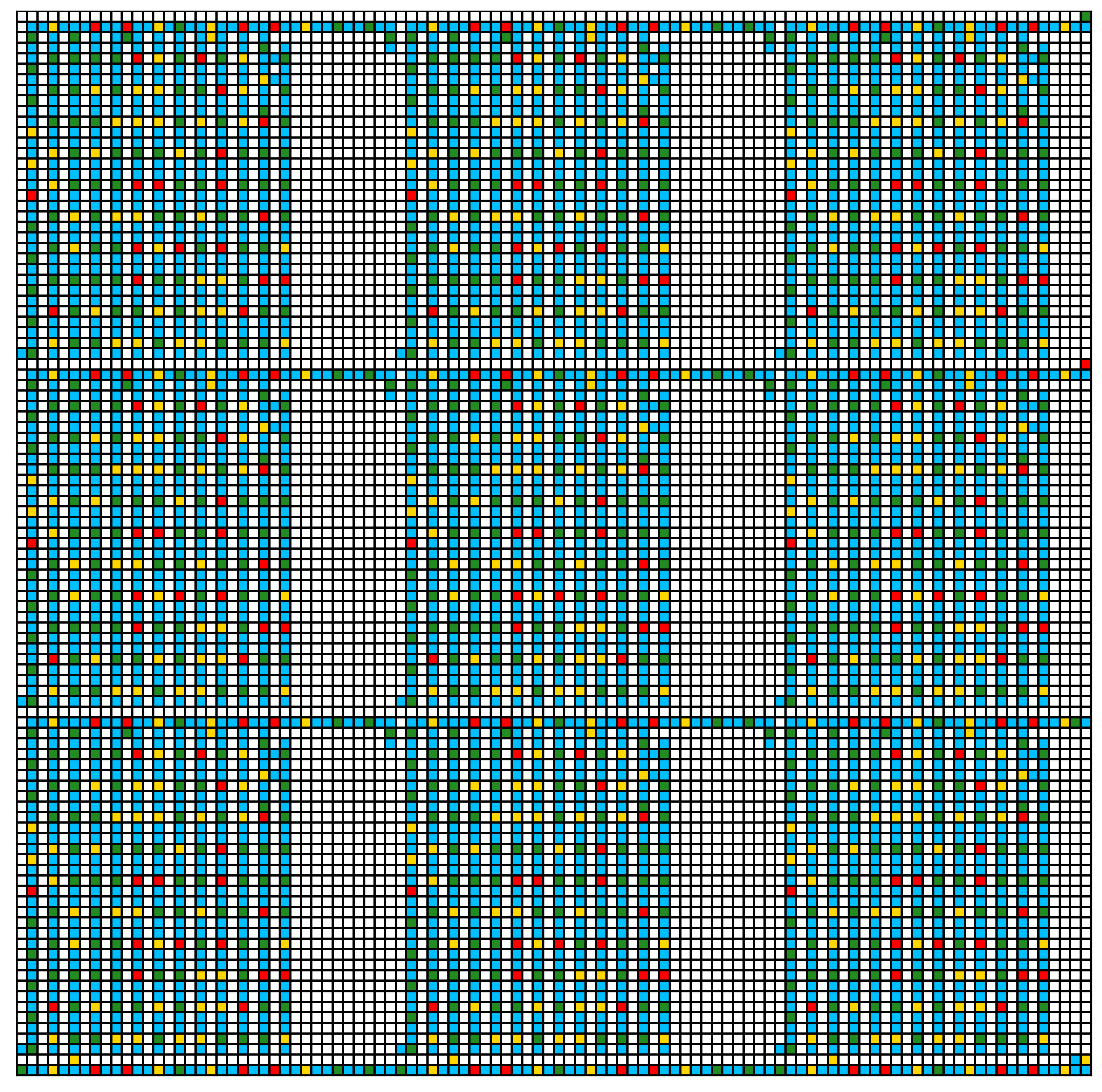}
        \caption{Manufacturing tiling environment}
        \label{fig:tile-baseline-manufacture}
    \end{subfigure}
    \hfill
    \begin{subfigure}[t]{\EvalEnvShowSize\textwidth}
        \centering
        \includegraphics[width=1\textwidth]{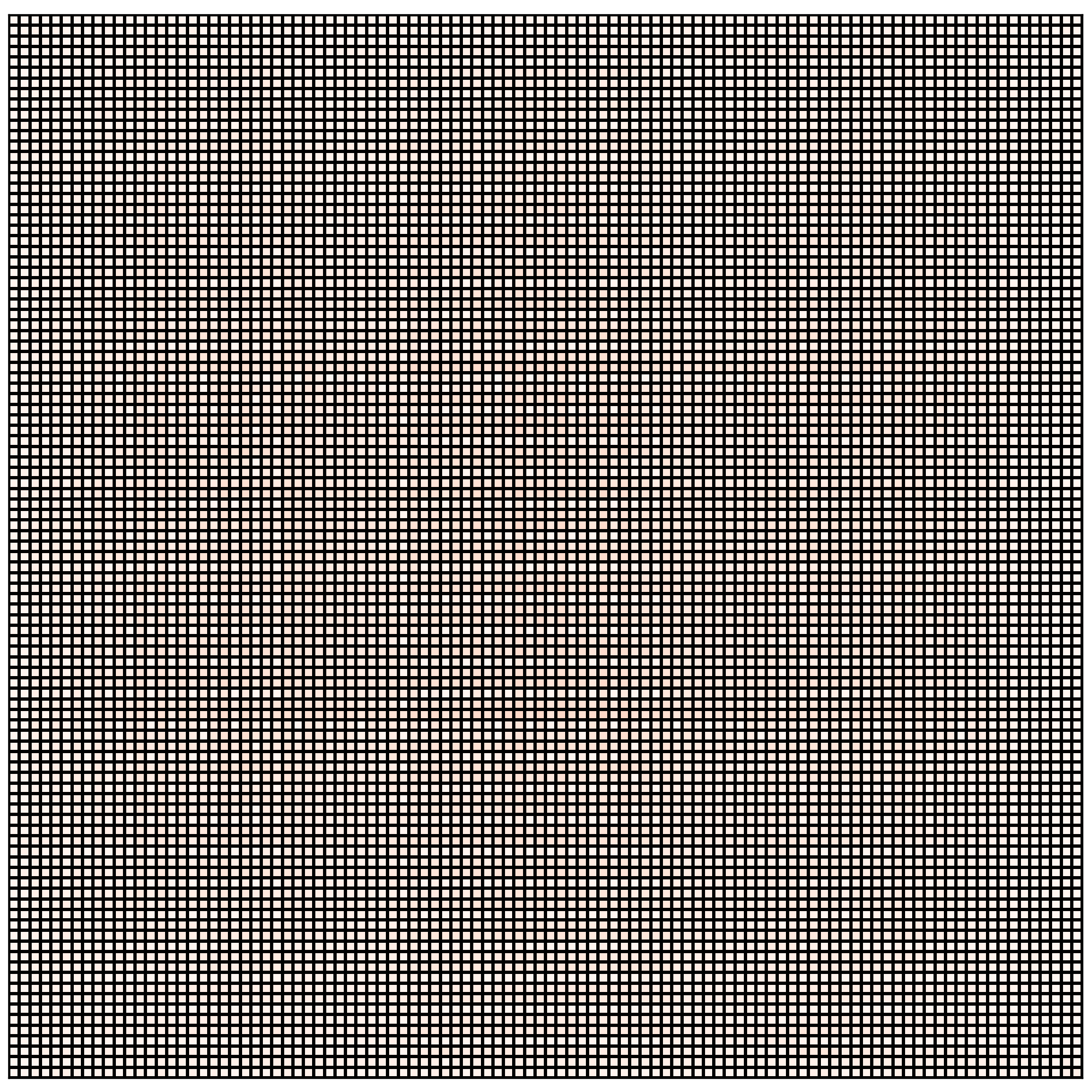}
        \caption{Manufacturing tile-usage map}
        \label{fig:tile-baseline-manufacture-tile-usage}
    \end{subfigure}\\
    \caption{Baseline environments as well as tile-usage maps of size $S_{eval}$ obtained by tiling smaller environments of size $S$.}
    \label{fig:tile-baseline}
\end{figure}

To further demonstrate the scalability of our method, we use trained NCA generators from \Cref{sec:result} to generate progressively larger environments and run simulations. \Cref{fig:scalability} shows the result. The y-axis illustrates two metrics: maximum mean throughput over 50 simulations (right) and the maximum scalability, defined as the agent count at this maximum (left).

We see an increasing trend for both maximum scalability and maximum mean throughput as the environment size increases. The NCA-generated environments generally scale better than the human-designed ones.

We see two exceptions: the maximum mean throughput in the 69$\times$69 warehouse (uneven) environment and both metrics in the 57$\times$58 manufacture environment. We can attribute this to the interaction between the MILP and the specific environment generated by the NCA. Since MILP makes numerous changes to the generated environments in these domains, certain combinations of generated environments and MILP random seeds can lead to repaired layouts that create congestion. However, if we encounter such issues in practice, we can either leverage a different NCA generator from the archive or re-run the MILP repair with a different random seed.

\subsection{Tiling Environments of size \texorpdfstring{$S$}{S}} \label{appen:tile-env-baseline}

Due to the similarity incurred in the NCA-generated environments of different sizes, one may argue that tiling small environments can create larger environments with competitive performance as the NCA-generated ones. To test this argument, we add a new baseline. We tile the environments of size $S$, namely \Cref{fig:warehouse-large-cma-mae-a=5-even} (warehouse (even)), \Cref{fig:warehouse-large-cma-mae-a=5-uneven} (warehouse (uneven)), and \Cref{fig:manufacture-large-cma-mae-a=5-opt} (manufacturing) in \Cref{appen:nca_gen}, to create environments of size $S_{eval}$ . We then use MILP to enforce constraints. 

We run 50 simulations with $N_{a\_eval}$ agents specified in \Cref{tab:search-space}. The new baseline achieves a success rate of only 0\% and 23\% in the warehouse (even) and warehouse (uneven) domains, respectively. \Cref{fig:tile-baseline} displays the tiled environments of size $S_{eval}$ and their tile-usage maps, which show the usage frequency of each tile in the simulation. As shown in \Cref{fig:tile-baseline-warehouse-even-tile-usage,fig:tile-baseline-warehouse-uneven-tile-usage}, the agents are congested, resulting in low success rates. In contrast, for the manufacturing domain, the baseline matches our method, with a success rate of 100\% and an average throughput of 22.73. This is because the tiling of \Cref{fig:manufacture-large-cma-mae-a=5-opt}, creating \Cref{fig:tile-baseline-manufacture}, resembles the NCA-generated patterns in \Cref{fig:manufacture-xxlarge-nca-repaired}. Thus, the tiling baseline may be a good method for the manufacturing domain, yet it falls short in the warehouse domains.

\subsection{QD Score and Archive Coverage}
\label{appen:add-result:qd-score-archive-cover}

\begin{figure}[!t]
    \centering
    \includegraphics[width=0.8\textwidth]{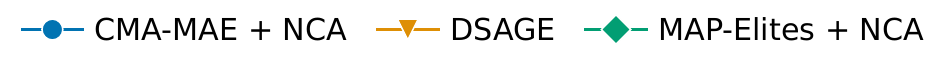}
    \begin{subfigure}[t]{0.49\textwidth}
        \includegraphics[width=1\textwidth]{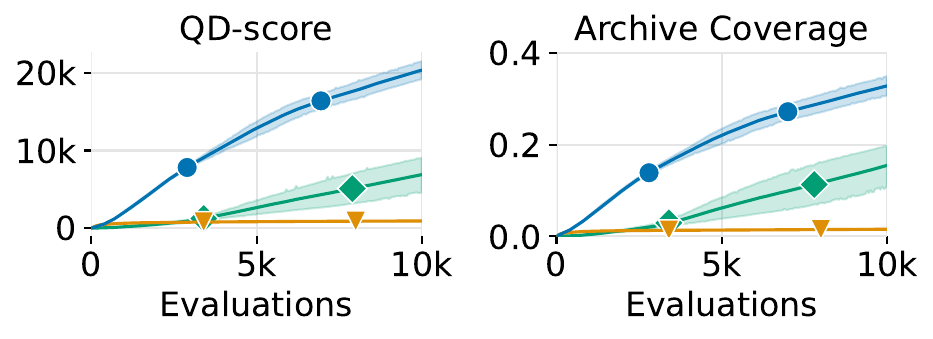}
        \caption{Warehouse (even) domain with environment entropy and number of connected shelf components measures.}
        \label{fig:qd-score-warehouse-even}
    \end{subfigure}
    \hfill
    \begin{subfigure}[t]{0.49\textwidth}
        \includegraphics[width=1\textwidth]{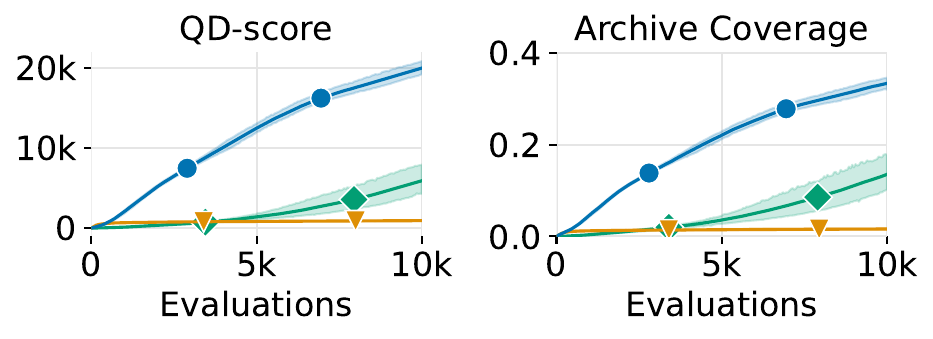}
        \caption{Warehouse (uneven) domain with environment entropy and number of connected shelf components measures.}
        \label{fig:qd-score-warehouse-uneven}
    \end{subfigure}\\
    \begin{subfigure}[t]{0.49\textwidth}
        \includegraphics[width=1\textwidth]{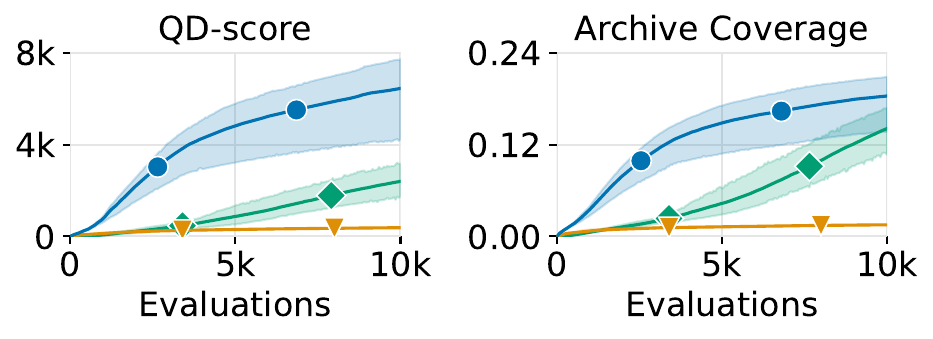}
        \caption{Manufacturing domain with environment entropy and number of workstations measures.}
        \label{fig:qd-score-manufacture}
    \end{subfigure}\\
        \begin{subfigure}[t]{0.49\textwidth}
        \includegraphics[width=1\textwidth]{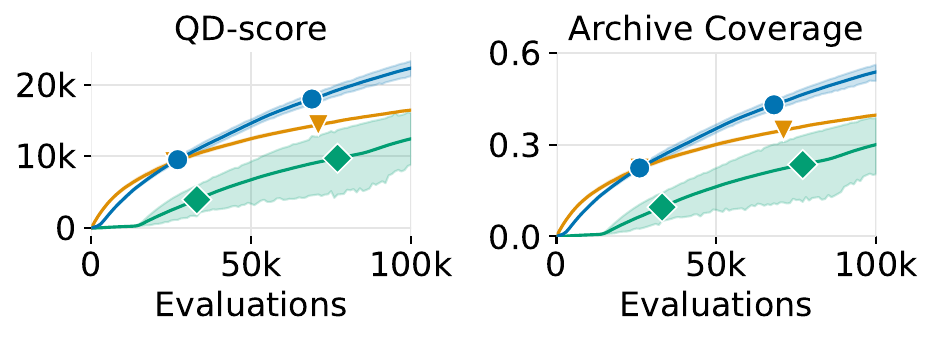}
        \caption{Maze domain with number of walls and average agent path length measures.}
        \label{fig:qd-score-maze-default}
    \end{subfigure}
    \hfill
    \begin{subfigure}[t]{0.49\textwidth}
        \includegraphics[width=1\textwidth]{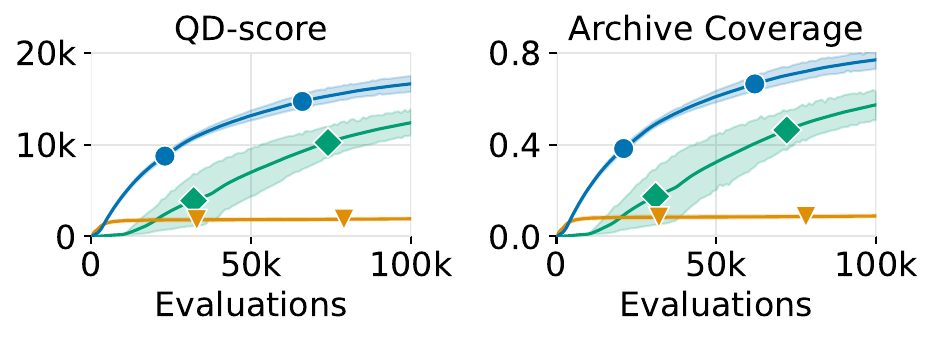}
        \caption{Maze domain with environment entropy and average agent path length measures.}
        \label{fig:qd-score-maze-entropy}
    \end{subfigure}
    \caption{QD-score and archive coverage of CMA-MAE + NCA, compared with MAP-Elites + NCA and DSAGE. The solid line shows the average and the shaded area shows the 95\% confidence interval. All algorithms use $\alpha = 5$.}
    \label{fig:qd-score}
\end{figure}

\begin{table}[!t]
\begin{center}
\resizebox{0.9\linewidth}{!}{
    \begin{tabular}{c|c||c|c}
    \toprule
    {} Domain & Algorithm   &                 QD-score & Archive Coverage \\
    \midrule
    \multirow{3}{*}{Maze w/o env entropy} & \textbf{CMA-MAE + NCA}    &    \textbf{22,299.00 $\pm$ 612.46} &  \textbf{0.54 $\pm$ 0.01} \\
     & DSAGE            &     16,446.60 $\pm$ 42.27 &  0.40 $\pm$ 0.00 \\
     & MAP-Elites + NCA &  12,446.80 $\pm$ 2,207.08 &  0.30 $\pm$ 0.05 \\
     \hline
    \multirow{3}{*}{Maze w/ env entropy} & \textbf{CMA-MAE + NCA}    &  \textbf{12,468.20 $\pm$ 342.09} &  \textbf{0.77 $\pm$ 0.02} \\
    & DSAGE            &    1,444.20 $\pm$ 52.60 &  0.09 $\pm$ 0.00 \\
    & MAP-Elites + NCA &   9,288.80 $\pm$ 590.13 &  0.57 $\pm$ 0.04 \\
    \hline
    \multirow{3}{*}{Warehouse (even)} & \textbf{CMA-MAE + NCA}    &   \textbf{20,366.07 $\pm$ 653.46} &  \textbf{0.33 $\pm$ 0.01} \\
    & DSAGE            &       914.60 $\pm$ 33.09 &  0.02 $\pm$ 0.00 \\
    & MAP-Elites + NCA &  6,893.49 $\pm$ 1,291.68 &  0.15 $\pm$ 0.03 \\
    \hline
    \multirow{3}{*}{Warehouse (uneven)} & \textbf{CMA-MAE + NCA}    &   \textbf{19,985.23 $\pm$ 493.98} &  \textbf{0.33 $\pm$ 0.01} \\
    & DSAGE            &       926.69 $\pm$ 23.42 &  0.02 $\pm$ 0.00 \\
    & MAP-Elites + NCA &  5,906.86 $\pm$ 1,027.29 &  0.14 $\pm$ 0.02 \\
    \hline
    \multirow{3}{*}{Manufacturing} & \textbf{CMA-MAE + NCA}    &  \textbf{6,449.56 $\pm$ 1,106.83} &  \textbf{0.18 $\pm$ 0.02} \\
    & DSAGE            &        378.33 $\pm$ 4.69 &  0.02 $\pm$ 0.00 \\
    & MAP-Elites + NCA &    2,398.99 $\pm$ 422.61 &  0.14 $\pm$ 0.02 \\
    \bottomrule
    \end{tabular}
}
\end{center}
\caption{QD-score and Archive coverage. $x \pm y$ denotes an average value of $x$ and a standard error of $y$ over 5 runs. All algorithms use $\alpha = 5$. ``Maze w/ env entropy'' refers to using environment entropy and average agent path length as the measures in the maze domain, while ``Maze w/o env entropy refers to using number of walls and average agent path length as the measures.}
\label{table:qd-score-archive-cover}
\end{table}

\begin{figure}[!t]
    \centering
    \begin{minipage}[b]{0.13\textwidth}
        \centering
        Algorithm:
    \end{minipage}
    \hfill
    \begin{minipage}[b]{0.27\textwidth}
        \centering
        CMA-MAE + NCA
    \end{minipage}
    \hfill
    \begin{minipage}[b]{0.27\textwidth}
        \centering
        MAP-Elites + NCA
    \end{minipage}
    \hfill
    \begin{minipage}[b]{0.27\textwidth}
        \centering
        DSAGE
    \end{minipage}\\
    \begin{minipage}[b]{0.13\textwidth}
        \centering
        Warehouse (even)
        \vspace{1.4cm}
    \end{minipage}
    \hfill
    \begin{subfigure}[t]{0.27\textwidth}
        \includegraphics[width=1\textwidth]{figs/archive/warehouse/even/cma-mae-a=5/heatmap_archive_200.pdf}
        \label{fig:archive-cma-mae-warehouse-even}
    \end{subfigure}
    \hfill
    \begin{subfigure}[t]{0.27\textwidth}
        \includegraphics[width=1\textwidth]{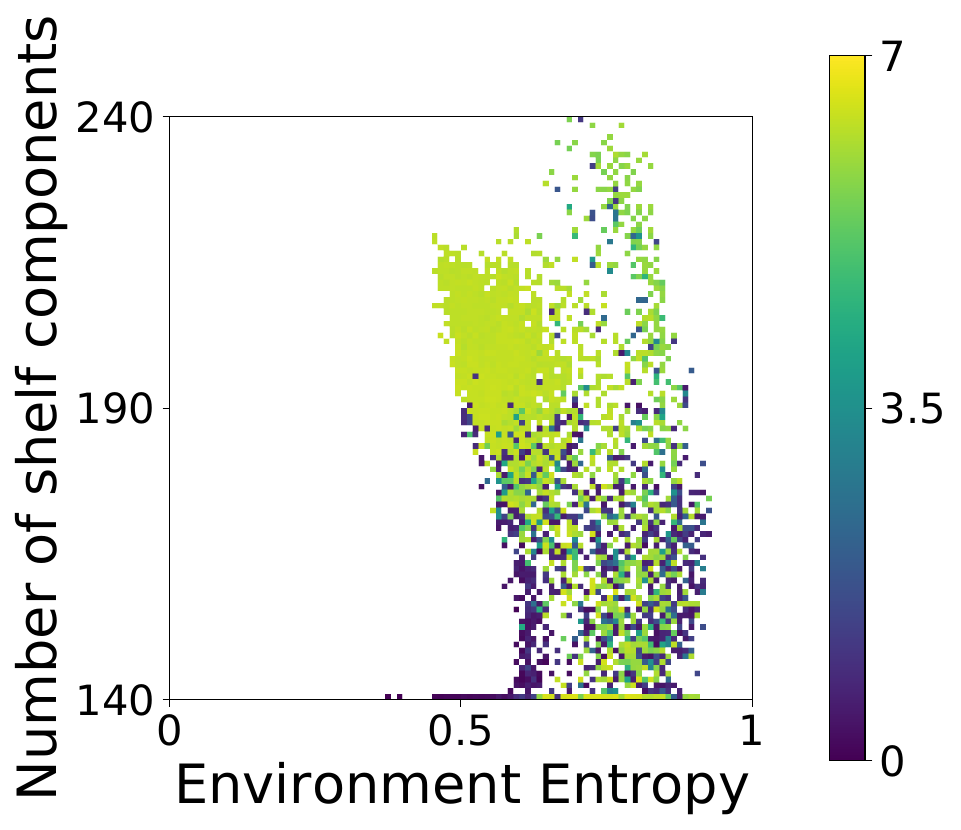}
        \label{fig:archive-me-warehouse-even}
    \end{subfigure}
    \hfill
    \begin{subfigure}[t]{0.27\textwidth}
        \includegraphics[width=1\textwidth]{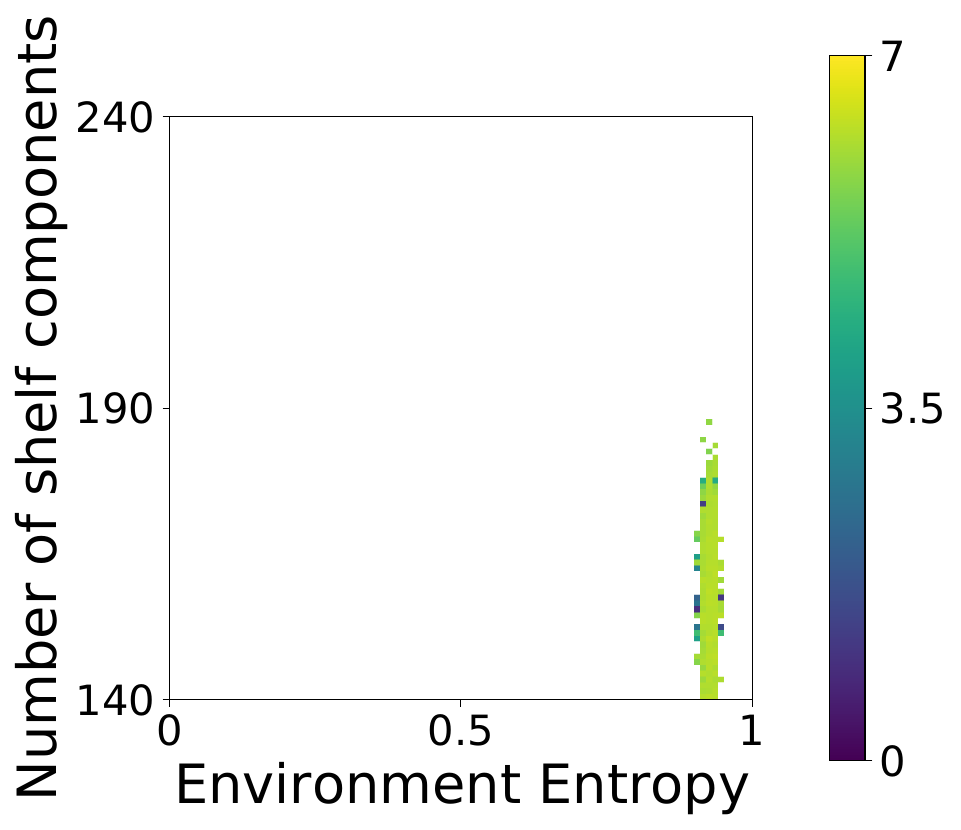}
        \label{fig:archive-dsage-warehouse-even}
    \end{subfigure}\\
    \begin{minipage}[b]{0.13\textwidth}
        \centering
        Warehouse (uneven)
        \vspace{1.4cm}
    \end{minipage}
    \hfill
    \begin{subfigure}[t]{0.27\textwidth}
        \includegraphics[width=1\textwidth]{figs/archive/warehouse/uneven/cma-mae-a=5/heatmap_archive_200.pdf}
        \label{fig:archive-cma-mae-warehouse-uneven}
    \end{subfigure}
    \hfill
    \begin{subfigure}[t]{0.27\textwidth}
        \includegraphics[width=1\textwidth]{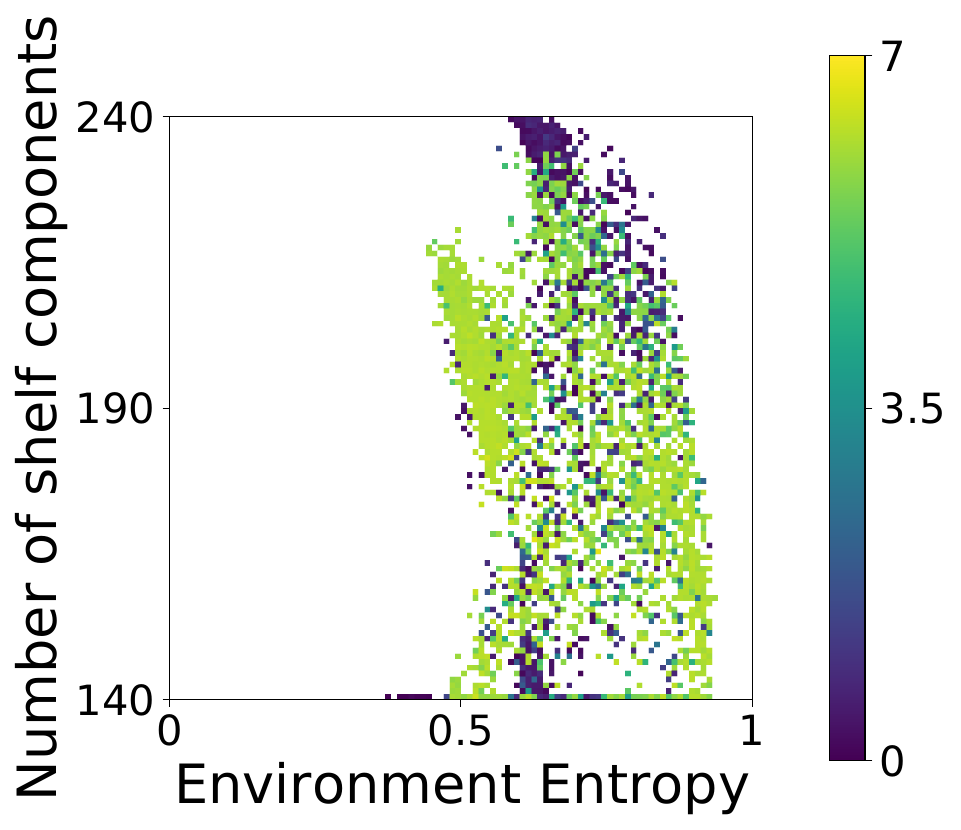}
        \label{fig:archive-me-warehouse-uneven}
    \end{subfigure}
    \hfill
    \begin{subfigure}[t]{0.27\textwidth}
        \includegraphics[width=1\textwidth]{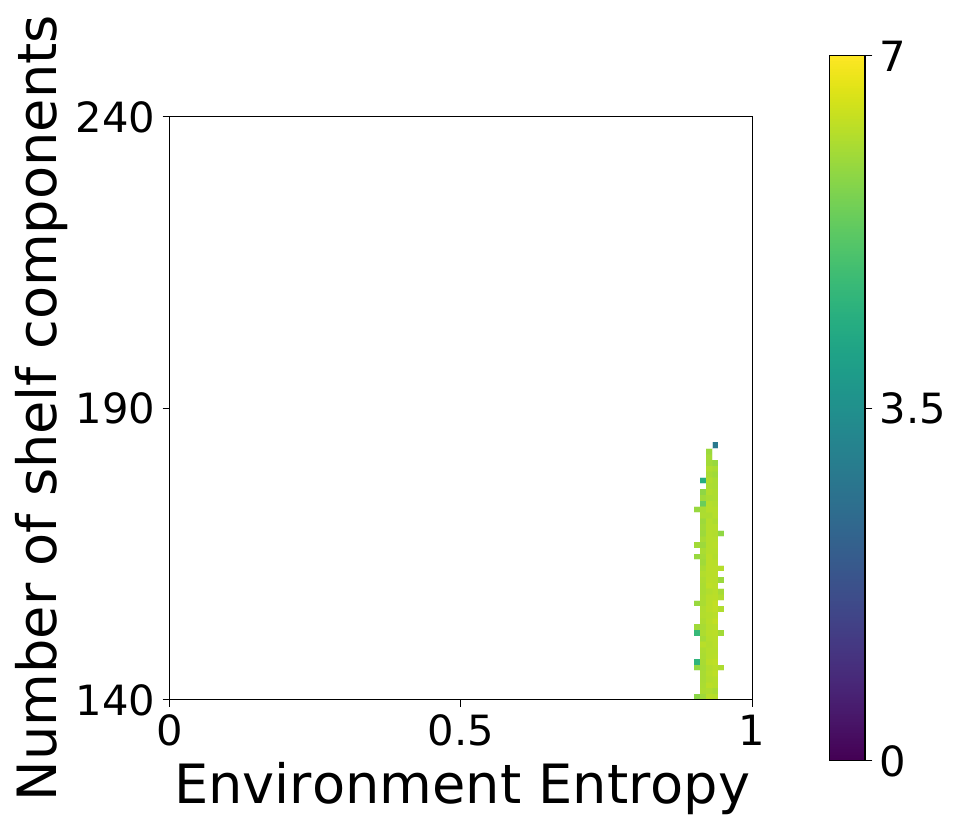}
        \label{fig:archive-dsage-warehouse-uneven}
    \end{subfigure}\\
    \begin{minipage}[b]{0.13\textwidth}
        \centering
        Manufacturing
        \vspace{1.2cm}
    \end{minipage}
    \hfill
    \begin{subfigure}[t]{0.27\textwidth}
        \includegraphics[width=1\textwidth]{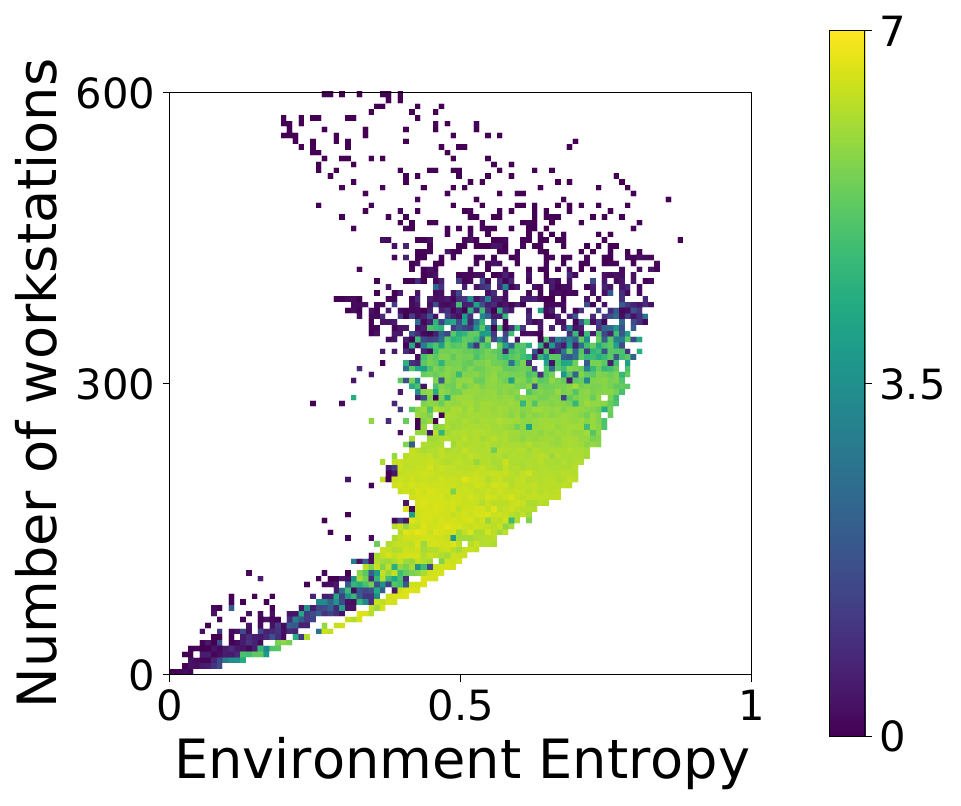}
        \label{fig:archive-cma-mae-manufacture}
    \end{subfigure}
    \hfill
    \begin{subfigure}[t]{0.27\textwidth}
        \includegraphics[width=1\textwidth]{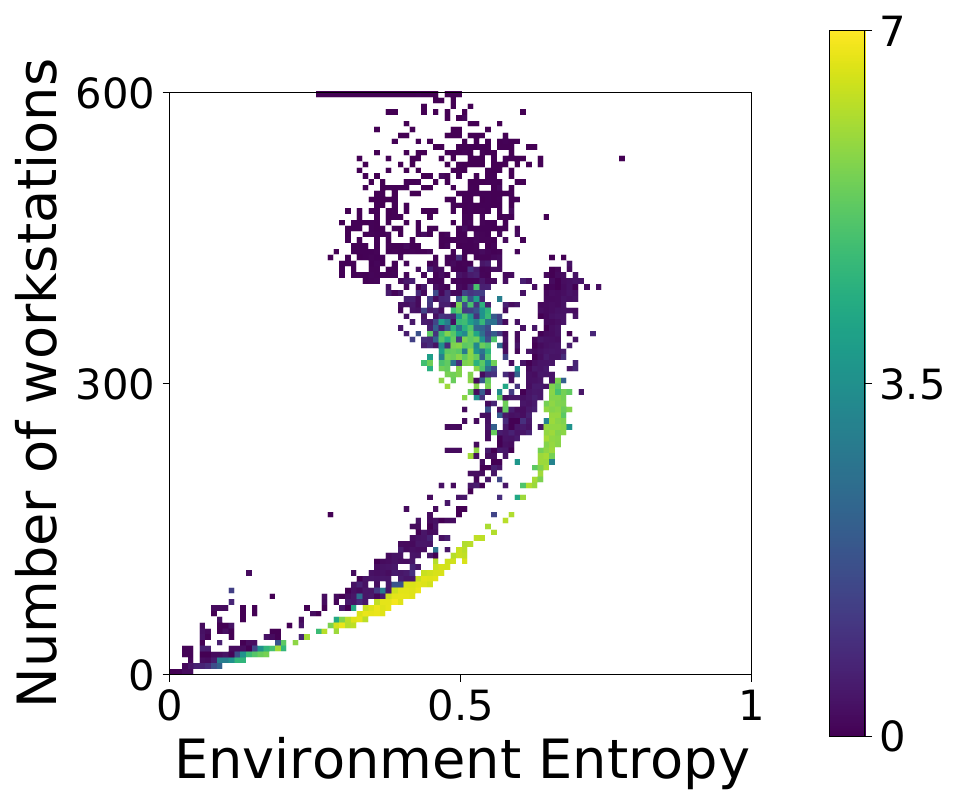}
        \label{fig:archive-me-manufacture}
    \end{subfigure}
    \hfill
    \begin{subfigure}[t]{0.27\textwidth}
        \includegraphics[width=1\textwidth]{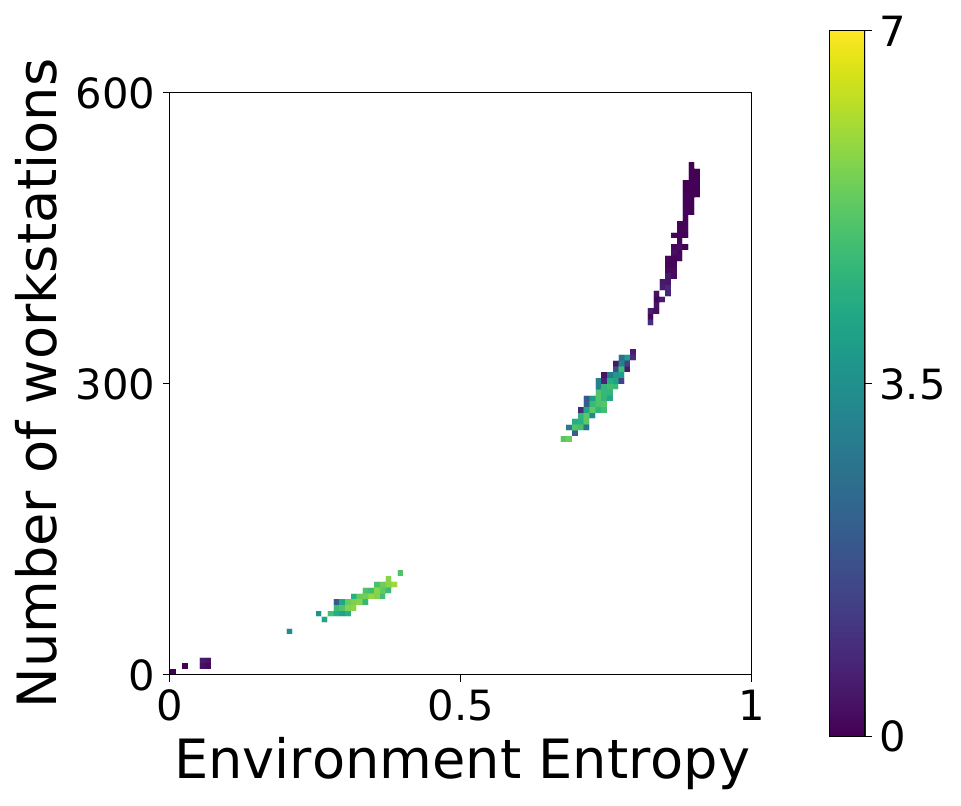}
        \label{fig:archive-dsage-manufacture}
    \end{subfigure}\\
    \begin{minipage}[b]{0.13\textwidth}
        \centering
        Maze w/o \\ env entropy
        \vspace{1.2cm}
    \end{minipage}
    \hfill
    \begin{subfigure}[t]{0.27\textwidth}
        \includegraphics[width=1\textwidth]{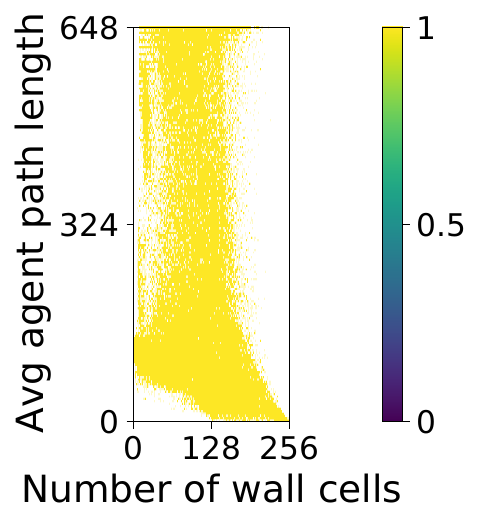}
        \label{fig:archive-cma-mae-maze-default}
    \end{subfigure}
    \hfill
    \begin{subfigure}[t]{0.27\textwidth}
        \includegraphics[width=1\textwidth]{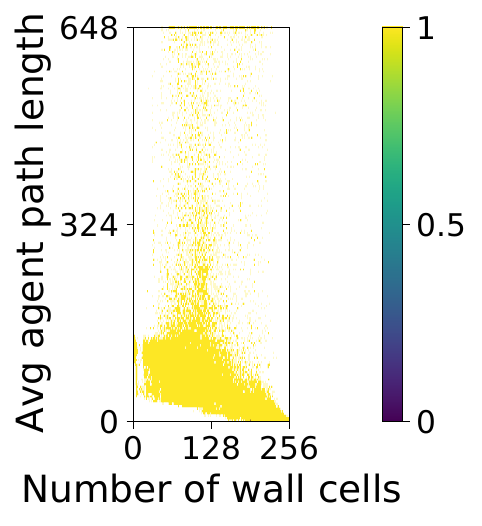}
        \label{fig:archive-me-maze-default}
    \end{subfigure}
    \hfill
    \begin{subfigure}[t]{0.27\textwidth}
        \includegraphics[width=1\textwidth]{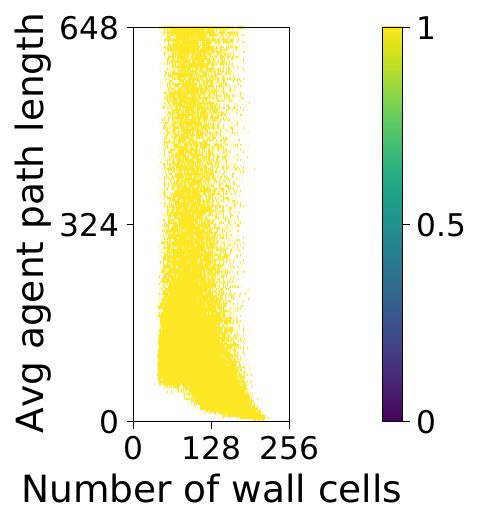}
        \label{fig:archive-dsage-maze-default}
    \end{subfigure}\\
    \begin{minipage}[b]{0.13\textwidth}
        \centering
        Maze w/ \\env entropy
        \vspace{1.2cm}
    \end{minipage}
    \hfill
    \begin{subfigure}[t]{0.27\textwidth}
        \includegraphics[width=1\textwidth]{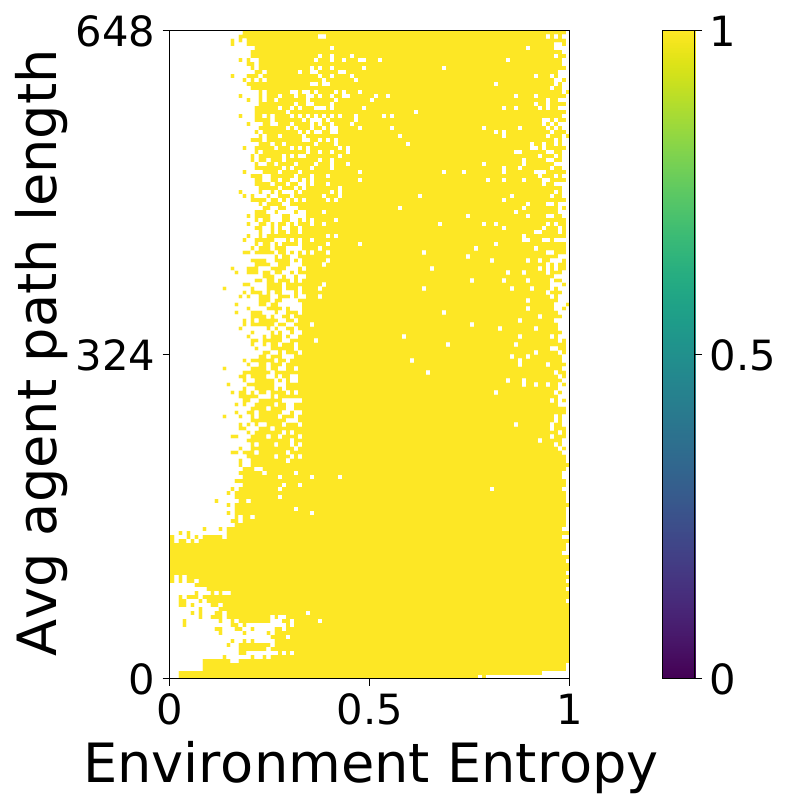}
        \label{fig:archive-cma-mae-maze-entropy}
    \end{subfigure}
    \hfill
    \begin{subfigure}[t]{0.27\textwidth}
        \includegraphics[width=1\textwidth]{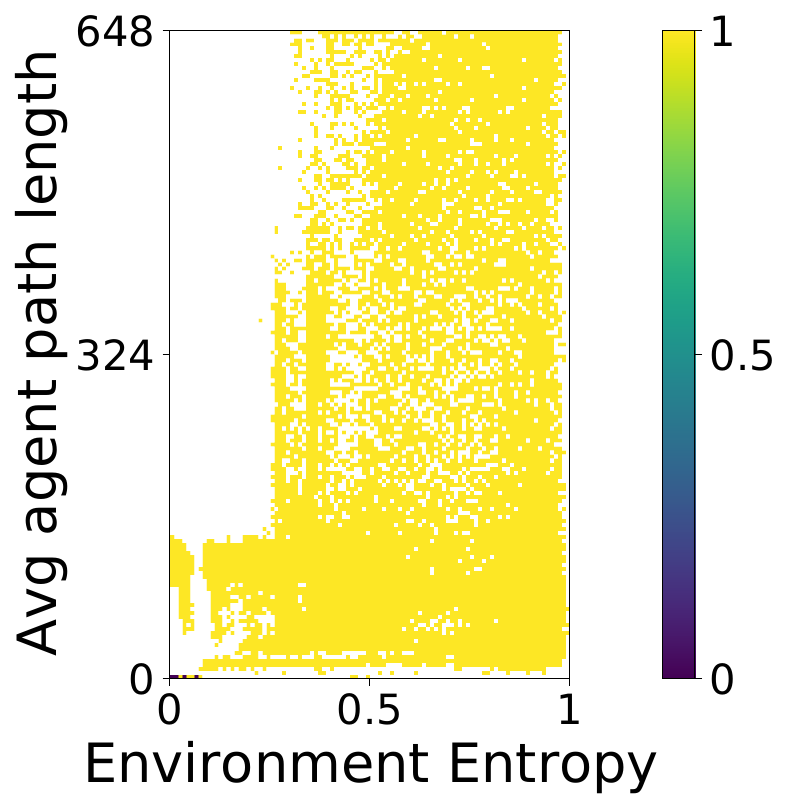}
        \label{fig:archive-me-maze-entropy}
    \end{subfigure}
    \hfill
    \begin{subfigure}[t]{0.27\textwidth}
        \includegraphics[width=1\textwidth]{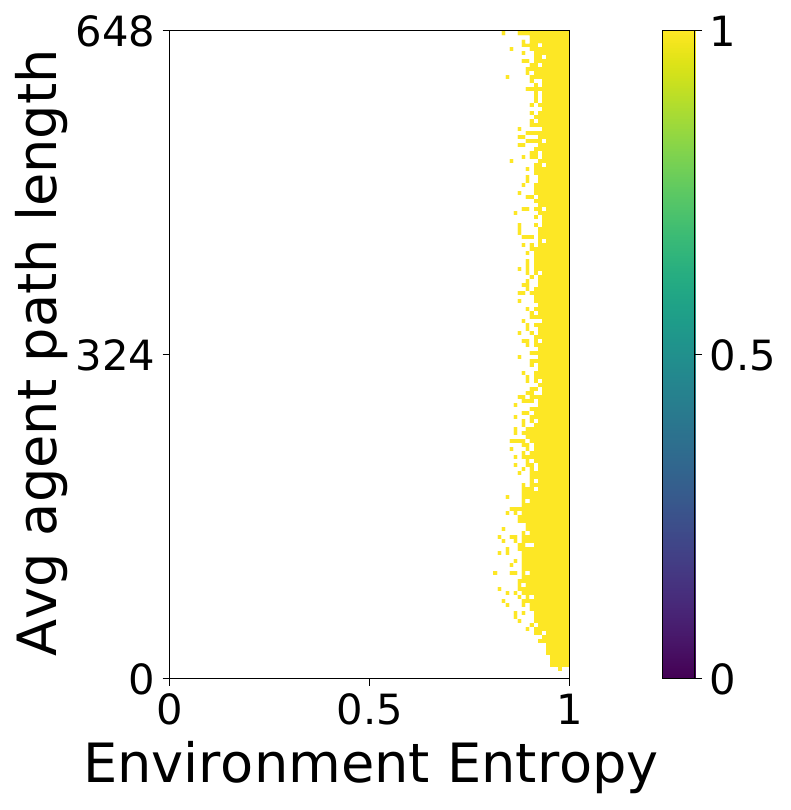}
        \label{fig:archive-dsage-maze-entropy}
    \end{subfigure}
    \caption{Example result archives of all domains. All algorithms use $\alpha = 5$.
    }
    \label{fig:archive}
\end{figure}

\Cref{fig:qd-score} shows the QD-score and the archive coverage over the number of evaluations during training. \Cref{table:qd-score-archive-cover} shows the corresponding numerical results. \Cref{fig:archive} then shows the result archive of one of the runs.

We compare CMA-MAE with MAP-Elites~\cite{mouret2015illuminating} to show the benefit of using CMA-MAE to train NCA. We use the state-of-the-art Iso+LineDD mutation operator~\cite{vassiliades2018isoline} on MAP-Elites with $\sigma_{line} = 0.2$ and $\sigma_{iso} = 0.01$. We also compare CMA-MAE + NCA with DSAGE to demonstrate the benefit of using NCA to generate environments with regularized patterns.
In the maze domain, in addition to the combination of diversity measures discussed in \Cref{sec:domain} (number of walls and average agent path length), we run experiments with the environment entropy measure, paired with average agent path length, to demonstrate the effect of the environment entropy measure on the QD-score and archive coverage.

We observe that CMA-MAE achieves the best QD-score and the best archive coverage in all domains, generating solutions with the best quality and diversity. We also observe from the result archives that CMA-MAE covers the largest number of cells. Notably, in all domains, DSAGE fails to diversify environment entropy, covering only the high entropy area of the archives and not exploring the low entropy region. This happens because we directly optimize environments instead of training NCA generators in DSAGE. Without NCA, DSAGE cannot generate candidate environments with low entropy (i.e., with regularized patterns). As a result, despite being high-quality, the DSAGE-optimized environments lack regularized patterns and cannot be scaled to arbitrary sizes.

\subsection{Compare CMA-ES with CMA-MAE}
\label{appen:add-result:cma-es}

\begin{figure}[!t]
    \centering
    \includegraphics[width=0.75\textwidth]{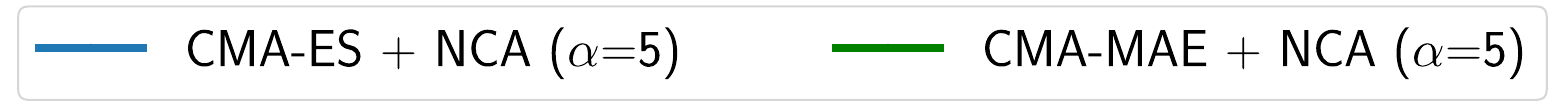}
    \begin{subfigure}[t]{0.32\textwidth}
        \centering
        \includegraphics[width=1\textwidth]{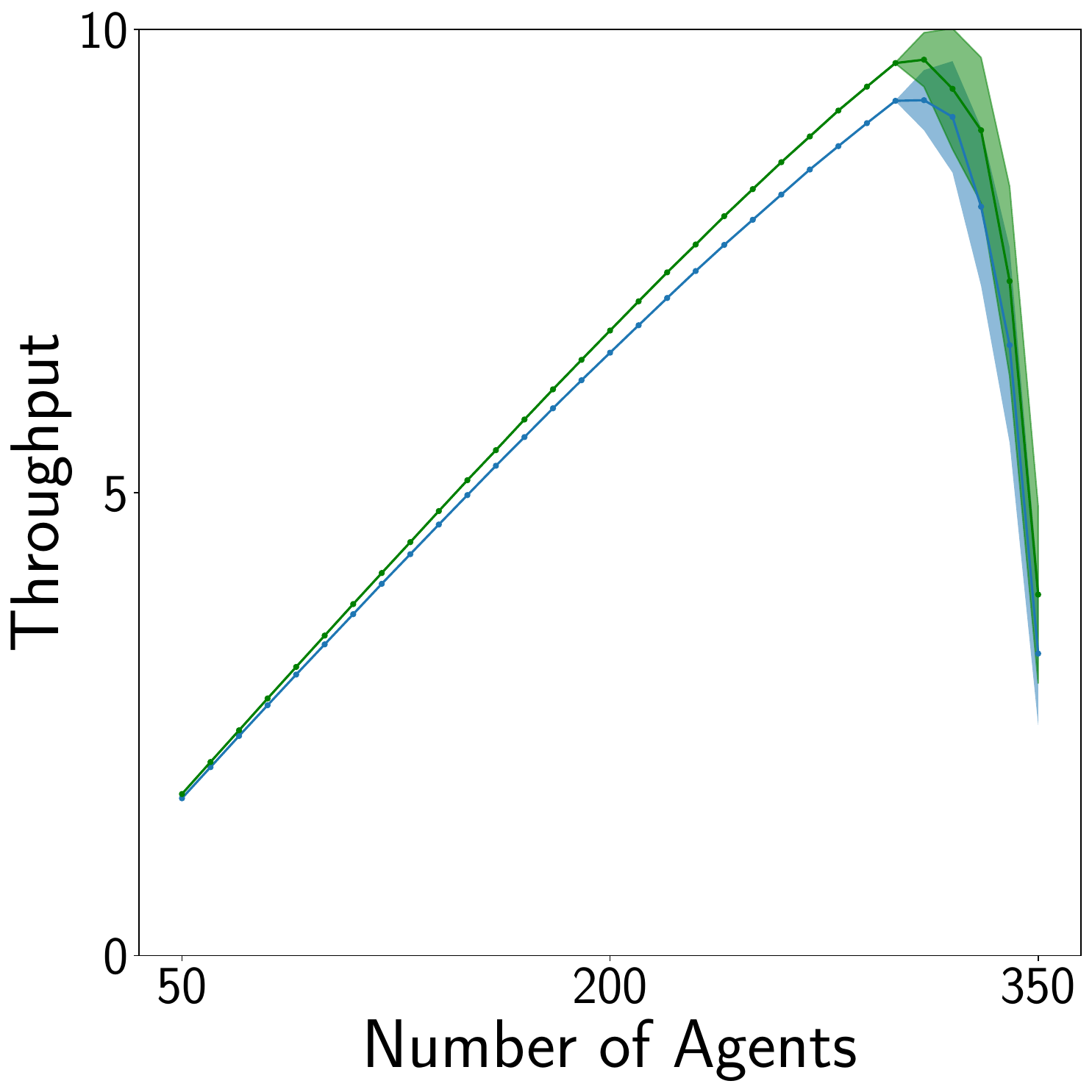}
        \caption{$S$: Warehouse (even)}
    \end{subfigure}
    \hfill
    \begin{subfigure}[t]{0.32\textwidth}
        \centering
        \includegraphics[width=1\textwidth]{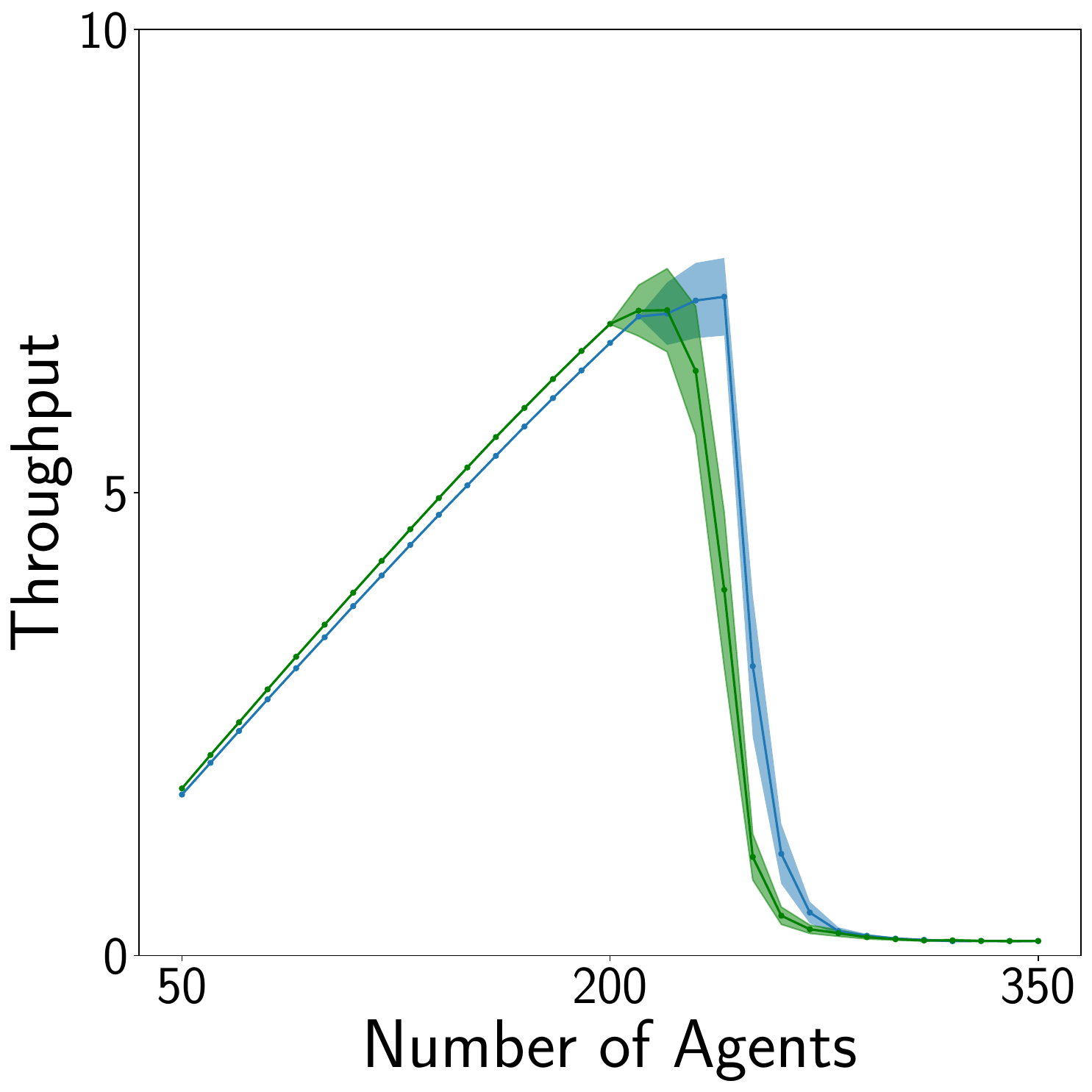}
        \caption{$S$: Warehouse (uneven)}
    \end{subfigure}
    \hfill
    \begin{subfigure}[t]{0.32\textwidth}
        \centering
        \includegraphics[width=1\textwidth]{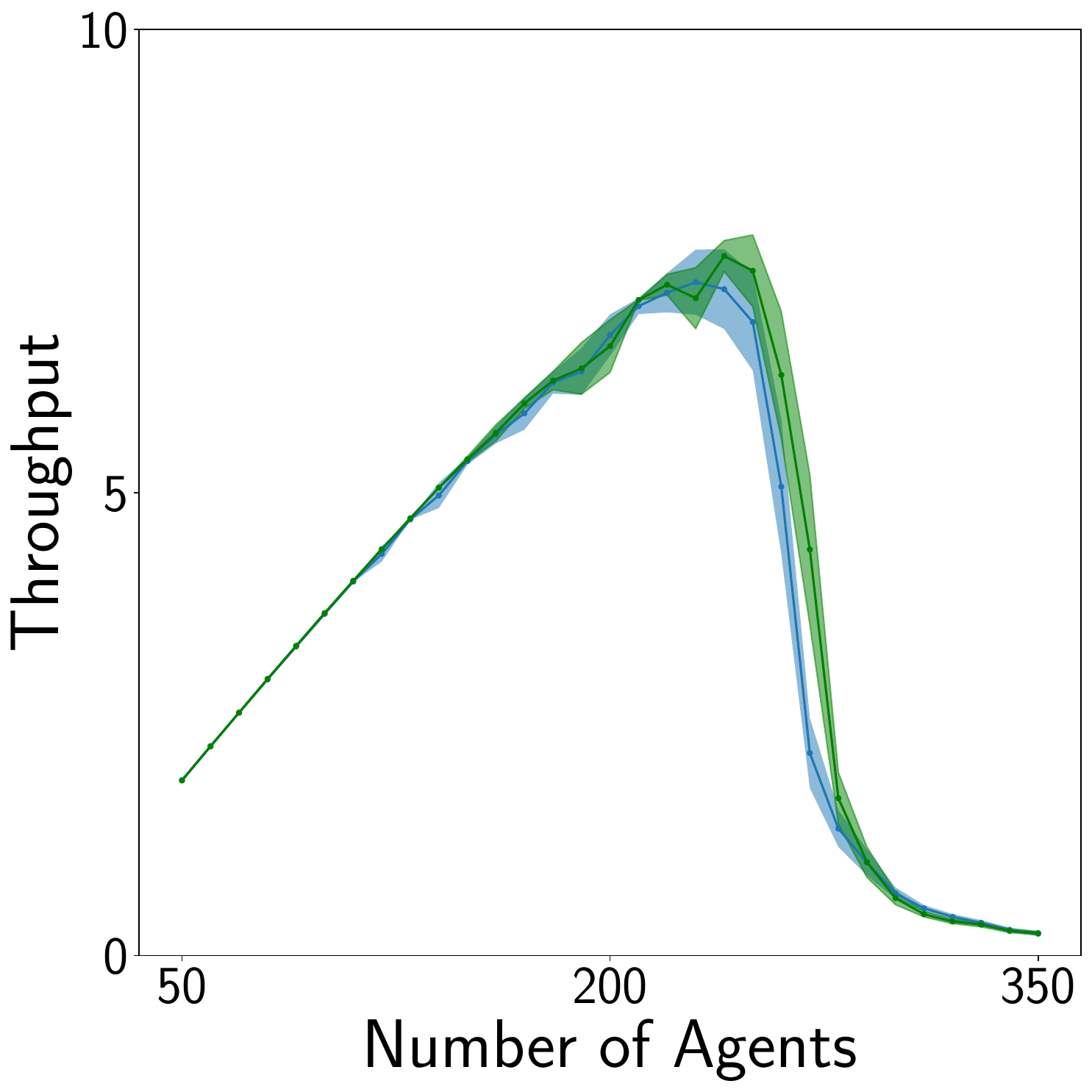}
        \caption{$S$: Manufacturing}
    \end{subfigure}\\
    \begin{subfigure}[t]{0.32\textwidth}
        \centering
        \includegraphics[width=1\linewidth]{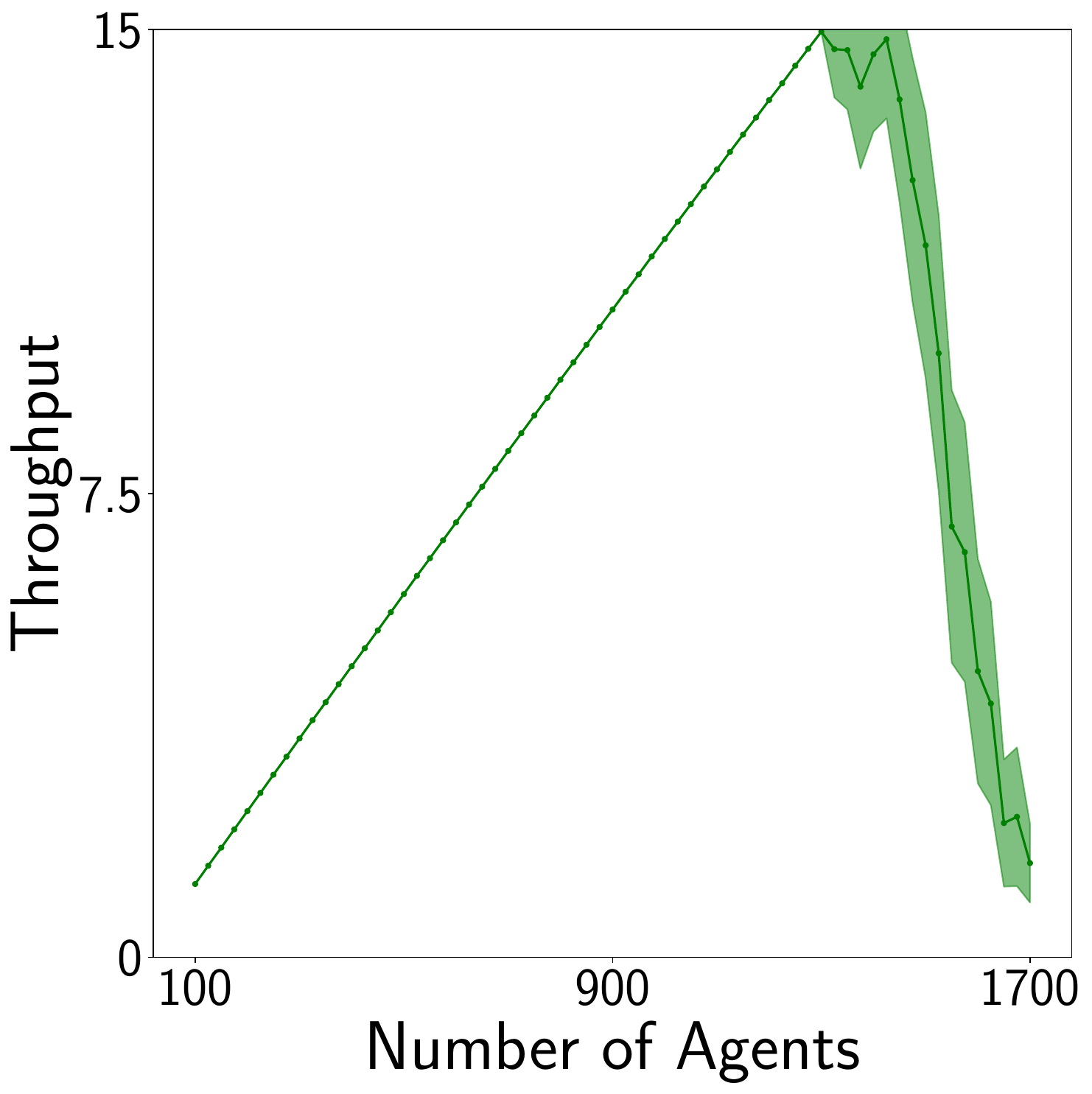}
        \caption{$S_{eval}$: Warehouse (even)}
    \end{subfigure}
    \begin{subfigure}[t]{0.32\textwidth}
        \centering
        \includegraphics[width=1\textwidth]{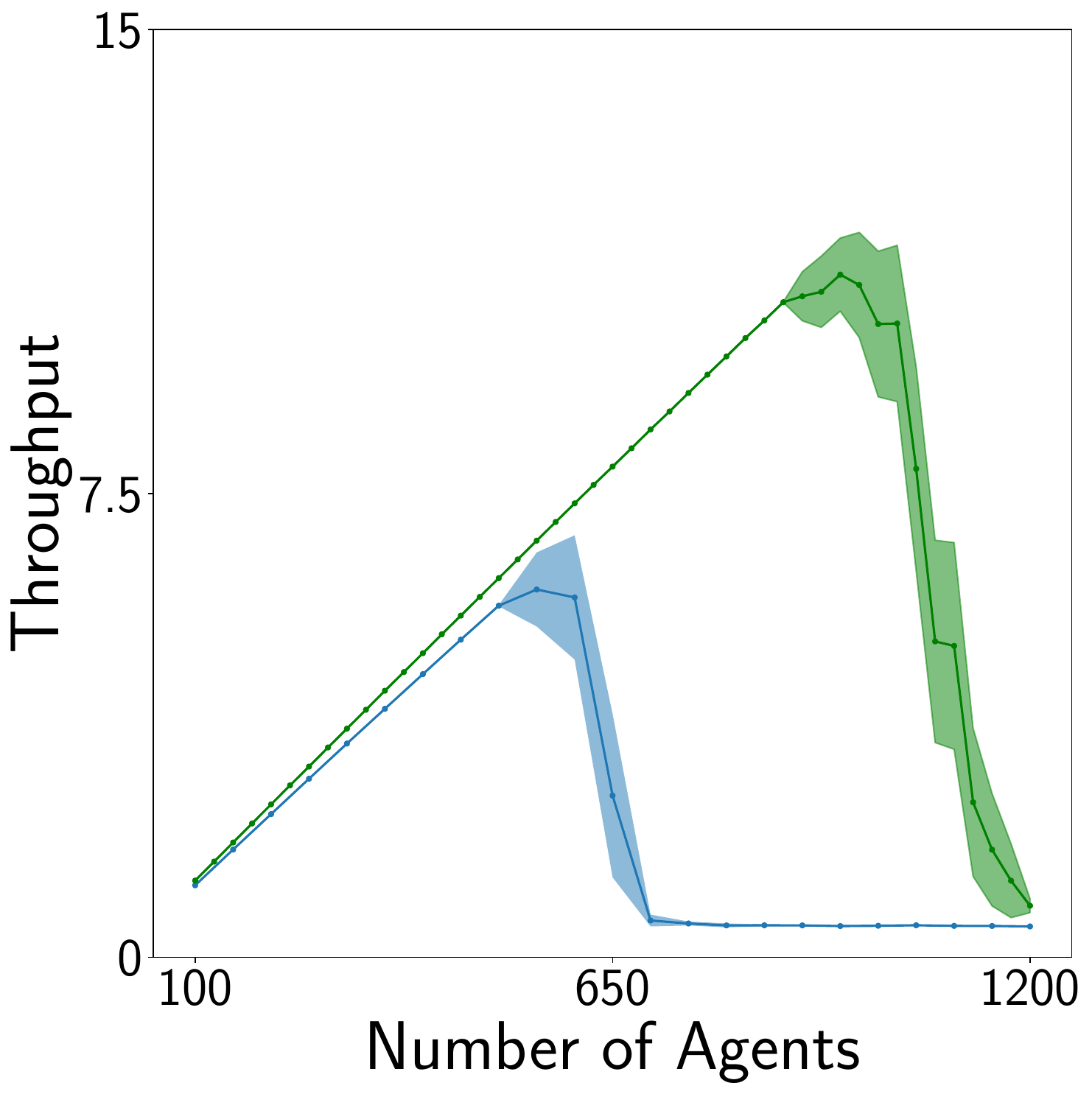}
        \caption{$S_{eval}$: Warehouse (uneven)}
    \end{subfigure}
    \begin{subfigure}[t]{0.32\textwidth}
        \centering
        \includegraphics[width=1\textwidth]{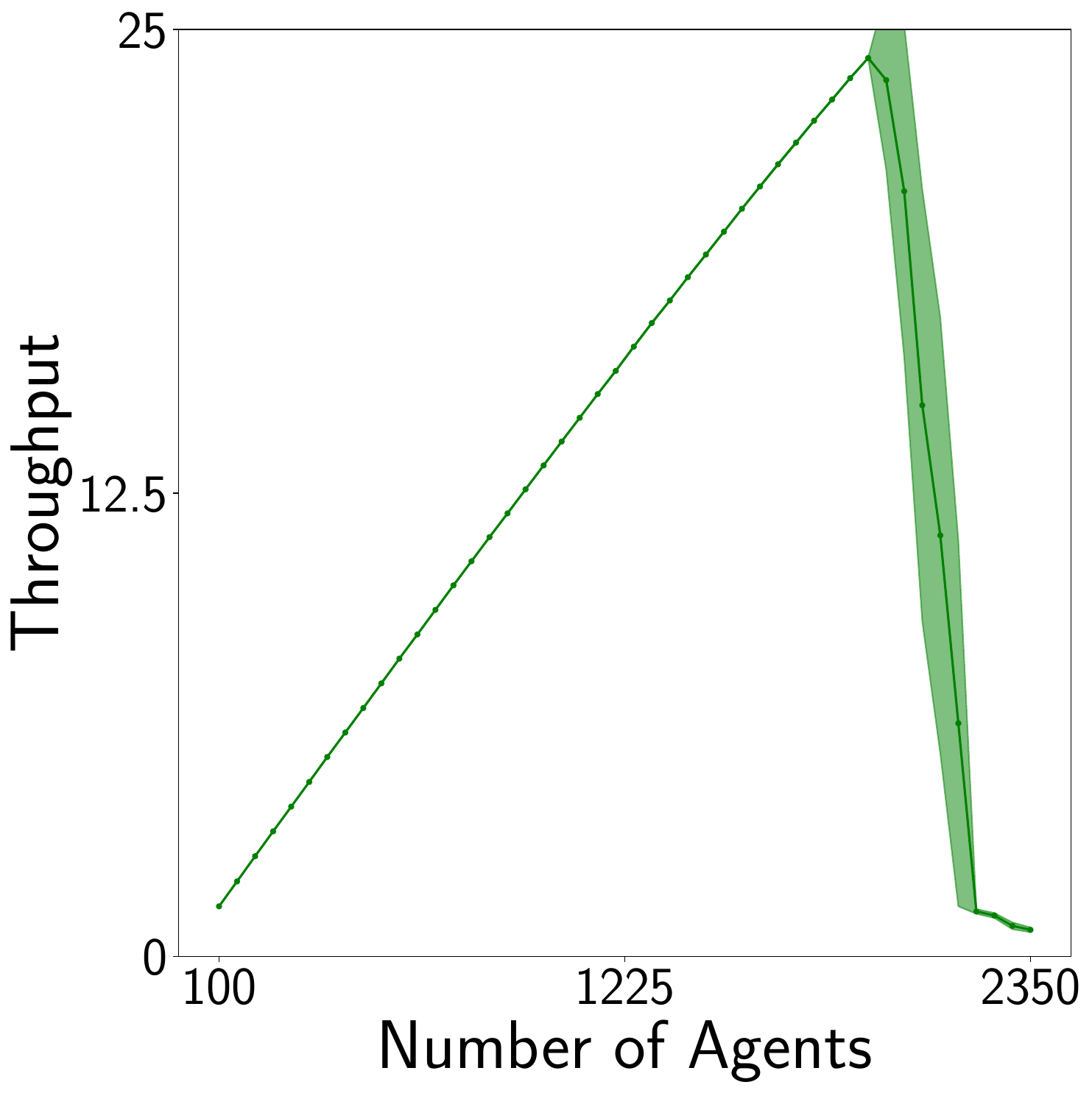}
        \caption{$S_{eval}$: Manufacturing}
    \end{subfigure}
    \caption{Throughput with an increasing number of agents in environments of sizes $S$ and $S_{eval}$ generated by CMA-ES + NCA and CMA-MAE + NCA. The NCA generators trained by CMA-ES fail to generate environments of size $S_{eval}$ in the warehouse (even) and manufacturing domains because the unrepaired environments deviate too much from the domain-specific constraints.
    The solid lines are the average throughput while the shaded area shows the 95\% confidence interval.}
    \label{fig:comp_cma-es_cma-mae}
\end{figure}

\begin{table}[!t]
    \centering
    \begin{center}
    \resizebox{\linewidth}{!}{
        \begin{tabular}{c|c||c|c||c|c}
        \toprule
                                         &      & \multicolumn{2}{c||}{Size $S$ with $N_a$ agents} & \multicolumn{2}{c}{Size $S_{eval}$ with $N_{a\_eval}$ agents}\\
        \hline
        Domain                           & Algorithm                 & Success Rate                & Throughput                         & Success Rate                & Throughput                      \\
        \hline
        \multirow{2}{*}{warehouse (even)} & CMA-ES + NCA & \textbf{100\%} & 6.51 $\pm$ 0.00 & N/A  & N/A \\
                                         & CMA-MAE + NCA & \textbf{100\%} & \textbf{6.74 $\pm$ 0.00} & \textbf{90\%} & \textbf{16.01 $\pm$ 0.00} \\
        \hline
        \multirow{2}{*}{warehouse (uneven)} & CMA-ES + NCA & \textbf{100\%} &  6.62 $\pm$ 0.00 & 0\% & N/A \\
                                           & CMA-MAE + NCA& \textbf{100\%} & \textbf{6.82 $\pm$ 0.00} & \textbf{84\%} & \textbf{12.03 $\pm$ 0.00}\\
        \hline
        \multirow{2}{*}{Manufacturing}       & CMA-ES + NCA   & \textbf{98\%}& 6.81 $\pm$ 0.01 & N/A & N/A \\
                                           & CMA-MAE + NCA  & 94\%  & \textbf{6.82 $\pm$ 0.00} & \textbf{100\%} & \textbf{23.11 $\pm$ 0.01}\\
        \bottomrule
        \end{tabular}
    }
    \end{center}
    \caption{Success rates and throughput of environments of sizes $S$ and $S_{eval}$ generated by CMA-ES + NCA and CMA-MAE + NCA.  Both algorithms use $\alpha=5$. We run 50 simulations for all environments except for the warehouse (uneven) environment of size $S_{eval}$ generated by CMA-ES + NCA, for which we run 20 simulations. We measure the throughput of only successful simulations and report both average and standard error.
}
    \label{tab:numerical-comp-cma-mae_cma-es}
\end{table}

\begin{figure}[!t]
    \centering
    \begin{subfigure}[t]{\EvalEnvShowSize\textwidth}
        \includegraphics[width=1\textwidth]{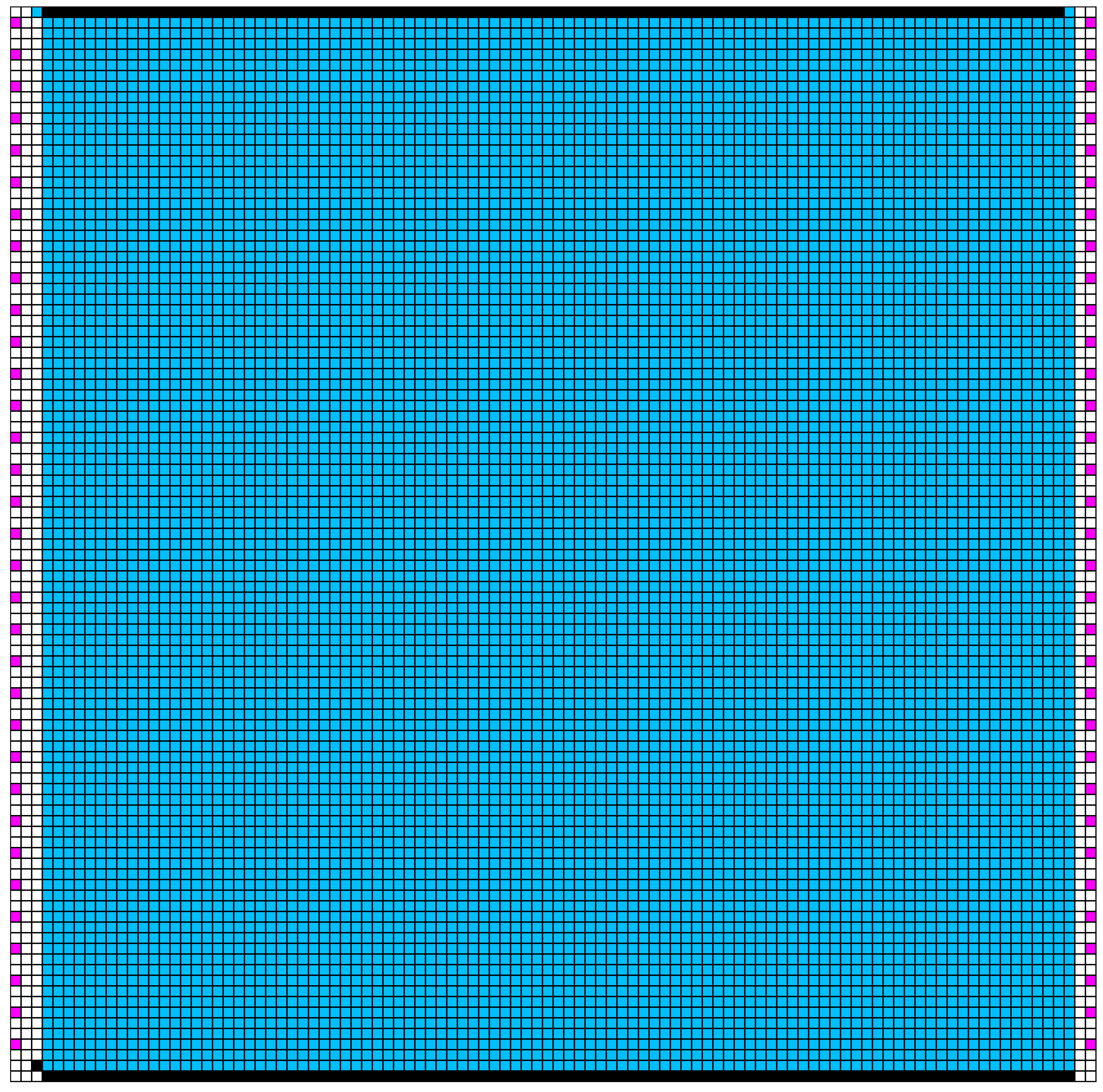}
        \caption{Warehouse (even)}
        \label{fig:warehouse-even-xxlarge-cma-es-unrepaired}
    \end{subfigure}
    \hfill
    \begin{subfigure}[t]{\EvalEnvShowSize\textwidth}
        \includegraphics[width=1\textwidth]{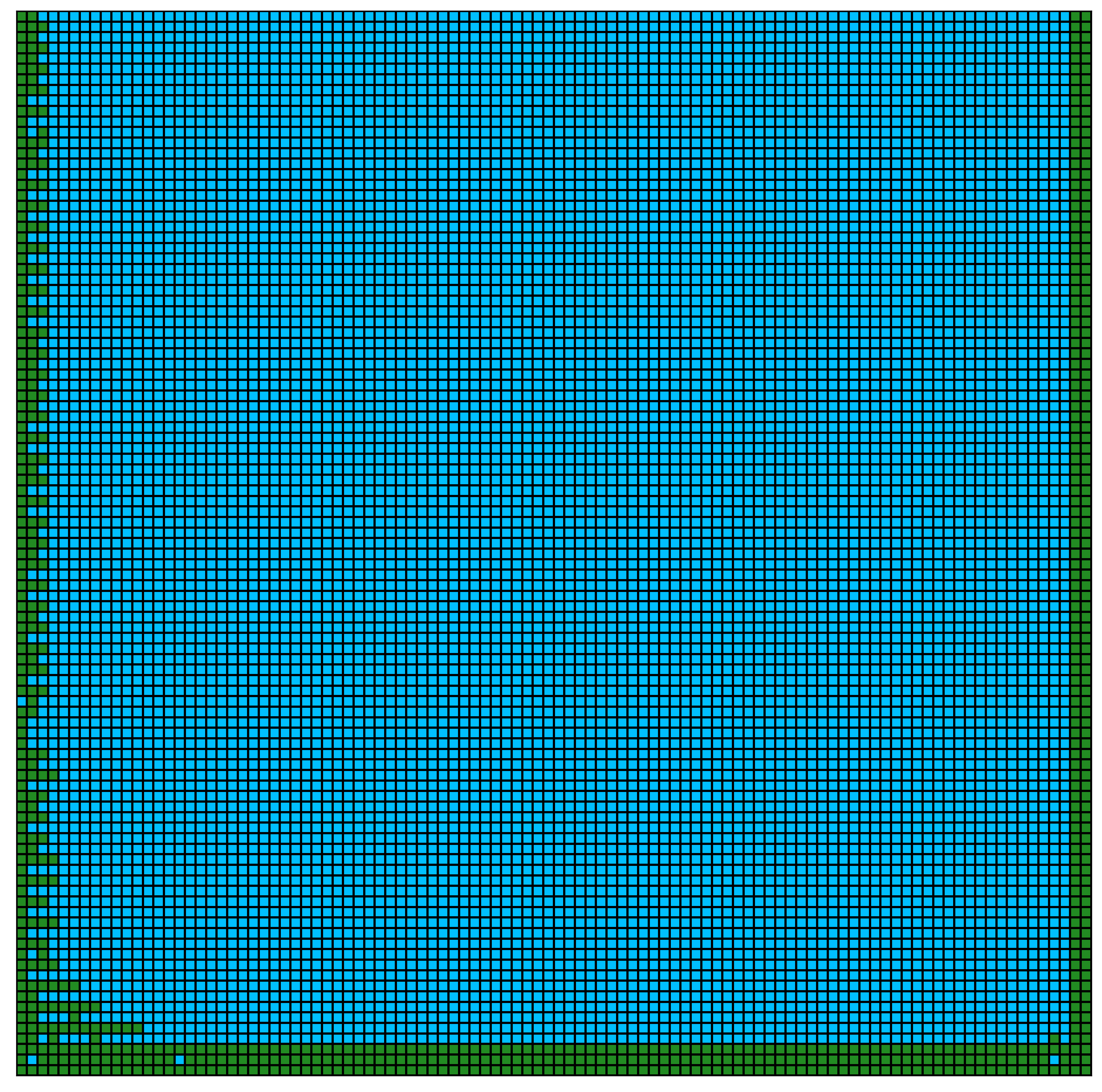}
        \caption{Manufacturing}
        \label{fig:manufacture-xxlarge-cma-es-unrepaired}
    \end{subfigure}
    \caption{Unrepaired NCA-generated environments of size $S_{eval}$ in the warehouse (even) and manufacturing domains for CMA-ES + NCA ($\alpha=5$).}
    \label{fig:xxlarge-cma-es-unrepaired}
\end{figure}

Derivative-free single objective optimizers such as CMA-ES~\cite{hansen2016cmaes} that do not diversify a given set of measures can also be used to train the NCA generators. To demonstrate the effect of diversifying measures on the trained NCA generators, we run CMA-ES in the warehouse and manufacturing domains with $\alpha = 5$ and compare the trained NCA generator with those trained by CMA-MAE. \Cref{tab:numerical-comp-cma-mae_cma-es} shows the numerical results, and \Cref{fig:comp_cma-es_cma-mae} shows the throughput over different numbers of agents. 

With an environment size of $S$, CMA-ES is competitive with CMA-MAE in all domains. In fact, CMA-ES ends up being slightly more scalable than CMA-MAE in the warehouse (uneven) domain with size $S$. However, CMA-MAE is significantly more scalable than CMA-ES with size $S_{eval}$. In particular, the MILP solver cannot find solutions for warehouse (even) and manufacturing domains of size $S_{eval}$ in the given computational budget (introduced in \Cref{appen:implementation}). 

\Cref{fig:xxlarge-cma-es-unrepaired} shows the unrepaired environments of size $S_{eval}$ in the warehouse (even) and manufacturing domains for CMA-ES. Both environments have a large number of endpoints. As a result, they rely on the MILP solver to both satisfy the domain-specific constraints and generate patterns, which takes a significant amount of time because almost no constraints are satisfied in the unrepaired environments. Given the deviation of the unrepaired environments, CMA-ES fails to optimize the similarity score, converging to a local optima while optimizing only the throughput. In comparison, CMA-MAE is less prone to falling into the deceptive local optima because of the diversity measures.

\section{Societal Impact} \label{appen:social-impact}

We propose using QD algorithms to generate arbitrarily large environments that enhance throughput beyond state-of-the-art environment optimization methods and human-designed environments. This is achieved by optimizing a diverse collection of NCA generators. Our method can be applied to generate any multi-robot system as long as an agent-based simulator is available for evaluation. 
Large companies such as Amazon and Alibaba have deployed multi-robot systems in warehouses to transport packages or inventory pods. Therefore, one real world application of our method is optimizing the layout of the automated warehouses to improve throughput. Since our method is agnostic to the specific agent simulator and only requires metrics such as throughput post-simulation, we can plug-in different simulators and apply our environment generation algorithm. Improving the throughput of their warehouses can present a significant economic impact on the industry.

Our method may have negative impacts. While designing a real-world automated warehouse or manufacturing environment and deploying large-scale multi-robot systems in reality, we shall take into account other factors such as safety measures. In our method, however, we ignore all factors except for the throughput and scalability of the environment. Consequently, while our method can contribute to optimizing operational efficiency, caution should be exercised to ensure that other crucial parameters are not compromised in real-world applications.

\end{document}